\documentclass[10pt,journal,compsoc]{IEEEtran}

\newif\ifFORM
\FORMfalse

\newif\iftr
\newif\ifcnf

\trfalse
\cnftrue

\trtrue
\cnffalse
 
\newif\ifnohl
\nohlfalse

\usepackage{etex}
\usepackage{relsize}

\newif\ifall    % Various stuff that might be useful but for now we don't want to use it
\newif\ifsq     % Squeeze space?
\newif\ifnonb   % Non blind submission
\newif\iftodos

\newif\ifsqCAP
\newif\ifsqVS
\newif\ifsqEN
\newif\ifsqTIT

\ifcnf
\sqtrue
\sqCAPtrue
\sqENtrue
\sqVStrue
\sqTITtrue

\else

\sqfalse
\sqCAPfalse
\sqENfalse
\sqVSfalse
\sqTITfalse

\fi

\newcommand{\vspaceSQ}[1]{\ifsqVS\vspace{#1}\fi}
\newcommand{\enlargeSQ}[1]{\ifsqEN\enlargethispage{\baselineskip}\fi}

\newcommand{\MspNN}[0]{\includegraphics[scale=0.4,trim=0 16 -8 0]{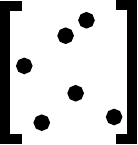}}

\newcommand{\MdnNK}[0]{\includegraphics[scale=0.4,trim=0 16 -8 0]{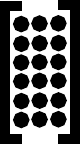}}
\newcommand{\MdnKN}[0]{\includegraphics[scale=0.4,trim=0 8 -8 0]{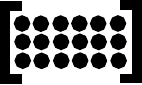}}
\newcommand{\MdnKK}[0]{\includegraphics[scale=0.4,trim=0 8 -8 0]{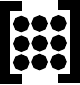}}
\newcommand{\MdnKO}[0]{\includegraphics[scale=0.4,trim=0 8 -8 0]{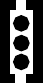}}
\newcommand{\MdnOK}[0]{\includegraphics[scale=0.4,trim=0 2 -8 0]{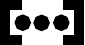}}

\ifcnf
\usepackage[font={scriptsize}]{caption}
\usepackage[font={scriptsize}]{subcaption}
\else
\usepackage[font=small]{caption}
\usepackage[font=small]{subcaption}
\fi

\usepackage{graphicx}
\usepackage{booktabs}
\usepackage{epstopdf}
\usepackage{float}
\usepackage{url}
\usepackage{placeins}
\usepackage{amsmath,amssymb,amsfonts}
\usepackage{mathtools,mathrsfs}
\usepackage{multirow}
\usepackage{rotating}
\usepackage{pbox}
\usepackage[normalem]{ulem}

\usepackage{inconsolata}
\usepackage{listings}

\newcommand{\subparagraph}{}
\usepackage[compact]{titlesec}
\titlespacing*{\section}{0pt}{6pt}{2pt}
\titlespacing*{\subsection}{0pt}{5pt}{1pt}
\titlespacing*{\subsubsection}{0pt}{5pt}{1pt}

\usepackage{etoolbox}

\makeatletter
\patchcmd{\ttlh@hang}{\parindent\z@}{\parindent\z@\leavevmode}{}{}
\patchcmd{\ttlh@hang}{\noindent}{}{}{}
\makeatother

\usepackage{cleveref}
\usepackage[utf8]{inputenc}
\crefname{section}{§}{§§}
\Crefname{section}{§}{§§}

\usepackage[10pt]{moresize}

\usepackage{xcolor}
\definecolor{darkgrey}{RGB}{70,70,70}
\definecolor{lightgrey}{RGB}{200,200,200}
\definecolor{lyellow}{RGB}{255,255,200}
\definecolor{ly}{RGB}{255,255,170}
\definecolor{darkred}{RGB}{150,0,0}
\definecolor{darkgreen}{RGB}{0,100,0}
\definecolor{darkblue}{RGB}{0,0,200}
\definecolor{lblue}{RGB}{100,100,255}

\usepackage{enumitem}

\usepackage{makecell}

\usepackage[linesnumbered,ruled,vlined]{algorithm2e}
\SetAlFnt{\scriptsize\tt}
\SetAlCapFnt{\scriptsize\sf}
\SetAlCapNameFnt{\scriptsize\sf}

\usepackage{algpseudocode}
\algblock{ParFor}{EndParFor}
\algnewcommand\algorithmicparfor{\textbf{parfor}}
\algnewcommand\algorithmicpardo{\textbf{do}}
\algnewcommand\algorithmicendparfor{\textbf{end\ parfor}}
\algrenewtext{ParFor}[1]{\algorithmicparfor\ #1\ \algorithmicpardo}
\algrenewtext{EndParFor}{\algorithmicendparfor}

\newcommand{\maciej}[1]{\textcolor{blue}{[Maciej: #1]}}

\usepackage[hang,flushmargin]{footmisc}

\usepackage{bm}

\usepackage{scalerel,stackengine}
\stackMath
\newcommand\rwh[1]{%
\savestack{\tmpbox}{\stretchto{%
  \scaleto{%
      \scalerel*[\widthof{\ensuremath{#1}}]{\kern-.6pt\bigwedge\kern-.6pt}%
          {\rule[-\textheight/2]{1ex}{\textheight}}%WIDTH-LIMITED BIG WEDGE
            }{\textheight}% 
}{0.5ex}}%
\stackon[1pt]{#1}{\tmpbox}%
}

\def\HiLiGA{\leavevmode\rlap{\hbox to \hsize{\color{black!10}\leaders\hrule height 1\baselineskip depth 1ex\hfill}}}
\def\HiLiGB{\leavevmode\rlap{\hbox to \hsize{\color{black!25}\leaders\hrule height 1\baselineskip depth 1ex\hfill}}}
\def\HiLiGC{\leavevmode\rlap{\hbox to \hsize{\color{black!40}\leaders\hrule height 1\baselineskip depth 1ex\hfill}}}
\def\HiLiGD{\leavevmode\rlap{\hbox to \hsize{\color{black!55}\leaders\hrule height 1\baselineskip depth 1ex\hfill}}}
\def\HiLiGE{\leavevmode\rlap{\hbox to \hsize{\color{black!70}\leaders\hrule height 1\baselineskip depth 1ex\hfill}}}
\def\HiLiGF{\leavevmode\rlap{\hbox to \hsize{\color{black!85}\leaders\hrule height 1\baselineskip depth 1ex\hfill}}}

\usepackage{algpseudocode}
\usepackage{tikz}
\usetikzlibrary{calc}
\usepackage{xcolor}
\makeatletter

\newcommand{\fRB}[1]{\left(#1\right)}
\newcommand{\fSB}[1]{\left[#1\right]}

\usepackage{fontawesome}
\usepackage{pifont}
\usepackage{textcomp}

\usepackage{xcolor}
\usepackage{soul}
\usepackage{dblfloatfix}

\usepackage{colortbl}

\ifcnf
\usepackage[sortcites, firstinits=true]{biblatex}
\ifcnf

\else

\fi
\fi

\newcommand{\noAnswer}{\textcolor{lightgray}{\faQuestionCircle}}
\newcommand{\comment}[1]{\ignorespaces}

\usepackage{physics}

\nohltrue

\ifnohl

\renewcommand{\hl}[1]{#1}
\renewcommand{\rowcolor}[1]{}
\renewcommand{\marginpar}[1]{}
\renewcommand{\colorbox}[2]{#2}

\fi

\newcolumntype{y}{>{}l}

\IEEEaftertitletext{\vspace{-2\baselineskip}}

\newif\ifHL
\HLfalse

\newcommand{\faY}[0]{\faBatteryFull}
\newcommand{\faH}[0]{\faBatteryHalf}
\newcommand{\faN}[0]{\faTimes}
\newcommand{\faU}[0]{\noAnswer}

\begin{document}

%\title{Parallel and Distributed Graph Neural Networks: An In-Depth Concurrency Analysis of Computational Complexity}
\title{Parallel and Distributed Graph Neural Networks: An In-Depth Concurrency Analysis}
%\title{Parallel and Distributed Graph Neural Networks: An In-Depth Concurrency Analysis of Models, Abstractions, Frameworks, and Accelerators}

% \iftr
% \author{Maciej Besta$^1$, Marc Fischer$^2$, Vasiliki Kalavri$^3$, Michael Kapralov$^4$, Torsten Hoefler$^1$\\
% \vspaceSQ{0.25em}{$^1$Department of Computer Science, ETH Zurich\\
% $^2$PRODYNA (Schweiz) AG;\\
% $^3$Department of Computer Science, Boston University\\
% $^4$School of Computer and Communication Sciences, EPFL\vspaceSQ{-0.25em}}}
% \else
\author{Maciej Besta and Torsten Hoefler\\
{\vspaceSQ{0.25em}{Department of Computer Science, ETH Zurich}}\vspace{-0.5em}}
% \fi

\IEEEtitleabstractindextext{%
\begin{abstract}
Graph neural networks (GNNs) are among the most powerful tools in deep
learning. They routinely solve complex problems on unstructured
networks, such as node classification, graph classification, or link
prediction, with high accuracy. However, both inference and training of GNNs are 
complex, and they uniquely combine the features of irregular graph processing
with dense and regular computations. 
%
\if 0
The complexity behind GNN computations is
further aggravated by the plethora of existing models and the sheer number of
different operations within each such model: aggregation, projection,
activation, update, and others.
\fi
%
This complexity makes it very challenging to execute GNNs efficiently on modern
massively parallel architectures.
To alleviate this, we first design a taxonomy of parallelism in GNNs,
considering data and model parallelism, and different forms of pipelining.
Then, we use this taxonomy to investigate the amount of parallelism in numerous
GNN models, GNN-driven machine learning tasks, software frameworks, or hardware
accelerators. We use the work-depth model, and we also assess
communication volume and synchronization.
We specifically focus on the sparsity/density of the associated tensors, in
order to understand how to effectively apply techniques such as vectorization.
We also formally analyze GNN pipelining, and we generalize the established
Message-Passing class of GNN models to cover arbitrary pipeline depths,
facilitating future optimizations.  Finally, we investigate different forms of
asynchronicity, navigating the path for future asynchronous parallel GNN
pipelines. 
The outcomes of our analysis are synthesized in a set of insights that help to
maximize GNN performance, and a comprehensive list of challenges and
opportunities for further research into efficient GNN computations.
Our work will help to advance the design of future GNNs.
\vspace{-0.5em}
\end{abstract}

\iftr
\fi
}

% make the title area
\maketitle

\IEEEdisplaynontitleabstractindextext
\IEEEpeerreviewmaketitle

\iftr
%{\vspace{-1.0em}\noindent \textbf{This is an extended version of a paper published at\\ ACM HPDC'15 under the same title}}
\else
{\vspace{-1.0em}\noindent \textbf{An extended version: \url{https://arxiv.org/abs/2205.09702}}\vspace{1em}}
\fi

\vspaceSQ{-1em}
\section{Introduction}
\label{sec:intro}
\vspaceSQ{-0.5em}

\iftr
Graph neural networks (GNNs) are taking over the world of machine learning (ML) by
storm~\cite{chami2020machine, wu2020comprehensive}. They have been used in a
plethora of complex problems such as node classification, graph classification,
or edge prediction~\cite{hu2020open, dwivedi2020benchmarking}. Example areas of
application are social sciences (e.g., studying human interactions),
bioinformatics (e.g., analyzing protein structures), chemistry (e.g., designing
compounds), medicine (e.g., drug discovery), cybersecurity (e.g., identifying
intruder machines), entertainment services (e.g., predicting movie popularity),
linguistics (e.g., modeling relationships between words), transportation (e.g.,
finding efficient routes), and others~\cite{wu2020comprehensive, zhou2020graph,
zhang2020deep, chami2020machine, hamilton2017representation,
bronstein2017geometric, cook2006mining, jiang2013survey, horvath2004cyclic,
chakrabarti2006graph, besta2021motif, gianinazzi2021learning}.
Some recent celebrated success stories are cost-effective and fast placement of
high-performance chips~\cite{mirhoseini2021graph}, simulating complex
physics~\cite{pfaff2020learning, sanchez2020learning}, guiding mathematical
discoveries~\cite{davies2021advancing}, or significantly improving the accuracy
of protein folding prediction~\cite{jumper2021highly}. 
\else
Graph neural networks (GNNs) are taking over the world of machine learning (ML) by
storm~\cite{wu2020comprehensive}. They have been used in a
plethora of complex problems such as node classification, graph classification,
or edge prediction. Example areas of
application are social sciences,
bioinformatics, chemistry, medicine, cybersecurity,
linguistics, transportation, and others~\cite{wu2020comprehensive}.
Some recent celebrated success stories are cost-effective and fast placement of
high-performance chips~\cite{mirhoseini2021graph}, guiding mathematical
discoveries~\cite{davies2021advancing}, or significantly improving the accuracy
of protein folding prediction~\cite{jumper2021highly}. 
\fi

\iftr
GNNs uniquely generalize both \emph{traditional deep learning
(DL)}~\cite{goodfellow2016deep, lecun2015deep, ben2019modular} and
\emph{graph processing}~\cite{lumsdaine2007challenges, sakr2021future,
gregor2005parallel}. 
Still, contrarily to the former, they do not operate on regular grids and
highly structured data (such as, e.g., image processing); instead, the
data in question is highly unstructured, irregular, and the resulting
computations are data-driven and lacking straightforward spatial or temporal
locality~\cite{lumsdaine2007challenges}. 
Moreover, contrarily to the latter, vertices and/or edges are associated with
complex data and processing. For example, in many GNN models, each vertex~$i$
has an assigned $k$-dimensional \emph{feature vector}, and each such vector is
combined with the vectors of $i$'s neighbors; this process is repeated
iteratively.  Thus, while the overall style of such GNN computations resembles
label propagation algorithms such as PageRank~\cite{page1999pagerank,
besta2017push}, it comes with additional complexity due to the high
dimensionality of the vertex features. 
\else
GNNs generalize both \emph{traditional deep learning
(DL)} and
\emph{graph processing}. 
Contrarily to the former, they do not operate on regular grids and
highly structured data (such as, e.g., image processing); instead, the
data is highly unstructured, irregular, and the resulting
computations are data-driven and lacking straightforward spatial or temporal
locality~\cite{lumsdaine2007challenges}. 
Moreover, contrarily to the latter, vertices and/or edges are associated with
complex data and processing. For example, in many GNN models, each vertex~$i$
has an assigned $k$-dimensional \emph{feature vector}, and each such vector is
combined with the vectors of $i$'s neighbors; this process is repeated
iteratively.  Thus, while the overall style of such GNN computations resembles
label propagation algorithms such as PageRank~\cite{page1999pagerank}, it comes
with additional complexity due to the high dimensionality of the vertex
features. 
\fi

Yet, this is only how \emph{the simplest} GNN models, such as basic Graph
Convolution Networks (GCN)~\cite{kipf2016semi}, work.  In many, if not most,
GNN models, high-dimensional data may also be attached to every edge, and
complex updates to the edge data take place at every iteration. For example, in
the Graph Attention Network (GAT) model~\cite{velivckovic2017graph}, to compute
the scalar weight of a single edge~$(i,j)$, one must first concatenate linear
transformations of the feature vectors of both vertices $i$ and $j$, and then
construct a dot product of such a resulting vector with a trained parameter
vector.
Other models come with even more complexity. For example, in Gated Graph
ConvNet (G-GCN)~\cite{bresson2017residual} model, the edge weight may be a
multidimensional vector.

\if 0
%
The above challenges are
further aggravated, for example possibly complex functions
for aggregating vertex feature vectors or for generating edge feature data,
  non-linear activations, projecting feature vectors, and others. These are all
  important parts of a GNN model, but their combination results in a highly
  complicated computation both in inference and in
  training~\cite{abadal2021computing}.
%
\fi

\iftr
\ifcnf
\begin{table}[h]
\else
\begin{table*}[t]
\fi
\vspaceSQ{-0.25em}
\centering
\setlength{\tabcolsep}{1pt}
\footnotesize
\ifcnf
\scriptsize
\ssmall
\sf
\fi
\begin{tabular}{ll@{}}
\toprule
\multicolumn{2}{c}{\textbf{Structure of graph inputs}} \\
\midrule
 $G = (V,E)$ & \makecell[l]{A graph; $V$ and $E$ are sets of vertices and edges.}\\
% $w(e)$ & \makecell[l]{The weight of an edge $e = (u,v)$.}\\
 $n, m$ & Numbers of vertices and edges in $G$; $|V| = n, |E| = m$.\\
 $N(i), N^+(i), \widehat{N}(i)$ & Neighbors of $i$, in-neighbors of $i$, and $\widehat{N}(i) = N(i) \cup \{i\}$.\\
% $N^+(i)$ & The set of incoming neighbors attached to $i$ ($i$'s in-neighbors).\\
% $\widehat{N}(i)$ & $i$'s neighbors together with $i$: $\widehat{N}(i) = N(i) \cup \{i\}$.\\
% $N(i)^s$ & The set of $s$-hop neighbors of a vertex~$i$.\\
%
\ifFORM
 $\overline{N(i)}^s$ & $\overline{N(i)}^s \equiv N(i) \cup N(i)^2 \cup ... \cup N(i)^s$.\\
 $\overline{\widehat{N}(i)}^s$ & $\overline{\widehat{N}(i)}^s \equiv \widehat{N}(i) \cup N(i)^2 \cup ... \cup N(i)^s$.\\
\fi
 $d_i, d$ & The degree of a vertex $i$ and the maximum degree in a graph.\\
%
%\midrule
%
%\multicolumn{2}{c}{\textbf{Graph structure (matrices)}} \\
%
%\midrule
%
 $\mathbf{A}, \mathbf{D} \in \mathbb{R}^{n \times n}$ & The graph adjacency and the degree matrices.\\
% $\mathbf{A} \in \{0,1\}^{n \times n}$ & The adjacency matrix of an unweighted graph.\\
% $\mathbf{D} \in \mathbb{R}^{n \times n}$ & The degree matrix; $\mathbf{D}(i,i) = d(i)$.\\
 $\mathbf{\widetilde{A}}, \mathbf{\widetilde{D}}$ & $\mathbf{A}$ and $\mathbf{D}$ matrices with self-loops ($\mathbf{\widetilde{A}} = \mathbf{A} + \mathbf{I}$, $\mathbf{\widetilde{D}} = \mathbf{D} + \mathbf{I}$).\\
%  $\mathbf{\widetilde{D}}$ & The degree matrix including self-loops.\\
 $\mathbf{\widehat{A}}, \mathbf{\overline{A}}$ & Normalization: $\mathbf{\widehat{A}} = \mathbf{\widetilde{D}}^{-\frac{1}{2}} \mathbf{\widetilde{A}} \mathbf{\widetilde{D}}^{-\frac{1}{2}}$ and $\mathbf{\overline{A}} = \mathbf{D}^{-1} \mathbf{A}$~\cite{wu2020comprehensive}.\\
% $\mathbf{L}$ & The Laplacian matrix; $\mathbf{L} = \mathbf{D} - \mathbf{A}$.\\
% $\delta_{ij}$ & Kronecker delta ($\delta_{ij} = 1$ if $i = j$, $\delta_{ij} = 0$ otherwise)\\
%
%
\midrule
\multicolumn{2}{c}{\textbf{Structure of GNN computations}} \\
\midrule
$L, k$ & The number of GNN layers and input features.\\
% $k_{(l)}$ & Dimensionality of a vertex feature vector (layer $l$). \\
$\mathbf{X} \in \mathbb{R}^{n \times k}$ & Input (vertex) feature matrix. \\
%$\mathbf{x}_i \in \mathbb{R}^{n}$ & The input feature vector of a vertex $i$. \\
$\mathbf{Y}, \mathbf{H}^{(l)} \in \mathbb{R}^{n \times O(k)}$ & Output (vertex) feature matrix, hidden (vertex) feature matrix. \\
%$\mathbf{y}_i \in \mathbb{R}^{n}$ & The output feature vector of a vertex $i$. \\
$\mathbf{x}_i, \mathbf{y}_i, \mathbf{h}^{(l)}_i \in \mathbb{R}^{n}$ & Input, output, and hidden feature vector of a vertex $i$ (layer $l$). \\
%$\mathbf{h}^{(l)}_i \in \mathbb{R}^{n}$ & The hidden feature vector of a vertex $i$ (layer $l$). \\
% $\mathbf{W}, \mathbf{W}_\cdot$ & A parameter matrix.\\ 
$\mathbf{W}^{(l)} \in \mathbb{R}^{O(k) \times O(k)}$ & A parameter matrix in layer $l$.\\ 
$\sigma(\cdot)$ & Element-wise activation and/or normalization.\\ 
% $Norm(\cdot)$ & Normalization (details depend on a model).\\ 
% $\psi, \phi, \bigoplus$ & Different operations (model dependant).\\ 
%
%\midrule
%
%\multicolumn{2}{c}{\textbf{Linear algebra operations}} \\
%
%\midrule
%
$\times$, $\odot$ & Matrix multiplication and element-wise multiplication.\\
\bottomrule
\end{tabular}
\vspaceSQ{-1em}
\caption{The most important symbols used in the paper.}
\vspaceSQ{-1em}
\label{tab:symbols}
%\vspace{-0.5em}
\ifcnf
\end{table}
\else
\end{table*}
\fi

\fi

\iftr
At the same time, \emph{parallel} and \emph{distributed} processing have
essentially become synonyms for computational efficiency. Virtually each modern
computing architecture is parallel: cores form a socket while sockets form a
non-uniform memory access (NUMA) compute node. Nodes may be further clustered
into blades, chassis, and racks~\cite{besta2014fault, schweizer2015evaluating,
gerstenberger2013enabling}. Numerous memory banks enable data
\emph{distribution}. All these parts of the architectural hierarchy run in
parallel.
\else
At the same time, \emph{parallel} and \emph{distributed} processing have
essentially become synonyms for computational efficiency. Virtually each modern
computing architecture is parallel: cores form a socket while sockets form a
non-uniform memory access (NUMA) compute node. Nodes may be further clustered
into blades, chassis, and racks. Numerous memory banks enable data
\emph{distribution}. All these parts of the architectural hierarchy run in
parallel.
\fi 
Even a single sequential core offers parallelism in the form of
\emph{vectorization}, \emph{pipelining}, or \emph{instruction-level parallelism
(ILP)}.
On top of that, such architectures are often \emph{heterogeneous}: Processing
units can be CPUs or GPUs, Field Programmable Gate Arrays (FPGAs), or others.
%
% are characterized by different performance properties. For example, CPUs are
% latency-oriented, GPUs are throughput-oriented, while FPGAs offer massive
% parallelism while having lower clock frequencies and thus higher latency.
% The same 
%
\emph{How to harness all these rich features to
achieve more performance in GNN workloads?}

To help answer this question, we systematically analyze different aspects of
GNNs, focusing on \emph{the amount of parallelism and distribution} in these
aspects. We use fundamental theoretical parallel computing machinery, for
example the Work-Depth model~\cite{blelloch2010parallel}, to reveal
architecture independent insights. We put special focus on the
linear algebra formulation of computations in GNNs, and we investigate the
sparsity and density of the associated tensors. This offers further insights
into performance-critical features of GNN computations, and facilitates
applying parallelization mechanisms such as vectorization.
\emph{In general, our investigation will help to develop more efficient GNN
computations.}

For a systematic analysis, \emph{we propose an in-depth taxonomy of
parallelism in GNNs}. The taxonomy identifies fundamental forms of parallelism
in GNNs. While some of them have direct equivalents in traditional
deep learning, we also illustrate others that are specific to GNNs.

To ensure wide applicability of our analysis, we cover a large number
of different aspects of the GNN landscape. Among others, we consider different
categories of GNN models (e.g., spatial, spectral, convolution, attentional,
message passing), a large selection of GNN models (e.g.,
GCN~\cite{kipf2016semi}, SGC~\cite{wu2019simplifying},
GAT~\cite{velivckovic2017graph}, G-GCN~\cite{bresson2017residual}), parts of
GNN computations (e.g., inference, training), building blocks of GNNs (e.g.,
layers, operators/kernels), programming paradigms (e.g.,
SAGA-NN~\cite{ma2019neugraph}, GReTA~\cite{kiningham2020greta}), execution
schemes behind GNNs (e.g., reduce, activate, different tensor operations), GNN
frameworks (e.g., NeuGraph~\cite{ma2019neugraph}), GNN accelerators (e.g.,
HyGCN~\cite{yan2020hygcn}) GNN-driven ML tasks (e.g., node classification, edge
prediction), 
mini-batching vs.~full-batch training, different forms of
sampling, and asynchronous GNN pipelines.

We finalize our work with \emph{general insights} into parallel
and distributed GNNs, and a set of \emph{research challenges and
opportunities}.
Thus, our work can serve as a guide when developing parallel and distributed
solutions for GNNs executing on modern architectures, and for choosing the next
research direction in the GNN landscape.

%\marginpar{\large\vspace{6em}\colorbox{yellow}{\textbf{R1}}}

\noindent
{Overall, the central contributions of our work are:}

\begin{itemize}[leftmargin=0.75em]
\item {We identify and analyze fundamental forms of parallelism in
GNNs, and we illustrate that they -- to some degree -- match those in
traditional DL. This will foster designing future GNN
systems more effectively, by empowering system designers with a
clear view of the space of parallelization approaches that they could
use, and how these approaches can be combined. Moreover, it will facilitate
reusing existing large-scale DL frameworks.}
\item {We analyze a broad spectrum of GNN models formally (covering all
major classes of models, i.e., Convolution, Attention, Message-Passing, and
Linear/Polynomial/Rational ones), for a total of 23 models, investigating how
parallelizable they are, and identifying their bottlenecks and the associated
tradeoffs. This will facilitate scaling these models to much larger sizes than
what is done today. It is an important factor in making them more powerful, as
indicated by the recent successes of large NLPs.}
\item {We design a broad theoretical framework for asynchronous GNNs, which
will serve as a blueprint for novel GNN models and implementations that will further
push the scalability and performance of GNNs.}
\item {We review challenges and opportunities, which 
will facilitate future research into large-scale GNNs.}
\end{itemize}

\if 0
%
\vspace{0.5em}
\noindent 
In general, we provide the following contributions:

\begin{itemize}[leftmargin=1em, noitemsep]
%
\item We provide the first taxonomy of different forms of parallelism and distribution in
the GNN landscape.
%
\item We conduct the first in-depth concurrency analysis that assesses the amount of
parallelism in different parts of the proposed taxonomy. The analysis results
in \emph{insights} and \emph{recommendations} into how to best harness the
architectural parallelism available in modern compute resources, when executing
complex GNN workloads.
%
\item We conduct the first analysis of distribution in GNN computations. Similarly, we
also offer insights into how to best utilize the available distributed compute
resources for more performance and scale in GNN workloads.
%
\end{itemize}
%
\fi

\iftr

\subsection{Complementary Analyses}

We discuss related works on the \emph{theory} and
\emph{applications} of GNNs. 
There exist general GNN surveys~\cite{bronstein2021geometric,
wu2020comprehensive, zhou2020graph, zhang2020deep, chami2020machine,
hamilton2017representation, bronstein2017geometric, zhang2019graph}, works on
theoretical aspects (spatial--spectral dichotomy~\cite{chen2021bridging,
balcilar2020bridging}, the expressive power of GNNs~\cite{sato2020survey}, or
heterogeneous graphs~\cite{yang2020heterogeneous, xie2021survey}), analyzes of
GNNs for specific applications (knowledge graph
completion~\cite{arora2020survey}, traffic forecasting~\cite{jiang2021graph,
tedjopurnomo2020survey}, symbolic computing~\cite{lamb2020graph}, recommender
systems~\cite{wu2020graph}, text classification~\cite{huang2019text}, or action
recognition~\cite{ahmad2021graph}), explainability of
GNNs~\cite{yuan2020explainability}, and on software (SW) and hardware (HW)
accelerators and SW/HW co-design~\cite{abadal2021computing}. We complement
these works as we focus on parallelism and distribution of GNN workloads.
\fi

\iftr

\subsection{Scope of this Work \& Related Domains}

We focus on GNNs, but we also cover parts of
the associated domains.
In the graph \textbf{embeddings} area, one develops methods for
finding low-dimensional representations of elements of graphs, most often
vertices~\cite{wang2020survey, cui2018survey, wang2017knowledge,
dai2020survey}. As such, GNNs can be seen as a part of this area, because one
can use a GNN to construct an embedding~\cite{wu2020comprehensive}.
However, we exclude non-GNN related methods for constructing embeddings, such
as schemes based on random walks~\cite{perozzi2014deepwalk, grover2016node2vec}
or {graph kernel} designs~\cite{vishwanathan2010graph, borgwardt2007graph,
kriege2020survey}. 

\fi

\section{Graph Neural Networks: Overview}
\label{sec:gnns}

We first overview GNNs; Table~\ref{tab:symbols} provides notation.
\iftr
We start by summarizing a GNN computation and GNN-driven downstream ML tasks
(\cref{sec:gnns-summary}).
We then discuss different parts of a GNN computation in more detail, providing
both the basic knowledge and general opportunities for parallelism and
distribution.
This includes the input GNN datasets (\cref{sec:gnns-inputs}), the mathematical
theory and formulations for GNN models that form the core of GNN computations
(\cref{sec:gnns-math}), GNN inference vs.~GNN training (\cref{sec:gnn-inf-tr}),
and the programmability aspects (\cref{sec:gnns-prog}). We finish with a
taxonomy of parallelism in GNNs (\cref{sec:gnns-taxonomy}) and parallel \&
distributed theory used for formal analyses (\cref{sec:par-algs}).
%
% We finish with a taxonomy of parallelism in GNNs (\cref{sec:gnns-taxonomy}).
%
\fi

\ifcnf

\fi

\if 0
%
\begin{figure*}[h]
%
\includegraphics[width=1.0\textwidth]{parallelism-overview-2.pdf}
\vspaceSQ{-1.5em}
\caption{\textbf{Overview of parallelism in traditional neural networks (left) and in GNNs (right).}
We provide pointers to parts of the paper with more details about each form of parallelism.}
\label{fig:par-overview}
\vspaceSQ{-1em}
%
\end{figure*}
%
\fi

\vspaceSQ{-0.5em}
\subsection{GNN Computation: A High-Level Summary}
\label{sec:gnns-summary}

\iftr
\begin{figure}[h]
\else
\begin{figure}[b]
\fi
\vspaceSQ{-1.5em}
\includegraphics[width=1.0\columnwidth]{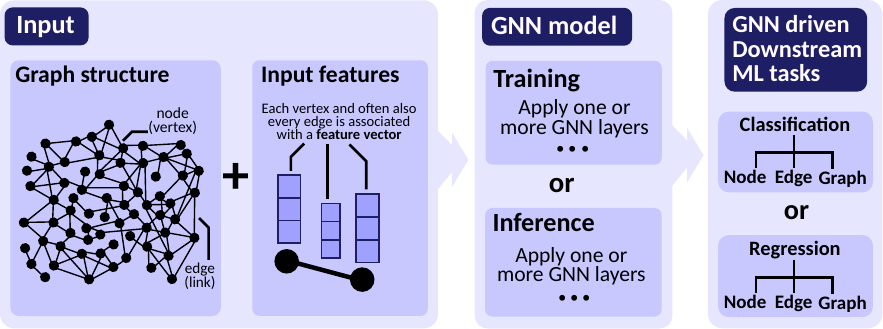}
\vspace{-1.5em}
\caption{\textbf{(\cref{sec:gnns-summary}) Overview of general GNN computation.}
Input comprises the graph structure and the accompanying feature vectors
(assigned to vertices/edges). The input is processed using a specific
GNN model (training or inference). Output
feature vectors are used in various downstream ML tasks.}
\label{fig:gnn-overview}
\vspaceSQ{-1em}
\end{figure}

We overview a \textbf{GNN computation} in Figure~\ref{fig:gnn-overview}.
The input is a \emph{graph dataset}, which can be a single graph (usually a
large one, e.g., a brain network), or several graphs (usually many small ones,
e.g., chemical molecules). The input usually comes with \emph{input feature
vectors} that encode the semantics of a given task. For example, if the input
nodes and edges model -- respectively -- papers and citations between these
papers, then each node could come with an input feature vertex being a one-hot
bag-of-words encoding, specifying the presence of words in the abstract of a
given publication.
Then, a \emph{GNN model} -- underlying the training and inference process --
uses the graph structure and the input feature vectors to generate the
\emph{output feature vectors}.
In this process, intermediate \emph{hidden latent vectors} are often created.
Note that hidden features may be updated \emph{iteratively} more than once (we
refer to a single such iteration, that updates all the hidden features, as a
\emph{GNN layer}). 
The output feature vectors are then used for the \emph{downstream ML tasks}
such as node classification or graph classification.

\begin{figure}[h]
\centering
\includegraphics[width=0.8\columnwidth]{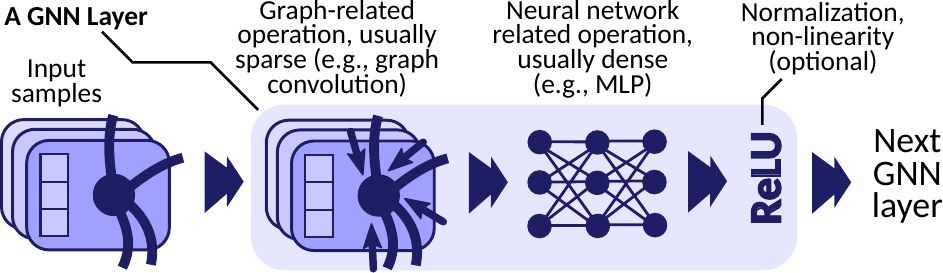}
\vspaceSQ{-0.5em}
\caption{\textbf{(\cref{sec:gnns-summary}) Overview of one GNN layer.} 
The input samples (e.g., vertices or graphs) are processed
with a graph-related operation such as graph convolution,
followed by a neural network related operation such as an MLP,
then optionally by a non-linearity such as ReLU, and potentially 
by some normalization.}
\label{fig:gnn-layer}
%\vspaceSQ{-0.5em}
%
\end{figure}

\if 0

\begin{figure}[h]
%
\centering
\includegraphics[width=0.8\columnwidth]{gnn-training-inference.pdf}
\vspaceSQ{-0.5em}
%
\caption{\textbf{(\cref{sec:gnns-summary}) Overview of GNN training and
inference.}} 
%
\label{fig:gnn-training-inference}
%
\end{figure}

\fi

A single \textbf{GNN layer} is summarized in Figure~\ref{fig:gnn-layer}.
In general, one first applies a certain \emph{graph-related} operation to the
features. For example, in the GCN model~\cite{kipf2016semi}, one aggregates the
features of neighbors of each vertex~$v$ into the feature vector of~$v$ using
summation.
Then, a selected operation related to \emph{traditional neural networks} is
applied to the feature vectors.  A common choice is an MLP or a plain linear
projection.
Finally, one often uses some form of non-linear activation (e.g.,
ReLU~\cite{kipf2016semi}) and/or normalization.

\begin{figure}[t]
\vspaceSQ{-1em}
\centering
\includegraphics[width=0.8\columnwidth]{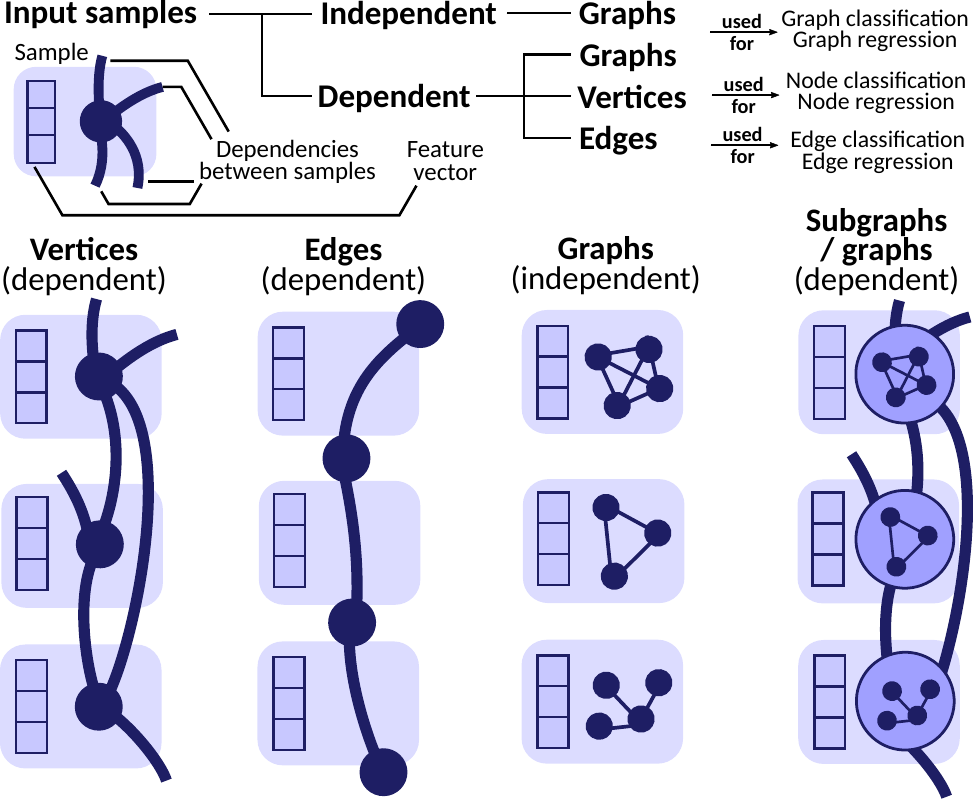}
\vspaceSQ{-0.5em}
\caption{\textbf{(\cref{sec:gnns-summary}) Overview of GNN samples.} GNN
downstream ML tasks aim at classification or regression of vertices, edges, or
graphs. While both vertex and edge samples virtually always have inter-sample
dependencies, graphs may be both dependent and independent.}
\label{fig:gnn-samples}
\vspaceSQ{-1.5em}
\end{figure}

One key difference between GNNs and traditional deep learning are
\emph{possible dependencies between input data samples} which make the
parallelization of GNNs much more challenging.
We show \textbf{GNN data samples} in Figure~\ref{fig:gnn-samples}.
A single sample can be a node (a vertex), an edge (a link), a subgraph, or a
graph itself.
One may aim to classify samples (assign labels from a discrete set) or conduct
regression (assign continuous values to samples).
Both vertices and edges have inter-dependencies: vertices are connected with
edges while edges share common vertices.
The seminal work by Kipf and Welling~\cite{kipf2016semi} focuses on node
classification. Here, one is given a single graph as input, data samples are
single vertices, and the goal is to classify all unlabeled vertices.

%\marginpar{\large\vspace{5em}\colorbox{yellow}{\textbf{R1}}\\\colorbox{yellow}{\textbf{(A)}}}

\iftr
Graphs -- when used as basic data samples -- are usually
independent~\cite{ying2018hierarchical, xinyi2018capsule}
(cf.~Figure~\ref{fig:gnn-samples}, 3rd column). 
\else
Graphs -- when used as basic data samples -- are usually
independent~\cite{ying2018hierarchical}
(cf.~Figure~\ref{fig:gnn-samples}, 3rd column). 
\fi
An example use case is
classifying chemical molecules. This setting
resembles traditional deep learning (e.g., image recognition), where samples
(single pictures) also have
no explicit dependencies.
Note that, as chemical molecules may differ in sizes, load balancing issues may
arise {(we discuss it in Section~\mbox{\ref{sec:mb-par}})}.
This also has analogies in traditional deep learning, e.g., sampled videos also may
have varying sizes~\cite{li2020taming}.
Graph classification may also feature graph samples \emph{with}
inter-dependencies (cf.~Figure~\ref{fig:gnn-samples}, 4th column). This is
useful when studying, for example, relations between network
communities~\cite{li2019semi}{; see Section~\mbox{\ref{sec:mb-par}} for details}.

%\marginpar{\large\vspace{-2em}\colorbox{yellow}{\textbf{R1}}\\\colorbox{yellow}{\textbf{(A)}}}

\if 0
In graph classification without dependencies, one can pick graph samples for
training independently. Hence, standard mini-batch data parallelism is
applicable. It becomes more  dependencies between samples.
This is because -- to process a given mini-batch -- we must consider graph
samples not only within but also possibly outside that mini-batch.
%
We discuss this aspect in more detail in Section~\ref{sec:data-par}. 
\fi

\iftr
We illustrate examples and taxonomy of GNN tasks in Figure~\ref{fig:gnn-tasks-dets}. 

\begin{figure*}[hbtp]
	\vspace{-1em}
\includegraphics[width=1.0\textwidth]{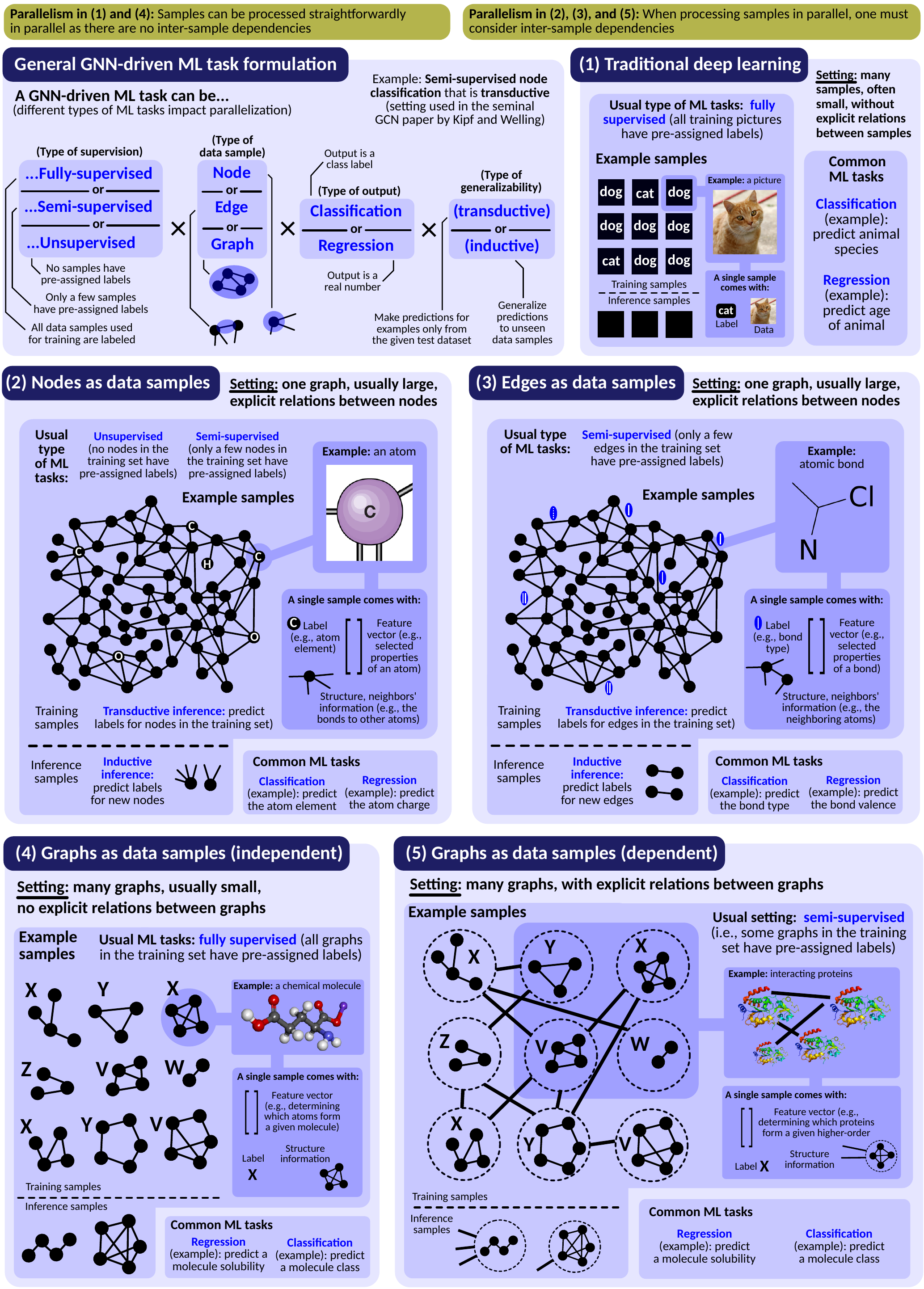}
\vspace{-1.5em}
	\caption{\textbf{Examples and taxonomy of GNN tasks.}}
\label{fig:gnn-tasks-dets}
\vspaceSQ{-1em}
\end{figure*}
\fi

\subsection{Input Datasets \& Output Structures in GNNs}
\label{sec:gnns-inputs}

A GNN computation starts with the input graph
$G$, modeled as a tuple $(V,E)$; $V$ is a set of vertices and
$E \subseteq V \times V$ is a set of edges; $|V|=n$ and $|E|=m$. $N(v)$ denotes
the set of vertices adjacent to vertex (node)~$v$, $d_v$ is $v$'s degree, and
$d$ is the maximum degree in $G$ (all symbols are listed in Table~\ref{tab:symbols}). 
%
% If $G$ is weighted, it is modeled by a tuple $(V,E,w)$. Then, $w(e)$ is the
% weight of an edge $e \in E$. 
%
The {adjacency matrix} (AM) of a graph is $\mathbf{A} \in \{0,1\}^{n \times
n}$. 
% 
% For a weighted graph, we have $\mathbf{A} \in \mathbb{R}^{n \times n}$
%
$\mathbf{A}$ determines the connectivity of vertices: $\mathbf{A}(i,j) = 1
\Leftrightarrow (i,j) \in E$.
The input, output, and hidden feature vector of a vertex~$i$ are denoted with,
respectively, $\mathbf{x}_i, \mathbf{y}_i, \mathbf{h}_i$. We have $\mathbf{x}_i
\in \mathbb{R}^k$ and $\mathbf{y}_i, \mathbf{h}_i \in \mathbb{R}^{O(k)}$, where 
$k$ is the number of input features.
These
vectors can be grouped in matrices, denoted respectively as $\mathbf{X},
\mathbf{Y}, \mathbf{H} \in \mathbb{R}^{n \times O(k)}$.
If needed, we use the iteration index $(l)$ to denote the latent features in an
iteration (GNN layer)~$l$ ($\mathbf{h}^{(l)}_i$, $\mathbf{H}^{(l)}$).
Sometimes, for clarity of equations, we omit the index~$(l)$.
%
% A GNN computation may involve multiple GNN layers. Each layer produces
% a different version of the hidden feature vectors. Whenever necessary,
% we indicate this version with index~$(l)$, i.e., $\mathbf{h}^{(l)}_i$
% and $\mathbf{H}^{(l)}$.

\if 0
%
Depending on the selected setting and downstream ML tasks, different parts of
the input can be treated as data samples in training/inference (vertices,
edges, graph components) and processed in parallel. Depending on whether there
are dependencies between these samples, one may have data parallelism (if there
are no dependencies between samples, e.g., if all samples are disconnected
components) or partition parallelism (if there are explicit dependencies, e.g.,
if samples are connected with edges).
%
\fi

\if 0
%
Then, a GNN model uses (1) and (2) to generate the output feature
matrix~$\mathbf{Y}$. 
%
Often, in this process, an intermediate matrix~$\mathbf{H}$ of hidden features
(often called the latent) is created.
%
The output matrix~$\mathbf{Y}$ is then used for the downstream ML tasks such as
node classification.
%
Most often, $\mathbf{X}$ and $\mathbf{Y}$ group feature vectors of vertices,
but one can also define them for grouping feature vectors of edges. In the
following, we focus on the former scenario, as it is prevalent in the
literature.
%
\fi

\subsection{GNN Mathematical Models}
\label{sec:gnns-math}

A GNN model defines a mathematical transformation that takes as input (1) the
graph structure~$\mathbf{A}$ and (2) the input features~$\mathbf{X}$, and
generates the output feature matrix~$\mathbf{Y}$. Unless specified otherwise,
$\mathbf{X}$ models vertex features.
The exact way of constructing $\mathbf{Y}$ based on $\mathbf{A}$ and
$\mathbf{X}$ is an area of intense research.
\iftr
Here, hundreds of different GNN models have been
developed~\cite{wu2020comprehensive, zhou2020graph, zhang2020deep,
chami2020machine, hamilton2017representation, bronstein2017geometric,
zhang2019graph}.
\else
Here, hundreds of different GNN models have been
developed~\cite{wu2020comprehensive, 
bronstein2017geometric}.
\fi
Importantly for parallel and distributed execution, \emph{one can formulate
most GNN models using either the \textbf{local formulation (LC)} based on
functions operating on single edges or vertices, or the \textbf{global
formulation (GL)}, based on operations on matrices grouping all vertex- and
edge-related vectors}.

\subsubsection{GNN Formulations: Local (LC) vs.~Global (GL)}

%\marginpar{\large\vspace{4em}\colorbox{yellow}{\textbf{R2}}\\\colorbox{yellow}{\textbf{1.1}},\\\colorbox{yellow}{\textbf{R2}}\\\colorbox{yellow}{\textbf{3.3}}}

{We explicitly distinguish LC and GL formulations because they have 
different potential for performance optimizations.
GL formulations can harness techniques from linear algebra and
matrix computations, such as communication
avoidance~\mbox{\cite{kwasniewski2019red, solomonik2014tradeoffs}}. They also
offer more potential for vectorization, as one operates on whole
feature and adjacency matrices and not on individual feature vectors.
LC formulations also have potential advantages. For example,
functions operating on single vertices/edges can be programmed more
effectively and scheduled more flexibly on low-end compute resources such as
serverless functions. Moreover, the fine-grained perspective
facilitates integration with vertex-centric graph processing paradigms,
benefiting from established parallel frameworks such as
Galois~\mbox{\cite{kulkarni2007optimistic}}.}

%\marginpar{\large\vspace{4em}\colorbox{yellow}{\textbf{R2}}\\\colorbox{yellow}{\textbf{1.2}}}

{
It is often highly non-trivial to provide both an LC and a GL variant of a GNN
model.  While some models have both formulations (e.g., GCN, GIN, Vanilla
Attention, CommNet; cf.~Table~\mbox{\ref{tab:models-fg-1}}), for most models,
this is not the case. Many models only have known local formulations (e.g.,
MoNet, GAT, AGNN, G-GCN, the ``pooling'' variant of GraphSAGE, EdgeConv
``choice 5''; cf.~Table~\mbox{\ref{tab:models-fg-1}}) or global ones (e.g.,
SGC, ChebNet, DCNN, GDC, LINE, PPNP; cf.~Table~\mbox{\ref{tab:models-la}}).
Very often, developing an LC variant of a GL model is hard, e.g., 
for the PPNP model, it would require finding the LC equivalent of inverting the
adjacency matrix. On the other hand, complex operations used to compute a score
for an edge in many LC formulations (e.g., in MoNet or G-GCN) are challenging to
express in GL formulations.
%
%\textcolor{red}{A recent work provides GL formulations for several A-GNNs~\cite{besta2023highgnns}.}
%
Hence, it is important to investigate both types of formulations to ensure all
these models can benefit from efficient parallel and distributed
execution.}

%\marginpar{\large\vspace{0em}\colorbox{yellow}{\textbf{R2}}\\\colorbox{yellow}{\textbf{1.2}}}

{Figure~\mbox{\ref{fig:gnn-models-cats}} shows the taxonomy of GNN
formulations. The LC sub-categories were proposed by Bronstein et
al.~\mbox{\cite{bronstein2021geometric}}; the GL sub-categories
are described by Chen et al.~\mbox{\cite{chen2021bridging}}.}
%
\if 0
%
Note that the LC and GL sub-categories differ.
%
This is because they have been developed separately (i.e., whenever a GNN model
is designed, it is usually only either LC or GL), and only recently has the
LC and GL dichotomy been explicitly targeted by researchers.
%
Moreover, while for some models LC and GL are merely two different
perspectives, many models are easily expressible with one formulation only, as
detailed in the above paragraph. We expect that future research will bring more
insights into the relationships between LC and GL. In this work, we analyze
them both in terms of parallel and distributed execution.
%
\fi

\begin{figure}[h]
\vspaceSQ{-1em}
\includegraphics[width=1.0\columnwidth]{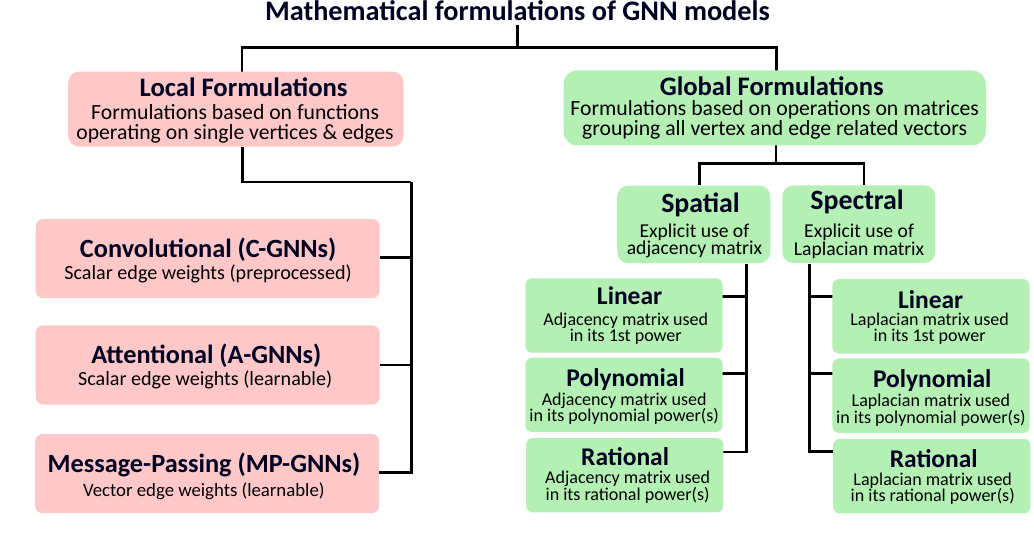}
\vspace{-2.0em}
\caption{\textbf{(\cref{sec:gnns-math}) Categories of GNN models.}
We classify the GNN model formulations into \textbf{local} and
\textbf{global}. Red/green refer to
formulation details in Figure~\ref{fig:operators}.
{}
}
\label{fig:gnn-models-cats}
\vspaceSQ{-1.5em}
\end{figure}

%\enlargeSQ

\subsubsection{Local GNN Formulations: Details}
\label{sec:local-gnns}

In many GNN models, the latent feature vector~$\mathbf{h}_i$ of a given
node~$i$ is obtained by applying a \emph{permutation invariant} aggregator
function~$\bigoplus$, such as sum or max, over the feature vectors of the
neighbors~$N(i)$ of~$i$ \hl{($N(i)$ is defined as the 1-hop
neighborhood)~\mbox{\cite{bronstein2021geometric}}}. Moreover, the feature
vector of each neighbor of~$i$ may additionally be transformed by a
function~$\psi$. Finally, the outcome of
$\bigoplus$ may be also transformed with another function~$\phi$. The sequence
of these three transformations forms one \emph{GNN layer}. 
We denote such a GNN model formulation (based on $\bigoplus, \psi, \phi$) as
\textbf{local (LC)}. 
Formally, the equation specifying the feature vector~$\mathbf{h}^{(l+1)}_i$ of
a vertex~$i$ in the next GNN layer $l+1$ is as follows:

\marginpar{\large\vspace{-8em}\colorbox{yellow}{\textbf{R2}}\\\colorbox{yellow}{\textbf{(3)}}}

\vspaceSQ{-1em}
\footnotesize
\begin{gather}
\mathbf{h}^{(l+1)}_i = \phi \left( \mathbf{h}^{(l)}_i, \bigoplus_{j \in N(i)} \psi\left(\mathbf{h}^{(l)}_i, \mathbf{h}^{(l)}_j \right) \right) \label{eq:mpgnn}
\end{gather}
\normalsize

\marginpar{\large\vspace{3em}\colorbox{yellow}{\textbf{R2}}\\\colorbox{yellow}{\textbf{(2)}}}

\hl{As an example, consider the seminal GCN model by Kipf and
Welling~\mbox{\cite{kipf2016semi}}. Here, $\bigoplus$ is a sum over $N(i) \cup
\{i\} \equiv \widehat{N}(i)$, $\psi$ acts on each neighbor~$j$'s feature vector
by multiplying it with a scalar~$1 / \sqrt{d_i d_j}$, and $\phi$ is a linear
projection with a trainable parameter matrix~$\mathbf{W}$ followed by $ReLU$.
Thus, the LC formulation is given by $\mathbf{h}_i^{(l+1)} = ReLU \fRB{
  \mathbf{W}^{(l)} \times \fRB{ \sum_{j \in \widehat{N}(i)} \frac{1}{\sqrt{d_i
  d_j}} \mathbf{h}_j^{(l)} }}$. Note that each iteration may have different
  projection matrices~$\mathbf{W}^{(l)}$.}

%\marginpar{\large\vspace{3em}\colorbox{yellow}{\textbf{R1}}\\\colorbox{yellow}{\textbf{(B)}}}

\marginpar{\large\vspace{3em}\colorbox{yellow}{\textbf{R2}}\\\colorbox{yellow}{\textbf{(1)}}}

Depending on the details of $\psi$, one can further distinguish three GNN
classes~\cite{bronstein2021geometric}: \emph{Convolutional GNNs} (C-GNNs),
\emph{Attentional GNNs} (A-GNNs), and \emph{Message-Passing GNNs} (MP-GNNs).
\hl{Example models from each class can be found in Table~\mbox{\ref{tab:models-fg-1}}.}
In short, in these three classes of models, {$\psi$ respectively applies --
as a weight on the features -- a fixed scalar coefficient (C-GNNs), a learnable
scalar coefficient (A-GNNs), or a learnable vector coefficient (MP-GNNs)}.

%\marginpar{\large\vspace{1em}\colorbox{yellow}{\textbf{R1}}\\\colorbox{yellow}{\textbf{(C)}},\\\colorbox{yellow}{\textbf{R2}}\\\colorbox{yellow}{\textbf{1.3}}}

{Importantly, these approaches form a hierarchy, i.e., C-GNNs
$\subseteq$ A-GNNs $\subseteq$ MP-GNNs~\mbox{\cite{bronstein2021geometric}}.
Specifically, A-GNNs can represent C-GNNs by implementing attention as a
look-up table $a(x_u, x_v) = c_{uv}$. Then, both C-GNNs and A-GNNs are special
cases of MP-GNNs: $\psi(x_u, x_v) = c_{uv} \psi(x_v)$ (for GNNs) and $\psi(x_u,
x_v) = a(x_u, x_v) \psi(x_v)$ for A-GNNs.}

%\marginpar{\large\vspace{0em}\colorbox{yellow}{\textbf{R1}}\\\colorbox{yellow}{\textbf{(C)}}}

{Note that we follow the taxonomy established by Bronstein et
al.~\mbox{\cite{bronstein2021geometric, petar-gnns}}, where MP-GNNs is a parent
class of C-GNNs, A-GNNs, but also more specialized message-passing model
classes such as MPNN by Gilmer et al.~\mbox{\cite{gilmer2017neural}} or Graph
Networks by Battaglia et al.~\mbox{\cite{battaglia2018relational}}.}

There are many ways in which one can parallelize GNNs in the LC formulation.
Here, the first-class citizens are ``fine-grained'' functions being evaluated
for vertices and edges. Thus, one could execute
 these functions in parallel over different vertices, edges, and graphs,
  parallelize a single function over the feature dimension or over the
  graph structure, pipeline a sequence of functions
  within a GNN layer or across GNN layers, or fuse parallel execution of functions.
We discuss all these aspects in the following sections.

\subsubsection{Global GNN Formulations: Details}
\label{sec:global-forms}

Many GNN models can also be formulated using operations on matrices
$\mathbf{X}$, $\mathbf{H}$, $\mathbf{A}$, and others. We will refer to this
approach as the \textbf{global (GL) linear algebraic} approach.

For example, the GL formulation of the GCN model is $\mathbf{H}^{(l+1)} =
ReLU(\mathbf{\widehat{A}} \mathbf{H}^{(l)} \mathbf{W}^{(l)})$. $\mathbf{\widehat{A}}$ is
the \emph{normalized adjacency matrix with self loops}~$\mathbf{\widetilde{A}}$
(cf.~Table~\ref{tab:symbols}): $\mathbf{\widehat{A}} =
\mathbf{\widetilde{D}}^{-\frac{1}{2}} \mathbf{\widetilde{A}}
\mathbf{\widetilde{D}}^{-\frac{1}{2}}$.
This normalization incorporates coefficients $1 / \sqrt{d_i d_j}$ shown in the
LC formulation above (the original GCN paper gives more
details about normalization).

Many GL models use higher powers of~$\mathbf{A}$ (or its normalizations).
Based on this criterion, GL models can be \emph{linear (L)} (if only the 1st power
of $\mathbf{A}$ is used), \emph{polynomial (P)} (if a polynomial power is used),
and \emph{rational (R)} (if a rational power is used)~\cite{chen2021bridging}.
This aspect impacts how to best parallelize a given model, as we illustrate in Section~\ref{sec:model-par}.
For example, the GCN model~\cite{kipf2016semi} is linear.

\iftr
Importantly, GNN computations involve both sparse and dense matrices. As the
performance patterns of operations on such matrices differ
vastly~\cite{kepner2016mathematical, kwasniewski2021parallel,
kwasniewski2021pebbles, kwasniewski2019red}, this comes with potential for
different parallelization routines.
We analyze this in more detail in Section~\ref{sec:model-par}.
\else
GNN computations involve both sparse and dense matrices,
which enbtail different performance patterns~\cite{kepner2016mathematical}. Hence, this comes with potential for
different parallelization routines.
We analyze this in more detail in Section~\ref{sec:model-par}.
\fi

\subsection{GNN Inference vs.~GNN Training}
\label{sec:gnn-inf-tr}

A series of GNN layers stacked one after another, as detailed in
Figure~\ref{fig:gnn-layer} and
in~\cref{sec:gnns-math}, constitutes GNN inference.
GNN training consists of three parts: forward pass, loss computation, and
backward pass.
The forward pass has the same structure as GNN inference.
For example, in classification, the loss~$\mathcal{L}$ is obtained as follows:
$\mathcal{L} = \frac{1}{|\mathcal{Y}|} \sum_{i \in \mathcal{Y}}
\text{loss}\fRB{\mathbf{y}_i, \mathbf{t}_i}$, where $\mathcal{Y}$ is a set of
all the labeled samples, $\mathbf{y}_i$ is the final prediction for sample~$i$,
and $\mathbf{t}_i$ is the ground-truth label for sample~$i$.
In practice, one often uses the cross-entropy loss~\cite{chiang2019cluster};
other functions may also be used~\cite{hamilton2020graph}.

Backpropagation outputs the gradients of all the trainable weights in the
model. A standard chain rule is used to obtain mathematical formulations for
respective GNN models. For example, the gradients for the first GCN layer,
assuming a total of two layers ($L=2$), are as
follows~\cite{thorpe2021dorylus}:

\vspaceSQ{-1em}
\small
$$
\grad_{\mathbf{W}^{(0)}} \mathcal{L} = \fRB{\mathbf{\widehat{A}} \mathbf{X}}^T \fRB{\sigma'\fRB{\mathbf{\widehat{A}} \mathbf{X} \mathbf{W}^{(0)}} \odot \mathbf{\widehat{A}}^T \text{loss}\fRB{\mathbf{Y} - \mathbf{T}} {\mathbf{W}^{(1)}}^T}
$$
\normalsize

\noindent
where $\mathbf{T}$ is a matrix grouping all the ground-truth vertex labels,
cf.~Table~\ref{tab:symbols} for other symbols. 
This equation reflects the forward propagation formula
(cf.~\cref{sec:global-forms}); the main difference is using transposed matrices
(because backward propagation involves propagating information in the reverse
direction on the input graph edges) and the derivative of the
non-linearity $\sigma'$.
\iftr

The structure of backward propagation depends on whether full-batch or
mini-batch training is used.
Parallelizing mini-batch training is more challenging due to the inter-sample
dependencies, we analyze it in Section~\ref{sec:data-par}.
\else
The structure of backward propagation depends on whether full-batch or
mini-batch training is used.
Parallelizing mini-batch training is more challenging due to the inter-sample
dependencies, see Section~\ref{sec:data-par}.

\fi

\ifcnf
\begin{figure}[t]
%\vspaceSQ{-1em}
\centering
\includegraphics[width=1.0\columnwidth]{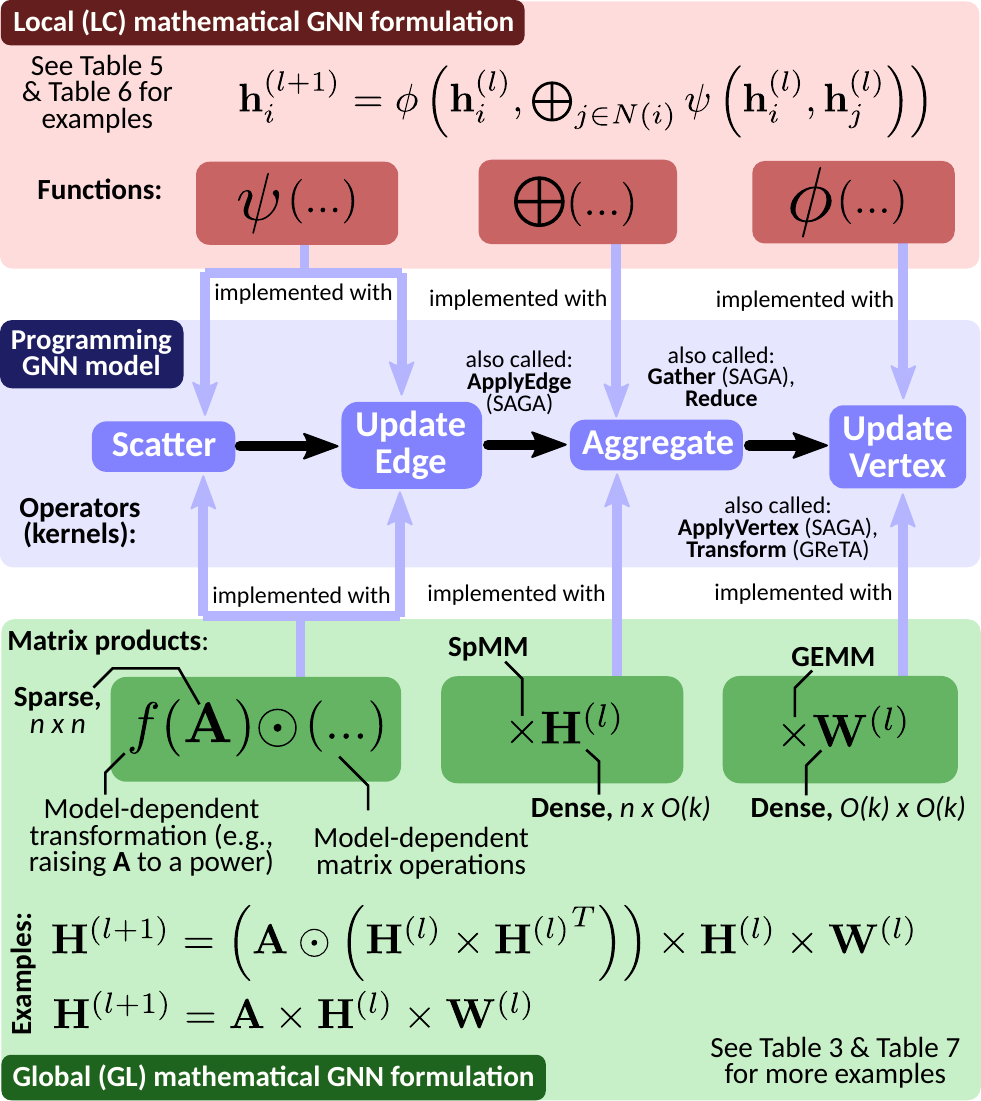}
\vspaceSQ{-2.0em}
\caption{\textbf{(\cref{sec:gnns-math}--\cref{sec:gnns-prog}) 
GNN model formulations} (top part: the local (LC) approach, bottom part:
the global (GL) approach), and \textbf{how they translate into GNN operators}
(central part). SAGA~\cite{ma2019neugraph}, NAU~\cite{wang2021flexgraph}, and
GReTA~\cite{kiningham2020greta} are GNN programming models. Red/green indicate
formulations from Figure~\ref{fig:gnn-models-cats}.}
\vspaceSQ{-1.0em}
\label{fig:operators}
\end{figure}

\fi

\subsection{GNN Programming Models and Operators}
\label{sec:gnns-prog}

\ifcnf
\begin{figure}[b]
\vspaceSQ{-1.5em}
\centering
\includegraphics[width=0.8\columnwidth]{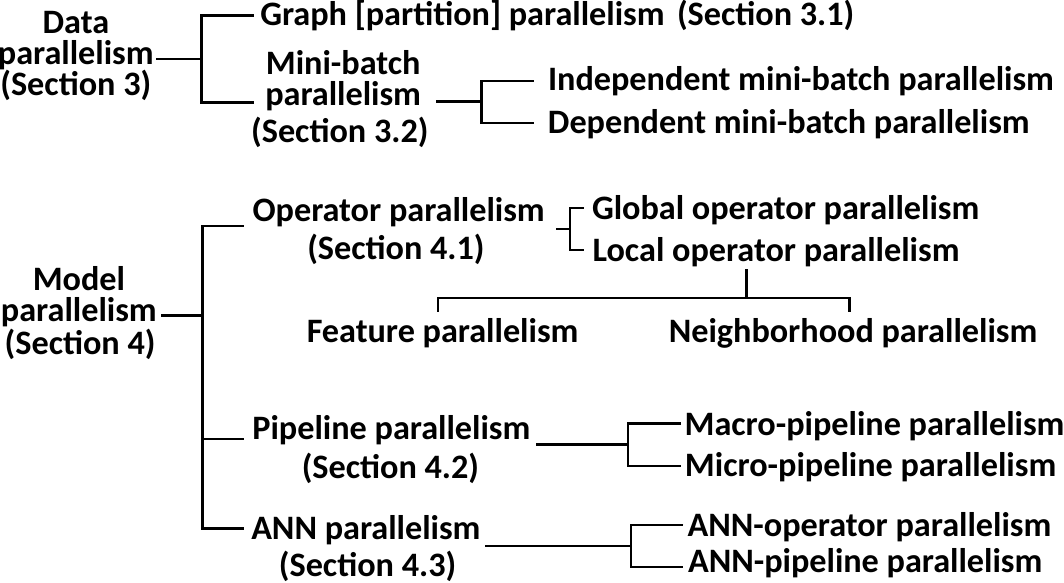}
\vspaceSQ{-0.5em}
\caption{\textbf{(\cref{sec:gnns-taxonomy}) Parallelism taxonomy in GNNs.}}
\label{fig:taxonomy-par}
\vspaceSQ{-1em}
\end{figure}

\begin{figure*}[t]
\vspaceSQ{-2em}
\includegraphics[width=1.0\textwidth]{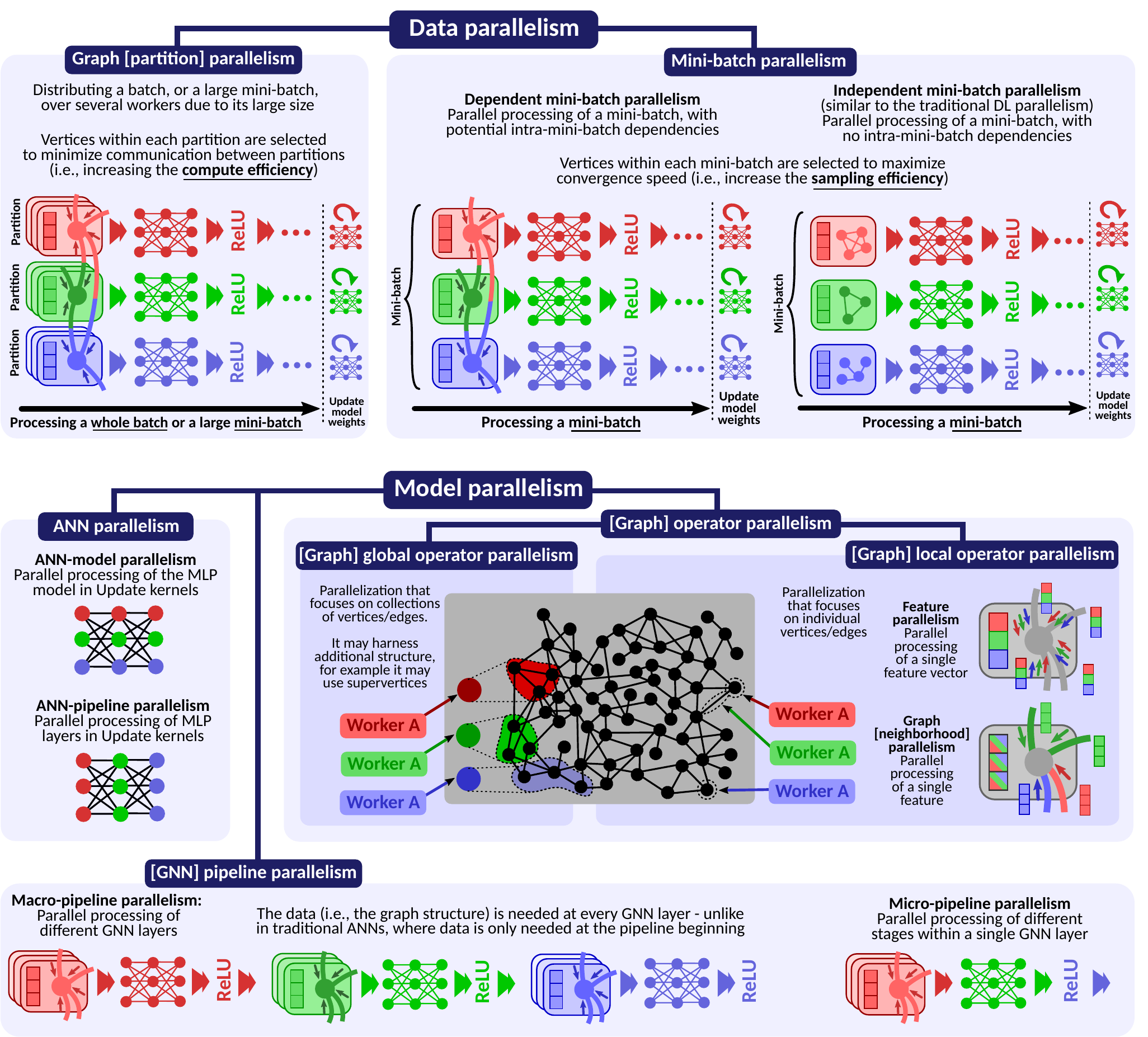}
\vspaceSQ{-1.5em}
\caption{\textbf{(\cref{sec:gnns-taxonomy}) Overview of parallelism in GNNs.
Different colors (\textcolor{red}{red}, \textcolor{green}{green},
\textcolor{blue}{blue}) correspond to different workers.}}
\label{fig:par-overview}
\vspaceSQ{-1em}
\end{figure*}
\fi

Recent works that originated in the systems community come with programming and
execution models.
These models facilitate GNN computations. In general, they each provide a set
of programmable \emph{kernels}, aka \emph{operators} (also referred to as UDFs
-- User Defined Functions) that enable implementing the GNN functions both in
the LC formulation ($\bigoplus, \psi, \phi$) and in the GL formulation (matrix
products and others). Figure~\ref{fig:operators} shows both LC and GL
formulations, and how they translate to the programming kernels.

The most widespread programming/execution model is SAGA~\cite{ma2019neugraph}
(``Scatter-ApplyEdge-Gather-ApplyVertex''), used in many GNN
libraries~\cite{zhang2020architectural}.
In the {Scatter} operator, the feature vectors of the vertices~$u,v$ adjacent
to a given edge~$(u,v)$ are processed (e.g., concatenated) to create the data
specific to the edge~$(u,v)$. Then, in {ApplyEdge}, this data is transformed
(e.g., passed through an MLP). Scatter and ApplyEdge together implement the
$\psi$ function.
Then, {Gather} aggregates the outputs of ApplyEdge for each vertex, using a
selected commutative and associative operation. This enables implementing the
$\bigoplus$ function. Finally, {ApplyVertex} conducts some user specified
operation on the aggregated vertex vectors (implementing $\phi$).

Note that, to express the edge related kernels Scatter and UpdateEdge, the LC
formulation provides a generic function~$\psi$. On the other hand, to express
these kernels in the GL formulation, one adds an element-wise product between
the adjacency matrix~$\mathbf{A}$ and some other matrix being a result of
matrix operations that provide the desired effect.  For example, to compute a
``vanilla attention'' model on graph edges, one uses a product of
$\mathbf{H}^{(l)}$ with itself transposed.

Other operators, proposed in GReTA~\cite{kiningham2020greta},
FlexGraph~\cite{wang2021flexgraph}, and others, are similar. For example, GReTA
has one additional operator, {Activate}, which enables a separate specification
of activation.  On the other hand, GReTA does not provide a kernel for applying
the $\psi$ function.

We illustrate the
relationships between operators and GNN functions from the LC and GL formulations, in
Figure~\ref{fig:operators}.
Here, we use the name \textbf{Aggregate} instead of
\textbf{Gather} to denote the kernel implementing the $\bigoplus$ function.
This is because ``Gather'' has traditionally been used to denote bringing
several objects together into an array~\cite{mpi3}\footnote{Another name
sometimes used in this context is ``Reduce''}.

\iftr
\begin{figure}[t]
%\vspaceSQ{-1em}
\centering
\includegraphics[width=1.0\columnwidth]{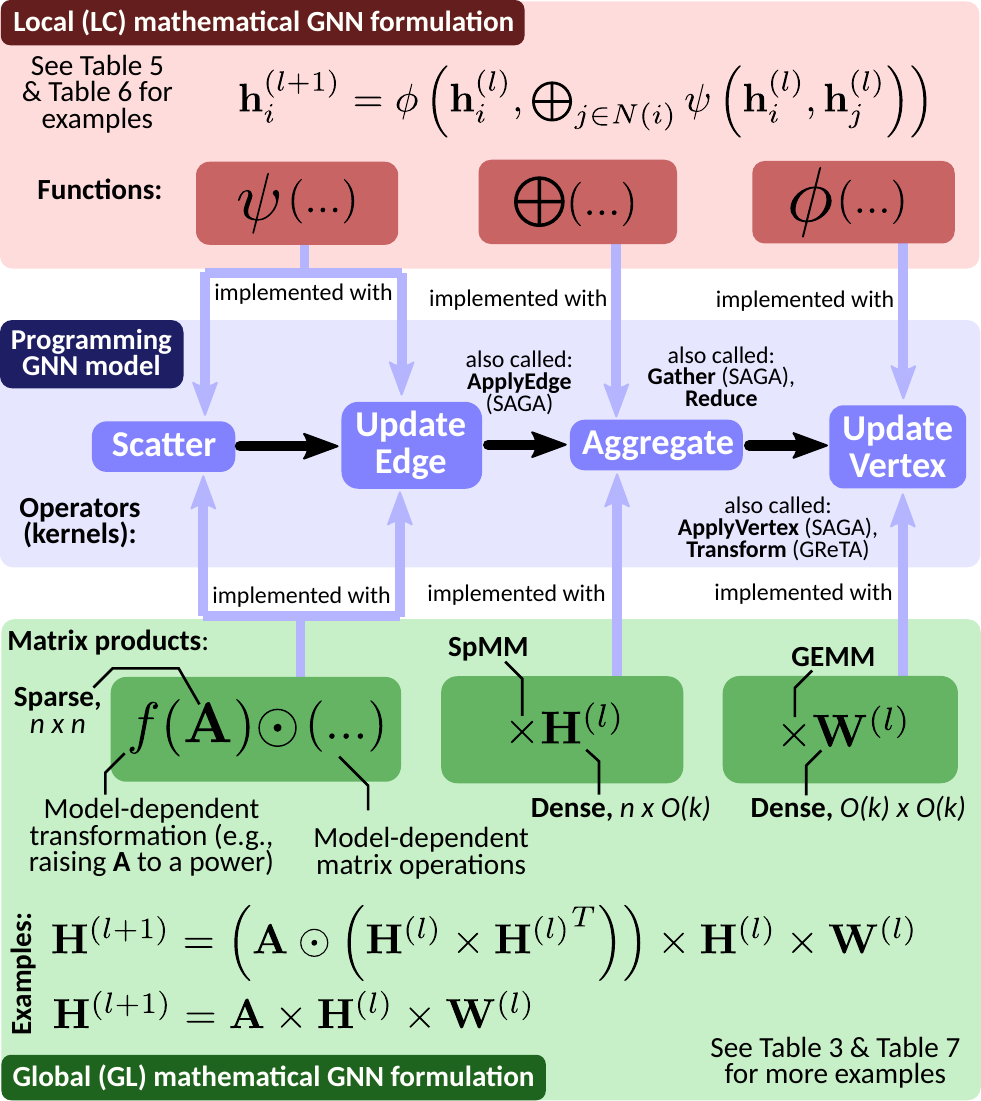}
\vspaceSQ{-2.0em}
\caption{\textbf{(\cref{sec:gnns-math}--\cref{sec:gnns-prog}) 
GNN model formulations} (top part: the local (LC) approach, bottom part:
the global (GL) approach), and \textbf{how they translate into GNN operators}
(central part). SAGA~\cite{ma2019neugraph}, NAU~\cite{wang2021flexgraph}, and
GReTA~\cite{kiningham2020greta} are GNN programming models. Red/green indicate
formulations from Figure~\ref{fig:gnn-models-cats}.}
\vspaceSQ{-1.0em}
\label{fig:operators}
\end{figure}

\fi

Parallelism in these programming and execution models is tightly related to
that of the associated GNN functions in LC and GL formulations; we discuss it
in Section~\ref{sec:model-par}. We also analyze parallel and distributed
frameworks and accelerators based on these models in Section~\ref{sec:systems}.

\subsection{Taxonomy of Parallelism in GNNs}
\label{sec:gnns-taxonomy}

{In traditional DL, there are two fundamental ways to parallelize the processing of
a neural network~\mbox{\cite{ben2019demystifying}}: \emph{data parallelism} and
\emph{model parallelism} where one partitions, respectively, data samples and
neural weights among different workers.}
% 
% Model parallelism can further be divided into
% pipeline parallelism (different NN layers are processed in parallel) and
% operator parallelism (a single sample or neural activity is processed in
% parallel).
%
Parallelism in GNNs also has {data parallelism} (detailed in
Section~\ref{sec:data-par}) and {model parallelism} (detailed in
Section~\ref{sec:model-par}). Yet, there are certain differences that we
identify and analyze. We overview the GNN parallelism taxonomy and the classes
of GNN parallelism in
Figure~\ref{fig:taxonomy-par} and
\ref{fig:par-overview}, respectively.

\iftr
\begin{figure}[b]
\includegraphics[width=1.0\columnwidth]{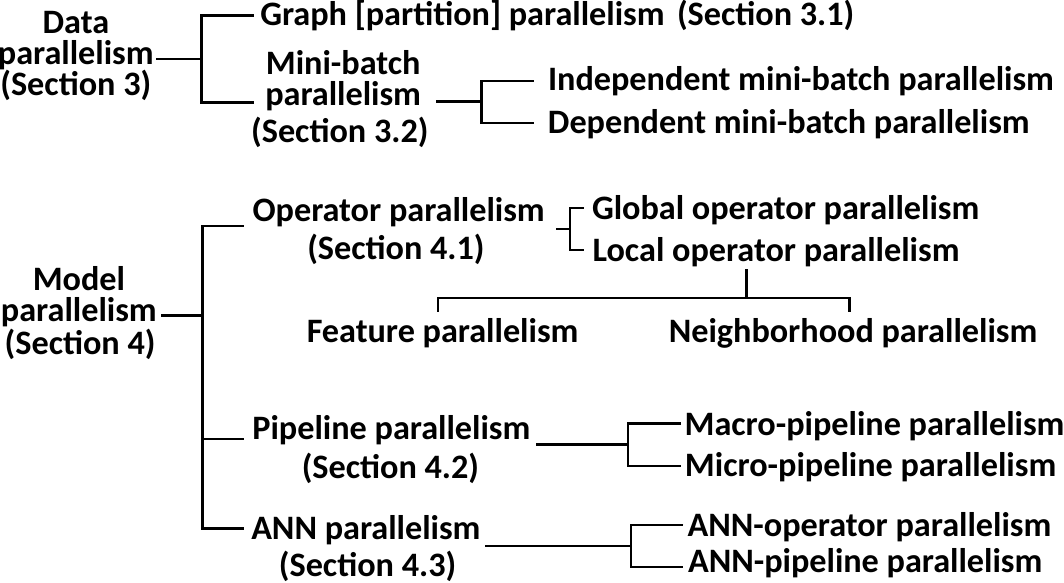}
\vspaceSQ{-1.5em}
\caption{\textbf{(\cref{sec:gnns-taxonomy}) Parallelism taxonomy in GNNs.}}
\label{fig:taxonomy-par}
\vspaceSQ{-1em}
\end{figure}

\begin{figure*}[h]
\includegraphics[width=1.0\textwidth]{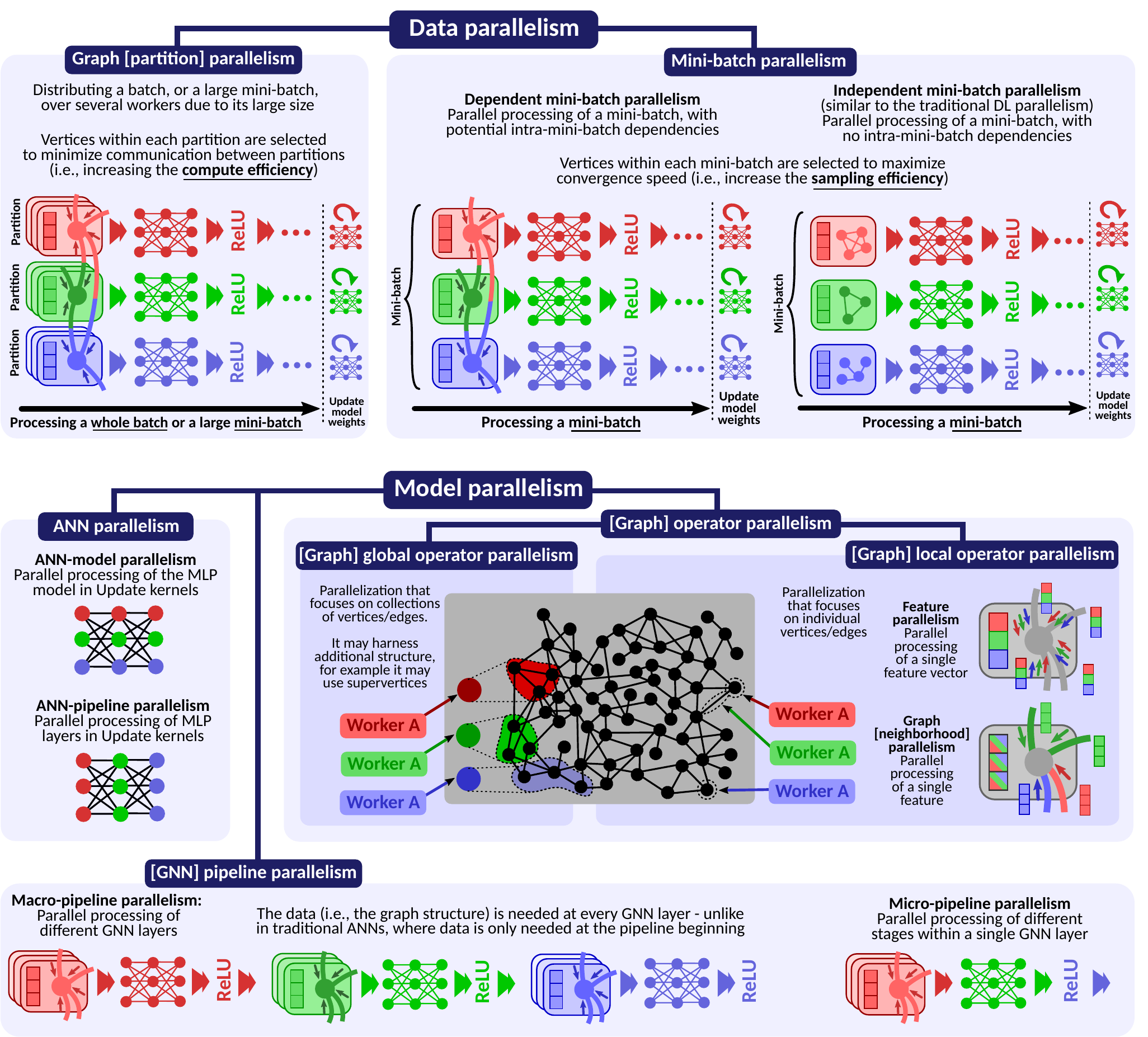}
\vspaceSQ{-1.5em}
\caption{\textbf{(\cref{sec:gnns-taxonomy}) Overview of parallelism in GNNs.
Different colors (\textcolor{red}{red}, \textcolor{green}{green},
\textcolor{blue}{blue}) correspond to different workers.}}
\label{fig:par-overview}
\vspaceSQ{-1em}
\end{figure*}
\fi

%\marginpar{\large\vspace{2em}\colorbox{yellow}{\textbf{R2}}\\\colorbox{yellow}{\textbf{(2.1)}}}

{The first form of data parallelism in GNNs
is
\emph{independent mini-batch parallelism}. Here, parallel workers process
a single mini-batch; the samples in this mini-batch have no
inter-sample dependencies (i.e., the samples are independent graphs, see Figure~\ref{fig:gnn-samples}).
This form of parallelism is analogous to the one in deep learning with images, where 
one parallelizes a mini-batch of pictures.
Second, GNNs also exhibit
\emph{dependent mini-batch parallelism}.
Here,
a mini-batch is also processed in parallel by multiple workers,
but the samples do have inter-sample dependencies (e.g., a mini-batch
could be a set of vertices and edges, sampled from a large input graph).
These dependencies make
parallelization much more complex, as we detail in Section~\ref{sec:data-par}.
Note that in GNNs, as in traditional DL, one can also use full-batch
training.
Finally, one can combine both mini-batch parallelism and full-batch processing
with \emph{graph [partition] parallelism}. Here, one distributes a given
mini-batch or a whole batch across different workers, {usually to fit it in
memory}.
%
% Different types of such dependencies are pictured in
% Figure~\ref{fig:gnn-samples}.
%
% Graph partition parallelism and dependent mini-batch parallelism are more
% challenging than their equivalent forms in traditional deep learning because
% of dependencies between data samples.
%
}

\marginpar{\large\vspace{0em}\colorbox{yellow}{\textbf{R2}}\\\colorbox{yellow}{\textbf{(4)}}}

{\hl{Model parallelism in GNNs can be divided into
\emph{operator parallelism}, \emph{artificial neural network (ANN)
parallelism}, and \emph{pipeline parallelism}.}
First, in operator parallelism, one parallelizes the Scatter and Reduce
kernels. Here, we further distinguish between \emph{[graph] local operator
parallelism} (parallel processing of individual vertices and edges) and
\emph{[graph] global operator parallelism} (parallel processing of collections of
vertices/edges).
Examples of local operator parallelism are \emph{feature parallelism}
(processing a feature vector of a given vertex in parallel) and \emph{graph
{[neighborhood]} parallelism} (processing in parallel the edges to the neighbors of a given
vertex).
Second, in \emph{ANN parallelism}, one parallelizes the UpdateEdge and
UpdateVertex kernels. These kernels can harness any form of parallelism that
has been developed for {traditional deep neural networks such as
MLPs}~\cite{ben2019demystifying}. Examples of ANN parallelism are
\emph{ANN-pipeline parallelism} (pipelining MLP layers) and \emph{ANN-operator
parallelism} (parallel processing of single NN operations).
Finally, in \emph{[GNN] pipeline parallelism}, one assigns different workers to
different stages of the GNN processing pipeline. Here, we distinguish
\emph{macro-pipeline parallelism} (pipelining the whole GNN layers) and
\emph{micro-pipeline parallelism} (pipelining the stages within
a single GNN layer).}

%\marginpar{\large\vspace{-12em}\colorbox{yellow}{\textbf{R2}}\\\colorbox{yellow}{\textbf{(3)}}}

%\marginpar{\large\vspace{-4em}\colorbox{yellow}{\textbf{R2}}\\\colorbox{yellow}{\textbf{(4)}}}

\begin{figure*}[t]
\centering
\includegraphics[width=1.0\textwidth]{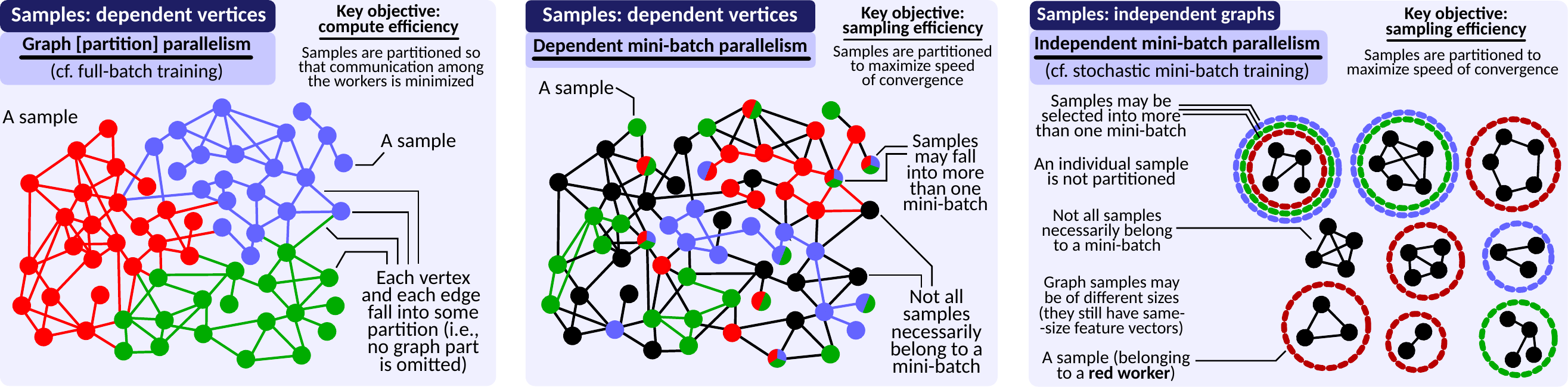}
\vspaceSQ{-1.5em}
\caption{\textbf{(\cref{sec:gp-par}, \cref{sec:mb-par}) Graph partition
parallelism vs.~dependent and independent mini-batch parallelism in GNNs.}
Different colors (\textcolor{red}{\textbf{red}}, \textcolor{green}{\textbf{green}},
\textcolor{blue}{\textbf{blue}}) indicate different graph partitions or mini-batches,
and the associated different workers. {Note that applying different colors to a vertex or to an independent graph
does not mean physical partitioning but it indicates that a given vertex or a given indepedent graph -- as a whole --
is used in more than a single mini-batch, indicated by the respective color}. Black vertices do not belong to any mini-batch.}
\label{fig:data-par-2}
\vspaceSQ{-1em}
\end{figure*}

\begin{figure*}[b]
\vspaceSQ{-1em}
\centering
\includegraphics[width=1.0\textwidth]{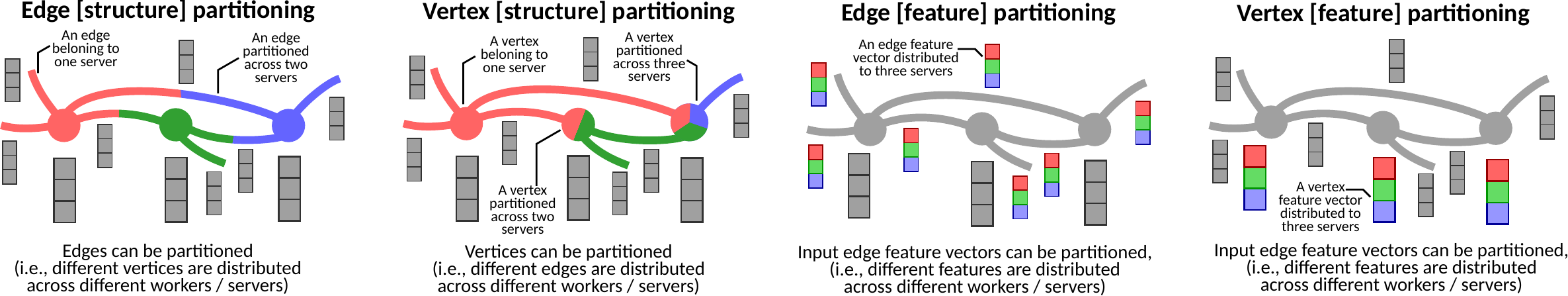}
\vspaceSQ{-1.5em}
\caption{\textbf{(\cref{sec:gp-par}) Different forms of graph partition
parallelism.} Different colors (\textcolor{red}{\textbf{red}}, \textcolor{green}{\textbf{green}},
\textcolor{blue}{\textbf{blue}}) indicate different graph partitions, and the associated
different workers.  The \textcolor{gray}{\textbf{gray}} graph element is oblivious to a
given form of partitioning. \textbf{Note that different partitioning schemes
can be combined together.}}
\label{fig:par-par}
\vspaceSQ{-1em}
\end{figure*}

\subsection{Parallel and Distributed Models and Algorithms}
\label{sec:par-algs}

%\marginpar{\large\vspace{-45em}\colorbox{yellow}{\textbf{R2}}\\\colorbox{yellow}{\textbf{(2)}}}

%\marginpar{\large\vspace{-27em}\colorbox{yellow}{\textbf{R2}}\\\colorbox{yellow}{\textbf{(3)}}}

We use formal models for reasoning about parallelism.
%
\if 0
%
For a \textbf{single-machine} (\textbf{shared-memory}), one often uses the
\emph{DAG model of dynamic multithreading}~\cite{blumofe1999scheduling,
blumofe1998space} as a \textbf{compute model}. Here, a parallel computation is
modeled as a \emph{directed acyclic graph} (DAG). A single DAG \emph{node}
models a single operation. Data used for this operation is modeled with
\emph{incoming edges} of the node. 
%
% The DAG model assumes constant-time operations, thus there are $O(1)$
% in-edges per node. 
%
The operation output is modeled with \emph{outgoing edges} of the node. A given
operation can be executed as soon as all its preceding operations are finished.
%
Then, to analyze the performance of parallel algorithms in the DAG model, we
use the \textbf{work-depth (WD) analysis}. The WD analysis enables bounding
run-times of parallel algorithms. The \emph{work}~$W$ of an algorithm is the
total number of nodes in the corresponding DAG, and the \emph{depth}~$D$ is the
longest directed path, and it forms a lower bound on the algorithm execution
time~\cite{Bilardi2011, blelloch2010parallel}.
%
One usually wants to minimize depth while preventing work from increasing too
much.
%
The WD analysis enables complexity analysis of parallel algorithms.  Executing
an algorithm using one processor and infinitely many processors takes $W$ and
$D$ time, respectively.
%
Moreover, any deterministic algorithm with work~$W$ and depth~$D$ can be
executed on $P$ processors in time~$T$ such that $\max\{W/P, D\} \le T \le W/P
+ D$~\cite{brent1974parallel}. 
%
\fi
%
For a \textbf{single-machine} (\textbf{shared-memory}), we use the {work-depth
(WD) analysis}, an established approach for bounding run-times of parallel
algorithms. The \emph{work} of an algorithm is the total number of operations
and the \emph{depth} is defined as the longest sequential chain of execution in
the algorithm (assuming infinite number of parallel threads executing the
algorithm), and it forms the lower bound on the algorithm execution
time~\cite{Bilardi2011, blelloch2010parallel}.
One usually wants to minimize depth while preventing work from increasing too
much.

\if 0
Following related work~\cite{whasenplaugh2014ordering,
dhulipala2018theoretically}, we assume that a parallel computation (modeled as
a DAG) runs on the \emph{ideal parallel computer} (\textbf{machine model}).
Each instruction executes in unit time and there is support for concurrent
reads, writes, and read-modify-write atomics (any number of such instructions
finish in $O(1)$ time).
%
These are standard assumptions used in all recent parallel graph coloring
algorithms~\cite{whasenplaugh2014ordering, dhulipala2018theoretically}.
\fi

In \textbf{multi-machine} (\textbf{distributed-memory}) settings, one is often
interested in understanding the algorithm cost in terms of the amount of
\emph{communication} (i.e., communicated data volume), \emph{synchronization}
(i.e., the number of global ``supersteps''), and
\emph{computation} (i.e., work), and minimizing these factors.
A popular model used in this setting is \emph{Bulk Synchronous Parallel
(BSP)}~\cite{valiant1990bridging}.

\section{Data Parallelism}
\label{sec:data-par}

In traditional deep learning, a basic form of data parallelism is to parallelize the processing
of input data samples within a mini-batch. Each worker processes its own portion of 
samples, computes partial updates of the model weights, and synchronizes these
updates with other workers using established strategies such as parameter
servers or allreduce~\cite{ben2019demystifying}. 
As samples (e.g., pictures) are independent, it is easy to parallelize their
processing, and synchronization is only required when updating the model
parameters.
%
% We show this in Figure~\ref{fig:data-par} (left). 
%
\if 0
%
Assume $n$ samples (such as images) in total, each with the dimensionality~$k$.
Then (assuming straightforward parallelization of processing one sample), work,
depth, and communication to obtain embeddings of all samples are $O\fRB{n W_s}$,
$O\fRB{D_s}$, and $O(1)$, respectively; $W_s$ and $D_s$ are work .
%
\fi
%
In GNNs, mini-batch parallelism is more complex because very often,
\emph{there are dependencies between data samples}
(cf.~Figure~\ref{fig:gnn-samples} and~\cref{sec:gnns-summary}. Moreover, the
input datasets as a whole are often massive. Thus, regardless of whether and
how mini-batching is used, one is often forced to resort to graph partition
parallelism because no single server can fit the dataset. We now detail both
forms of GNN data parallelism.
We illustrate them in Figure~\ref{fig:data-par-2}.

\subsection{Graph Partition Parallelism}
\label{sec:gp-par}

\iftr
Some graphs may have more than 250 billion vertices and beyond 10 trillion
edges~\cite{lin2018shentu, besta2021enabling}, and each vertex and/or edge may
have a large associated feature vector~\cite{hu2020open}. Thus, one inevitably
must distribute such graphs over different workers as they
do not fit into one server memory.
\else
Some graphs may have more than 250 billion vertices and beyond 10 trillion
edges~\cite{lin2018shentu}, and each vertex and/or edge may
have a large associated feature vector~\cite{hu2020open}. Thus, one inevitably
must distribute such graphs over different workers as they
do not fit into one server memory.
\fi
We refer to this form of GNN parallelism as the graph partition parallelism,
because it is rooted in the established problem of graph
partitioning~\cite{bulucc2016recent, karypis1995metis} and the associated
mincut problem~\cite{gianinazzi2018communication, karypis1995metis,
geissmann2018parallel}.
The main objective in graph partition parallelism is to distribute the graph
across workers in such a way that both communication between the workers and
work imbalance among workers are minimized.

We illustrate variants of graph partitioning in Figure~\ref{fig:par-par}.
When distributing a graph over different workers and servers, one can
specifically distribute vertices (\emph{edge [structure] partitioning}, i.e.,
edges are partitioned), edges (\emph{vertex [structure] partitioning}, i.e.,
vertices are partitioned), or edge and/or vertex input features
(\emph{edge/vertex [feature] partitioning}, i.e., edge and/or vertex input
feature vectors are partitioned). Importantly, these methods can be combined,
e.g., nothing prevents using both edge and feature vector partitioning
together.
Edge partitioning is probably the most widespread form of graph partitioning,
but it comes with large communication and work imbalance when partitioning
graphs with skewed degree distributions. Vertex partitioning alleviates it
to a certain degree, but if a high-degree vertex is distributed among many
workers, it also incurs overheads in maintaining a consistent
distributed vertex state.
\ifcnf
Differences between edge and vertex partitioning are covered in
rich literature~\cite{ccatalyurek2001fine, devine2006parallel,
gonzalez2012powergraph, bulucc2016recent, karypis1995metis}.
\else
Differences between edge and vertex partitioning are covered extensively in
rich literature~\cite{ccatalyurek2001fine, devine2006parallel,
gonzalez2012powergraph, bulucc2016recent, karypis1995metis, devine2006parallel,
buluc2013graph, bulucc2008towards, bader2013graph, kim2012sbv, ding2001min,
hendrickson1995improved, karypis1995analysis}.
\fi
Feature vertex partitioning was not addressed in the graph processing
area because in traditional distributed graph algorithms,
vertices and/or edges are usually associated with scalar values.

Partitioning entails communication when a given part of a graph depends on
another part kept on a different server.
This may happen during a graph related operator (Scatter, Aggregate) if edges
or vertices are partitioned, and during a neural network related operator
(UpdateEdge, UpdateVertex) if feature vectors are partitioned.

\marginpar{\large\vspace{1em}\colorbox{yellow}{\textbf{R2}}\\\colorbox{yellow}{\textbf{(6)}}}

\hl{Partition parallelism usually does not allow a single vertex to belong
to multiple partitions (unlike mini-batch parallelism, where a single
sample may belong to more than one mini-batch). However, there are strategies 
for reducing communication, in which vertices are cached on remote partitions.
Such schemes would involve maintaining multiple copies
of a given vertex on several partitions.}

\hl{Note that, while graph partition is usually conducted once, before training
starts, it could also be in principle reapplied during training, to
alleviate potential load imbalance (e.g., due to inserting new vertices or edges). Such
schemes are an interesting direction for future work.}

\marginpar{\large\vspace{-2em}\colorbox{yellow}{\textbf{R2}}\\\colorbox{yellow}{\textbf{(5)}}}

\subsubsection{Full-Batch Training}
\label{sec:fbt}

%\marginpar{\large\vspace{-15em}\colorbox{yellow}{\textbf{R1}}\\\colorbox{yellow}{\textbf{(D)}}}
%\marginpar{\large\vspace{-10em}\colorbox{yellow}{\textbf{R1}}\\\colorbox{yellow}{\textbf{(D)}}}

Graph partition parallelism is commonly used to alleviate large memory
requirements of full-batch training.
In full-batch training, one must store \emph{all the activations} for
\emph{each feature} in \emph{each vertex} in \emph{each GNN layer}).
Thus, a common approach for executing and parallelizing this scheme is using
distributed-memory large-scale clusters that can hold the massive input
datasets in their combined memories, together with graph partition parallelism.
Still, using such clusters may be expensive, and it still does not alleviate
the slow convergence.
Hence, mini-batching is often used.

\subsection{Mini-Batch Parallelism}
\label{sec:mb-par}

%\marginpar{\large\vspace{5em}\colorbox{yellow}{\textbf{R1}}\\\colorbox{yellow}{\textbf{(A)}}}

In GNNs, if data samples are independent graphs, then {mini-batch parallelism}
is similar to traditional deep learning. First, one mini-batch is a
set of such graph samples, with no dependencies between them. Second, samples
(e.g., molecules) may have different sizes. This may cause \textbf{load imbalance}),
similarly to, e.g., videos~\cite{li2020taming}. 
{For example, a single dataset (e.g., ChemInformatics)~\mbox{\cite{nr-sigkdd16}}
may contain graphs both 5 vertices and 18 edges as well as with 121 vertices and 298 edges.}
This setting is common in graph classification or
graph regression. We illustrate this in Figure~\ref{fig:data-par-2} (right),
and we refer to it as \emph{independent mini-batch parallelism}. Note that --
while graph samples may have different sizes (e.g., molecules can have
  different counts of atoms and bonds) -- their feature counts are
  the same. 

Still, in most GNN computations, mini-batch parallelism is much more challenging
because of inter-sample dependencies (\emph{dependent mini-batch parallelism}).
As a concrete example, consider node classification.
Similarly to graph partition parallelism, one may experience \textbf{load
imbalance} issues, e.g., because vertices may differ in their degrees.
\hl{Several works alleviate this~\mbox{\cite{su2021adaptive, mondal2022gnnie}}.}
%
% Now, assuming that the graph is connected (which is virtually always the
% case~\cite{hu2020open}), no matter how target vertices are selected to

\marginpar{\large\vspace{-1em}\colorbox{yellow}{\textbf{R2}}\\\colorbox{yellow}{\textbf{(7)}}}

\iftr
{While the graph prediction setting has been explored
in data science works~\mbox{\cite{wang2022molecular, gasteiger2021gemnet,
hu2020open, dwivedi2020benchmarking, wu2020comprehensive, xu2018powerful}}, it
has been largely unaddressed in system design
studies~\mbox{\cite{abadal2021computing}}.} 
\else
{While the graph prediction setting has been explored in data science
works~\mbox{\cite{hu2020open, dwivedi2020benchmarking, wu2020comprehensive}},
it has been largely unaddressed in system design
studies~\mbox{\cite{abadal2021computing}}.} 
\fi
{As such, to the best of our
knowledge, there are no detailed existing load balancing studies or schemes for
independent mini-batch parallelism (unlike for node or edge
predictions~\mbox{\cite{abadal2021computing}}).
Dependent mini-batch parallelism \emph{with graphs as samples} is even more
scarcely researched; for example, it does not even have a representing dataset
in the Large-Scale OGB challenge~\mbox{\cite{hu2020open}}.
Hence, we list these as research opportunities in
Section~\mbox{\ref{sec:challenges}}.}

%%\marginpar{\large\vspace{-3em}\colorbox{yellow}{\textbf{R1}}\\\colorbox{yellow}{\textbf{(A)}}}

Another key challenge in GNN mini-batching is the \textbf{information loss}
when selecting the \emph{target vertices} forming a given mini-batch. In
traditional deep learning, one picks samples randomly. In GNNs,
straightforwardly applying such a strategy would result in very low accuracy.
This is because, when selecting a random subset of nodes, this subset may not
even be connected, but most definitely it will be very sparse and due to the
missing edges, a lot of information about the graph structure is lost during
the Aggregate or Scatter operator. 
This information loss challenge was circumvented in the early GNN works with
{full-batch training}~\cite{kipf2016semi, wu2020comprehensive} (cf.~\cref{sec:fbt}).
Unfortunately, full-batch training comes with slow convergence (because the
model is updated only once per epoch, which may require processing billions of
vertices), and the above-mentioned large memory requirements.
Hence, two recent approaches that address specifically mini-batching proposed 
{\emph{sampling neighborhoods},} and 
\emph{appropriately selecting target vertices}.

\begin{figure*}[t]
\vspaceSQ{-1em}
\includegraphics[width=1.0\textwidth]{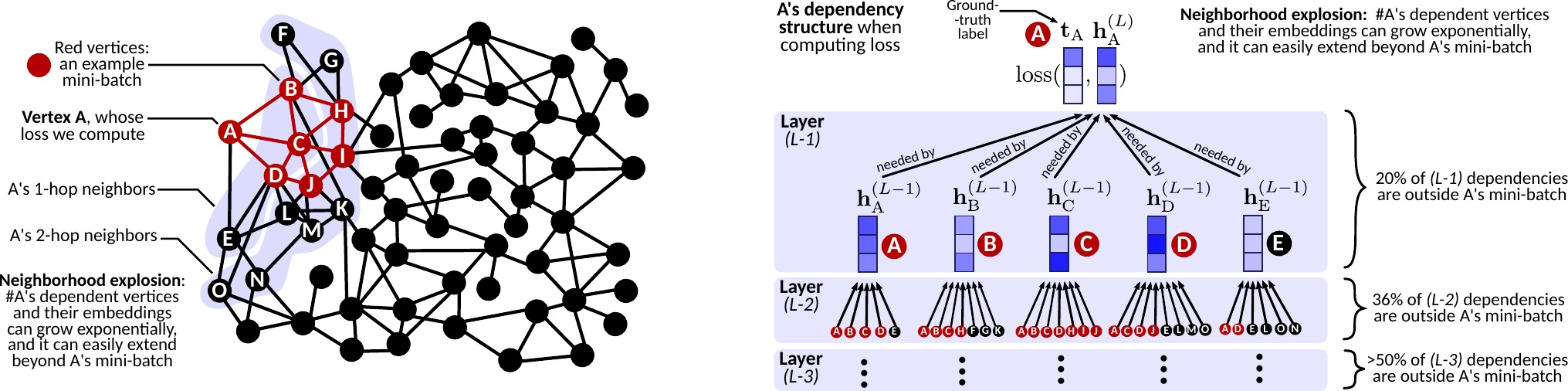}
\vspaceSQ{-1.5em}
	\caption{\textbf{(\cref{sec:mb-par}) Neighborhood explosion} in mini-batching in GNNs. \hl{This phenomenon is characteristic to schemes based on node-wise sampling~\mbox{\cite{zou2019layer}}.}}
\label{fig:neigh-exp}
\vspaceSQ{-1em}
\end{figure*}

%\marginpar{\large\vspace{1em}\colorbox{yellow}{\textbf{R1}}\\\colorbox{yellow}{\textbf{(E)}},\\\colorbox{yellow}{\textbf{(G)}}}

\subsubsection{Neighborhood Sampling}
\label{sec:sv}

In a line of works initiated by GraphSAGE~\cite{hamilton2017inductive}, one
{adds sampled neighbors of each selected target vertex~$v$ to
the mini-batch.} 
\iftr 
These neighbors are used to increase the accuracy
of predictions for target vertices (i.e., they are not used as target vertices
in that mini-batch). 
Specifically, when executing the Scatter and Aggregate kernels for each
of target vertices in a mini-batch, one also considers the pre-selected neighbors.
Hence, the results of Scatter and Aggregate are more accurate.
\fi
{Sampled neighbors of}~$v$ usually come
from not only 1-hop, but also from {$H$}-hop neighborhoods of $v$, 
where {$H$} may be as large as graph's diameter.
%
% Here, note that
% the target vertices within each mini-batch may be clustered but may also be
% spread across the graph (i.e., how exactly a mini-batch is selected depends on
% a specific scheme~\cite{li2021training, hamilton2017inductive,
% chen2017stochastic, chen2018fastgcn}). Support vertices, indicated with darker
% shades of each mini-batch color, are located up to 2 hops away from their
% target vertices. 
%
The exact selection of {sampled neighbors} depends on the
details of each scheme. In GraphSAGE, they are sampled (for each target
vertex) for each GNN layer before the actual
training.

\iftr

We illustrate {neighborhood sampling} in Figure~\ref{fig:sampling}. Here, note
that the target vertices within each mini-batch may be clustered but may also
be spread across the graph (depending on a specific
scheme~\cite{li2021training, hamilton2017inductive, chen2017stochastic,
chen2018fastgcn}). {Sampled neighbors}, indicated with darker shades of each
mini-batch color, are located up to 2 hops away from their target vertices.

\begin{figure}[h]
\includegraphics[width=1.0\columnwidth]{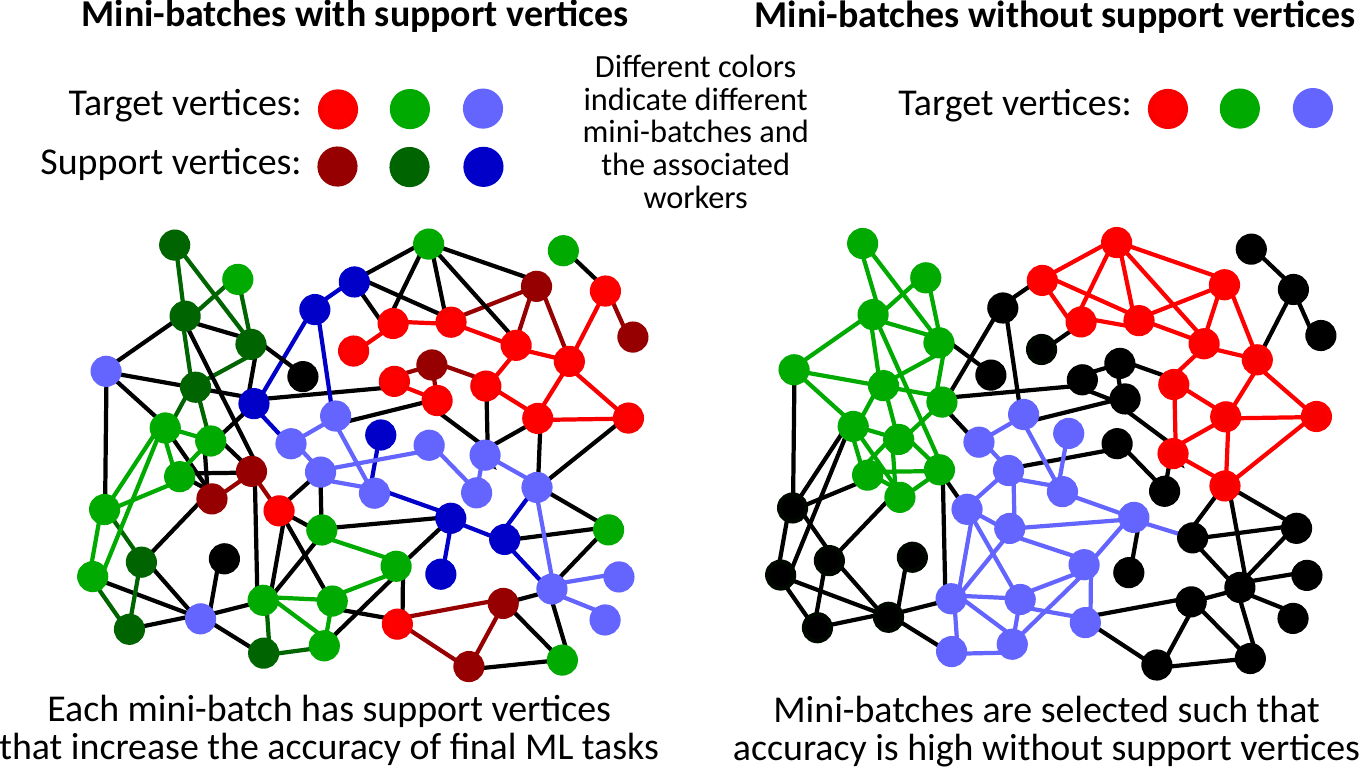}
\vspaceSQ{-1.5em}
\caption{\textbf{(\cref{sec:mb-par}) Two variants of mini-batching}: with and
w/o support vertices.  Light colors (\textcolor{red}{\textbf{red}},
\textcolor{green}{\textbf{green}}, \textcolor{blue}{\textbf{blue}}) indicate different
mini-batches (target vertices) and the associated workers. Dark
colors (\textcolor{darkred}{\textbf{dark red}}, \textcolor{darkgreen}{\textbf{dark green}},
\textcolor{darkblue}{\textbf{dark blue}}) indicate support vertices for each respective
mini-batch. \textbf{Black} vertices do not belong to any mini-batch.
Example schemes: GraphSAGE~\cite{hamilton2017inductive} (left), Cluster-GCN~\cite{chiang2019cluster} (right).}
\label{fig:sampling}
\vspaceSQ{-1em}
\end{figure}

\fi

%\marginpar{\large\vspace{16em}\colorbox{yellow}{\textbf{R1}}\\\colorbox{yellow}{\textbf{(F)}}}

One challenge related to {neighborhood sampling} is the \textbf{overhead of
their pre-selection}. For example, in GraphSAGE, one has to -- in addition to
the forward and backward propagation passes -- conduct as many sampling steps
as there are layers in a GNN, to {conduct sampling} for each layer and for
each target vertex. While this can be alleviated with parallelization schemes
also used for forward and backward propagation, it inherently increases the
depth of a GNN computation by a multiplicative constant factor.

Another associated challenge is called the \textbf{neighborhood explosion} and
is related to the memory overhead due to maintaining potentially many such
vertices. In the worst case, for each vertex in a mini-batch, assuming keeping
all its neighbors up to $H$ hops, one has to maintain $O(k d^H)$
state\footnote{\ssmall\sf The above bound is not tight because not all overlaps (e.g.,
between sampled neighbors of different target vertices) are considered.
However, we reflect the approach taken by all the considered GNN schemes
analyzed in Table~\mbox{\ref{tab:training}}.
To enhance the bound, one needs additional assumptions, e.g., on the graph
structure or its generating model.}.
Even if
some of these vertices are target vertices in that mini-batch and thus are
already maintained, when increasing $H$, their ratio becomes lower.
GraphSAGE alleviates this by {sampling a constant fraction of vertices from
each neighborhood} instead of keeping all the neighbors, but the memory
overhead may still be large~\cite{zeng2019graphsaint}. 
\iftr
Other works also explore efficient sampling to avoid neighborhood explosion.
For example, shaDow-GNN focuses on extracting a small number of critical
neighbors, while excluding irrelevant ones~\cite{zeng2021decoupling}.
\fi
We show an example
neighborhood explosion in Figure~\ref{fig:neigh-exp}.

\subsubsection{Appropriate Selection of Target Vertices}
\label{sec:as}

More recent GNN mini-batching works focus on the appropriate selection of
target nodes included in mini-batches, such that support vertices are not
needed for high accuracy. For example, Cluster-GCN first clusters a graph and
then assigns clusters to mini-batches~\cite{chiang2019cluster,
ying2018graph}. This way, one reduces the loss of
information because a mini-batch usually contains a tightly knit community of
vertices. 
\iftr
We illustrate this in Figure~\ref{fig:sampling} (right).  
\fi
\iftr
However,
one has to additionally compute graph clustering as a form of preprocessing.
This can be parallelized with one of many established parallel clustering
routines~\cite{karypis1995metis, besta2021sisa, besta2021graphminesuite,
besta2020communication}.
\else
However,
one has to additionally compute graph clustering as a form of preprocessing.
This can be parallelized with one of many established parallel clustering
routines~\cite{karypis1995metis}.
\fi

\if 0
%
``GCNs training is different from the challenge of classical distributed DNN
training where (1) data samples are small yet the model is large (model
parallelism (Krizhevsky, 2014; Harlap et al., 2018)) and (2) data samples do
not have dependency (data parallelism (Li et al., 2020; 2018b;a)), both
violating the nature of GCNs.'' (BNS-GCN)
%
\fi

\if 0
%
``graph and the node features in memory. As a result, for example, an L-layer GCN
model has time complexity $O(Lnd^2)$ and memory complexity $O(Lnd +Ld^2)$ [7],
prohibitive even for modestly-sized graphs.''
%
\fi

\if 0
%
``In graphs, on the other hand, the fact that the nodes are inter-related via
edges creates statistical dependence between samples in the training set.
Moreover, because of the statistical dependence between nodes, sampling can
introduce bias — for instance, it can make some nodes or edges appear more
frequently than on others in the training set — and this ‘side-effect’ would
need proper handling. Last but not least, one has to guarantee that the sampled
subgraph maintains a meaningful structure that the GNN can exploit.''
%
\fi

%\subsubsection{Discussion}

\subsection{Graph Partitions vs.~Mini-Batch/Full-Batch Training}

%\marginpar{\large\vspace{1em}\colorbox{yellow}{\textbf{R2}}\\\colorbox{yellow}{\textbf{2.2}}}

{Graph partitions are used primarily to contain the graph fully in memory
(avoiding expensive disk accesses), while mini-batches are used to speed up
convergence.
The first fundamental difference between these two is the key objective when
splitting the graph dataset across workers. For graph partition parallelism,
one aims to maximize \textbf{compute efficiency}, i.e., minimize runtimes by
minimizing the amount of communication between workers. For the former, one
focuses on increasing \textbf{sampling efficiency}, i.e., creating
mini-batches in such a way that the convergence speed is maximized.
%
% Thus, the objectives when selecting vertices belonging to a partition or a
% mini-batch differ. For example, when constructing partitions, one would focus
% on minimizing communication volume, while when selecting mini-batches, the
% priority is accelerating convergence. 
%
With certain schemes, these objectives
could result in selecting similar vertices. For example, Cluster-GCN selects
dense clusters as mini-batches and such clusters could also be effective graph
partitions~\mbox{\cite{chiang2019cluster}}. However, this is not always the
case as other mini-batching schemes do not necessarily focus on dense
clusters~\mbox{\cite{liu2021sampling}}.}
{Another difference is that, while a single graph partition is
processed by a single worker (i.e., multiple workers process multiple partitions),
one mini-batch is processed by multiple workers.
We also note that one could consider the parallel processing of different
mini-batches. This would entail \textbf{asynchronous GNN training}, with model
updates being conducted asynchronously. Such a scheme could slow down
convergence, but would offer potential for more parallelism.}

{Commonly, one uses graph partition
parallelism with full-batch training~\cite{wan2022pipegcn, wan2022bns, hoang1050efficient,
thorpe2021dorylus}. However, in principle, graph partition and mini-batch
parallelism are orthogonal to each other, and could thus be used together.} For
example, a large mini-batch running on workers with not much memory could
utilize graph partition parallelism to avoid I/O.  Such an approach has also
been proposed in traditional DL~\mbox{\cite{stich2018local,
oyama2018accelerating}}.

%\marginpar{\large\vspace{1em}\colorbox{yellow}{\textbf{R2}}\\\colorbox{yellow}{\textbf{2.1}}}

\if 0
\maciej{??}
Another {key difference between graph partition parallelism} and
{mini-batch parallelism} is the {timing of updating model weights}.
It takes place after the whole batch (for the former) and after each mini-batch
(for the latter).
%
\fi
%
\iftr
Other differences are as follows.
First, the primary objective when partitioning a graph is to minimize
communication and work imbalance across workers. Contrarily, in mini-batching,
one aims at a selection of target vertices that maximizes accuracy.
Second, each vertex belongs to some partition(s), but not each vertex is
necessarily included in a mini-batch.
Third, while mini-batch parallelism has a variant
with no inter-sample dependencies, graph partition parallelism nearly always
deals with a connected graph and has to consider such dependencies.
\fi
%
\if 0
These forms of parallelism also have similarities.
%
While a vertex can be partitioned across different workers,
a vertex could also be included in more than one mini-batch.
\fi
%

\subsection{Work-Depth Analysis}
\label{sec:data-fb-vs-mb}

We analyze work/depth of different GNN training schemes that use full-batch
or mini-batch training, see Table~\ref{tab:training}.

\begin{table}[t]
%\vspace{-1em}
\centering
\setlength{\tabcolsep}{1pt}
%\renewcommand{\arraystretch}{1.8}
% \ifcnf
% \renewcommand{\arraystretch}{0.5}
% \else
% \renewcommand{\arraystretch}{1.2}
% \fi
\scriptsize
%\footnotesize
%\small
%\sf
%
\begin{tabular}{@{}lll@{}}
\toprule
\makecell[l]{\textbf{Method}} &
\multicolumn{2}{c}{\textbf{Work \& depth in one training iteration}} \\
\midrule
\multicolumn{3}{c}{Full-batch training schemes:} \\
\midrule
Full-batch~\cite{kipf2016semi}  & $O\fRB{Lmk + Lnk^2}$ & $O\fRB{L \log k + L \log d}$ \\
Weight-tying~\cite{li2021training}  &  $O\fRB{Lmk + Lnk^2}$ & $O\fRB{L \log k + L \log d}$ \\
RevGNN~\cite{li2021training}  & $O\fRB{Lmk + Lnk^2}$ & $O\fRB{L \log k + L \log d}$ \\
%WT-RevGNN~\cite{li2021training}  & $O\fRB{Lmk + Lnk^2}$ & $O\fRB{L \log k + L \log d}$ \\
%
\midrule
\multicolumn{3}{c}{Mini-batch training schemes:} \\
\midrule
GraphSAGE~\cite{hamilton2017inductive}  & $O\fRB{Lmk + Lnk^2 + c^L nk^2}$ & $O\fRB{L \log k + L \log c}$ \\
VR-GCN~\cite{chen2017stochastic}  & $O\fRB{Lmk + Lnk^2 + c^Lnk^2}$ & $O\fRB{L \log k + L \log c}$ \\
FastGCN~\cite{chen2018fastgcn}  & $O\fRB{Lmk + Lnk^2 + cLnk^2}$ & $O\fRB{L \log k + L \log c}$ \\
Cluster-GCN~\cite{chiang2019cluster}  & $O\fRB{W_{pre} + Lmk + Lnk^2}$ & $O\fRB{D_{pre} + L \log k + L \log d}$ \\
GraphSAINT~\cite{zeng2019graphsaint}  & $O\fRB{W_{pre} + Lmk + Lnk^2}$ & $O\fRB{D_{pre} + L \log k + L \log d}$ \\
%
% RevGNN + sampling~\cite{li2021training} & MB & $O\fRB{Lmk + Lnk^2}$ & $O\fRB{}$ \\
% WT-RevGNN + sampling~\cite{li2021training} & MB & $O\fRB{Lmk + Lnk^2}$ & $O\fRB{}$ \\
%
\bottomrule
\end{tabular}
\vspaceSQ{-1em}
\caption{\textbf{(\cref{sec:data-fb-vs-mb}) Work-depth analysis of GNN training methods.}
$c$ is the number of vertices sampled per neighborhood or per GNN layer. 
%
% ``\textbf{[F]}'': Full-batch.
% ``\textbf{[M]}'': Mini-batch.
}
\label{tab:training}
\vspaceSQ{-1em}
\end{table}

First, all methods have a common term in work being $O(Lmk + Lnk^2)$ that
equals the number of layers $L$ times the number of operations conducted in
each layer, which is $mk$ for sparse graph operations (Aggregate) and $nk^2$
for dense neural network operations (UpdateVertex).
This is the total work for full-batch methods.  Mini-batch schemes have
additional work terms. Schemes based on support vertices (GraphSAGE, VR-GCN,
FastGCN) have terms that reflect how they pick these vertices.  GraphSAGE and
VR-GCN have a particularly high term $O(c^L n k^2)$ due to the neighborhood
explosion ($c$ is the number of vertices sampled per neighborhood).
FastGCN alleviates the neighborhood explosion by sampling $c$ vertices per
whole layer, resulting in $O(cL n k^2)$ work.
Then, approaches that focus on appropriately selecting target vertices
(GraphSAINT, Cluster-GCN) do not have the work terms related to the
neighborhood explosion.  Instead, they have preprocessing terms indicated with
$W_{pre}$. Cluster-GCN's $W_{pre}$ depends on the selected clustering method,
which heavily depends on the input graph size ($n$, $m$).  GraphSAINT, on the
other hand, does stochastic mini-batch selection, the work of which does not necessarily
grow with $n$ or $m$. 

In terms of depth, all the full-batch schemes depend on the number of
layers~$L$. Then, in each layer, two bottleneck operations 
are the dense neural network operation (UpdateVertex, e.g., a
matrix-vector multiplication) and the sparse graph operation (Aggregate).  They
take $O(\log k)$ and $O(\log d)$ depth, respectively.
Mini-batch schemes are similar, with the main difference being the $O(\log c)$
instead of $O(\log d)$ term for the schemes based on support vertices.  This is
because Aggregate in these schemes is applied over $c$ sampled neighbors.
Moreover, in Cluster-GCN and GraphSAINT, the neighborhoods may have up to $d$
vertices, yielding the $O(\log d)$ term. They however have the
additional preprocessing depth term $D_{pre}$ that depends on the used sampling
or clustering scheme.

To summarize, full-batch and mini-batch GNN training schemes have similar
depth. Note that this is achieved using graph partition parallelism in
full-batch training methods, and mini-batch parallelism in mini-batching
schemes.
Contrarily, overall work in mini-batching may be larger due to the overheads from
support vertices, or additional preprocessing when selecting target vertices
using elaborate approaches.
However, mini-batching comes with faster convergence and usually lower memory
pressure.

\ifall

All aggregations $\bigoplus_{j \in N(i)}$ within a single layer can be executed
in parallel, as the only data being modified is specific to each vertex. Thus,
the total work for all aggregations in a layer is $O\left(n k
d\right)$ as there are $n$ vertices. The depth of aggregations is
$O\left(\log d\right)$.

\fi

\subsection{Tradeoff Between Parallelism \& Convergence}

\marginpar{\large\vspace{-12em}\colorbox{yellow}{\textbf{R2}}\\\colorbox{yellow}{\textbf{(8)}}}

The \emph{efficiency tradeoff} between the amount of parallelism
in a mini-batch and the convergence speed, controlled with the
mini-batch size, is an important part of parallel traditional
ANNs~\cite{ben2019demystifying}. In short, 
small mini-batches would accelerate convergence but may limit
parallelism while large mini-batches may slow down convergence
but would have more parallelism.
In GNNs, finding the ``right'' mini-batch size is much more
complex, because of the inter-sample dependencies. For example,
a large mini-batch consisting of vertices that are not even
connected, would result in very low accuracy. On the other hand,
if a mini-batch is small but it consists of tightly connected
  vertices that form a cluster, then the accuracy of the updates
  based on processing that mini-batch can be high~\cite{chiang2019cluster}.

\section{Model Parallelism}
\label{sec:model-par}

In traditional neural networks, models are often large. 
In GNNs, models~($\mathbf{W}$) are usually small and often fit into
the memory of a single machine.
However, numerous forms of model parallelism are heavily used to improve
throughput; we provided
an overview in~\cref{sec:gnns-taxonomy} and in Figure~\ref{fig:par-overview}.
%
\if 0
%
First, there is parallelism within a single operator (compute kernel),
\textbf{operator parallelism}, with its variants that we refer to as \textbf{feature
parallelism} (parallel processing of features) and \textbf{graph
parallelism} (parallel processing of the graph structure when computing a
single feature).
%
%\textbf{step parallelism} (parallelism across operators),
%
Other forms of model parallelism include \textbf{micro-pipeline parallelism}
(parallel pipelined execution of operators within one GNN layer), and
\textbf{pipeline parallelism} (parallel pipelined execution of GNN layers).
%
\fi

In the following model analysis, we often picture the used linear algebra
objects and operations. For clarity, we indicate their shapes, densities, and
dimensions, using small figures, see
Table~\ref{tab:la-ops} for a list.
Interestingly, GNN models in the LC formulations heavily use dense matrices and
vectors with dimensionalities dominated by $O(k)$, and the associated operations.
On the other hand, the GL formulations use both sparse and dense matrices of
different shapes (square, rectangular, vectors), and the used matrix
multiplications can be dense--dense (GEMM, GEMV), dense--sparse (SpMM), and
sparse--sparse (SpMSpM). Other operations are elementwise matrix products or
rational sparse matrix powers.
This rich diversity of operations immediately illustrates a huge
potential for parallel and distributed techniques to be used with different
classes of models.

\begin{table}[t]
\centering
\renewcommand{\arraystretch}{0.7}
\footnotesize
%\scriptsize
%\ssmall
%\sf
\begin{tabular}{lll@{}}
\toprule
\textbf{Symbol} & \textbf{Description} & \textbf{Used often in} \\
\midrule
\multicolumn{3}{c}{{Matrices and vectors}} \\
\midrule
\MdnKO, \MdnOK & \makecell[l]{\makecell[l]{Dense vectors, \textbf{dimensions}:\\ $O(k) \times 1$, $1 \times O(k)$}} & LC models \\
\MdnKK & \makecell[l]{Dense matrices, \textbf{dimensions}:\\ $O(k) \times O(k)$} & GL \& LC models \\
\MdnNK, \MdnKN & \makecell[l]{Dense matrices, \textbf{dimensions}:\\ $n \times O(k)$, $O(k) \times n$} & GL models \\
\vspace{0.5em}
\hspace{-0.28em}\MspNN & \makecell[l]{Sparse matrix, \textbf{dimensions}:\\ $n \times n$} &  GL models \\
\midrule
\multicolumn{3}{c}{{Matrix multiplications (dimensions as stated above)}} \\
\midrule
\MdnNK $\times$ \MdnKK & \makecell[l]{GEMM, dense tall matrix $\times$\\ dense square matrix} & GL models\\
\MdnKK $\times$ \MdnKK & \makecell[l]{GEMM, dense square matrix $\times$\\ dense square matrix} & GL models\\
\vspace{0.5em}
\hspace{-0.28em}\MdnNK $\times$ \MdnKN & \makecell[l]{GEMM, dense tall matrix $\times$\\ dense tall matrix} & GL models\\
\vspace{0.5em}
\hspace{-0.28em}\MdnKK $\times$ \MdnKO & \makecell[l]{GEMV, dense matrix $\times$\\ dense vector} & LC models\\
\vspace{0.5em}
\hspace{-0.28em}\MspNN $\times$ \MdnNK & \makecell[l]{SpMM, sparse matrix $\times$\\ dense matrix} & GL models\\
\midrule
\multicolumn{3}{c}{{Elementwise matrix products and other operations}} \\
\midrule
\vspace{0.25em}
%\hspace{-0.28em}\MspNN $\odot$ \MdnNK $\times$ $\MdnNK^T$ & \makecell[l]{Elementwise product of a sparse matrix\\and a dense product of two tall matrices} & GL models \\
\hspace{-0.28em}\MspNN $\odot$ $(...)$ & \makecell[l]{Elementwise product of a\\ sparse matrix and some object} & GL models \\
% & SpMV,  sparse matrix -- dense vector product\\
%\vspace{0.5em}
%\hspace{-0.28em}
\vspace{0.25em}
\hspace{-0.28em}$\MspNN^x, x \in \mathbb{N}$ & \makecell[l]{SpMSpM, sparse matrix $\times$\\ sparse matrix} & GL models\\
\vspace{0.75em}
\hspace{-0.28em}$\MspNN^x, x \in \mathbb{Z}$ & Rational sparse matrix power & GL models\\
% & SpMSpV, sparse matrix -- sparse vector product\\
\vspace{0.5em}
\hspace{-0.28em}\MdnOK $\cdot$ \MdnKO, \MdnKO $\odot$ \MdnKO & \makecell[l]{Vector dot product,\\ elementwise vector product} & LC models \\
\vspace{0.5em}
\hspace{-0.28em}$\MdnKO \big\Vert\ \MdnKO$, $\sum$ \MdnKO & \makecell[l]{Vector concatenation,\\ sum of $d$ vectors, $d \le n$} & LC models \\
% \vspace{0.5em}
% \hspace{-0.28em}\MdnKO $\odot$ \MdnKO & Elementwise product of two vectors \\
% $\sum$ \MdnKO & Sum of up to $d$ vectors, $d \le n$\\
%
\bottomrule
\end{tabular}
\vspaceSQ{-1em}
\caption{Important objects and operations from linear algebra used in GNNs. \textbf{We
assign these operations to specific GNN models in Tables~\ref{tab:models-fg-1},
\ref{tab:models-fg-2}, and~\ref{tab:models-la}}.}
\vspaceSQ{-1em}
\label{tab:la-ops}
%\vspace{-0.5em}
\end{table}

\subsection{Local Operator Parallelism}

{In local operator parallelism, one focuses on parallelizing executions of
Scatter and Gather on individual vertices or edges (i.e., local graph
elements).
We further structure our investigation by considering separately local operator
parallelism over LC and GL GNN model formulations.}

\subsubsection{Parallelism in LC Formulations of GNN Models}

\if 0
%
However, having multiple formulations makes it complex to conduct the
work-depth analysis. Specifically, one is forced to consider separately work
and depth of both the function~$\psi$ and then, in case of C-GNNs and A-GNNs,
also separately the work and depth of coefficients~$c_{ij}$ and $a(\cdot,
\cdot)$. For these reasons, we apply the most generic MP-GNN form, which
encompasses C-GNNs and A-GNNs, when analyzing all the considered models:

\vspaceSQ{-1em}
\begin{align}
%
\mathbf{h}^{(l+1)}_i &= \phi \left( \mathbf{h}^{(l)}_i, \bigoplus_{j \in N(i)} \psi\left(\mathbf{h}^{(l)}_i, \mathbf{h}^{(l)}_j \right) \right) \label{eq:gen-gnn} 
%
\end{align}

In the MP-GNN formulation, each edge~$(i,j)$ is associated with a
function~$\psi$ evaluated on the feature vectors of attached vertices:
$\psi\fRB{\mathbf{h}_i, \mathbf{h}_j}$. This enables a uniform work-depth
analysis: for each model, one only considers the work and depth of $\psi$ when
assessing the amount of parallelism required to compute an edge weight.

Note that, in the following analysis, we assume that $\bigoplus$ conducts
reduction over \emph{1-hop neighborhoods}. This assumption is prevalent in
existing GNN works. However, technically, it is also possible to execute
$\bigoplus$ over general $H$-hop neighborhoods. We discuss this briefly
in~\cref{sec:h-hop}.
%
\fi

We illustrate generic work and depth equations of LC GNN formulations in
Figure~\ref{fig:work-depth}. Overall, work is the sum of any preprocessing
costs~$W_{pre}$, post-processing costs~$W_{post}$, and work of a single GNN
layer~$W_l$ times the number of layers~$L$. In the considered generic
formulation in Eq.~(\ref{eq:mpgnn}), $W_l$ equals to work needed to evaluate
$\psi$ for each edge ($m W_\psi$), $\bigoplus$ for each vertex ($n
W_{\oplus}$), and $\phi$ for each vertex ($n W_\phi$).
Depth is analogous, with the main difference that the depth of a single GNN
layer is a plain sum of depths of computing $\psi$, $\bigoplus$, and $\phi$
(each function is evaluated in parallel for each vertex and edge, hence no
multiplication with $n$ or $m$).

\vspaceSQ{-1em}
\begin{figure}[h]
\footnotesize
\begin{align}
\text{Work (generic)} & = W_{pre} + L  W_{l} + W_{post} \nonumber\\
\text{Depth (generic)} & = D_{pre} + L  D_{l} + D_{post} \nonumber\\
\text{Work of Eq.~(\ref{eq:mpgnn})} & = W_{pre} + L \cdot \fRB{m  W_\psi + n  W_\oplus + n  W_\phi} + W_{post}\nonumber\\
\text{Depth of Eq.~(\ref{eq:mpgnn})} & = D_{pre} + L \cdot \fRB{D_\psi + D_\oplus + D_\phi} + D_{post}\nonumber
\end{align}
\vspaceSQ{-2em}
\caption{Generic equations for work and depth in GNN LC formulations.}
\label{fig:work-depth}
\vspaceSQ{-1em}
\end{figure}

We now analyze work and depth of many specific GNN models, by focusing on the
three functions forming these models: $\psi$, $\bigoplus$, and $\phi$. The
analysis outcomes are in Tables~\ref{tab:models-fg-1} and~\ref{tab:models-fg-2}. We
select the representative models based on a recent
survey~\cite{chen2021bridging}.
We also indicate whether a model belongs to the class of convolutional (C-GNN),
attentional (A-GNN), or message-passing (MP-GNN) models~\cite{petar-gnns}
(cf.~\cref{sec:local-gnns}).

\textbf{Analysis of $\psi$}
We show the analysis results in Table~\ref{tab:models-fg-1}. We provide
the formulation of $\psi$ for each model, and we also illustrate all algebraic
operations needed to obtain~$\psi$.
All C-GNN models have their $\psi$ determined during preprocessing. This
preprocessing corresponds to the adjacency matrix row normalization ($c_{ij} =
1/d_i$), the column normalization ($c_{ij} = 1/d_j$), or the symmetric
normalization ($c_{ij} = 1/\sqrt{d_i d_j}$)~\cite{wu2020comprehensive}.
In all these cases, their derivation takes $O(1)$ depth and $O(m)$ work.
Then, A-GNNs and MP-GNNs have much more complex formulations of~$\psi$ than
C-GNNs. Details depend on the model, but - importantly - nearly all the models
have $O(k^2)$ work and $O(\log k)$ depth. The most computationally intense
model, GAT, despite having its work equal to $O(dk^2)$, has also logarithmic
depth of $O(\log k + \log d)$.  This means that computing~$\psi$ in all the
considered models can be effectively parallelized.
As for the sparsity pattern and type of operations involved in
evaluating~$\psi$, most models use GEMV. All the considered A-GNN
models also use transposition of dense vectors. GAT also uses vector
concatenation and sum of up to $d$ vectors. Finally, one considered MP-GNN
model uses an elementwise MV product.
In general, each considered GNN model uses dense matrix and vector operations
to obtain $\psi$ for each of the associated edges.

%\marginpar{\large\vspace{3em}\colorbox{ly}{\textbf{R2}}\\\colorbox{ly}{\textbf{3.1}}}

\sethlcolor{ly}
{Note that, by default, $\psi$ corresponds to edge feature vectors that are
``transient'', i.e., they are computed on the fly and are not stored explicitly
(unlike vertex feature vectors).  However, in some cases, one may also want to
explicitly instantiate edge feature vectors. Such instantiation would be used
in, for example, edge classification or edge regression tasks. An example GNN
formulation that enables this is Graph Networks by Battaglia et
al.~\mbox{\cite{battaglia2018relational}}, also an LC formulation. The insights
about parallelism in such a formulation are not different than the ones
provided in this section; the main different is the additional memory overhead
of $O(mk)$ needed for storing all edge feature vectors.}
\sethlcolor{yellow}

\if 0
{The LC formulations as specified by Eq.~(\ref{eq:mpgnn})
(C-GNNs, A-GNNs, MP-GNNs) enable explicit instantiation of vertex feature
vectors. However, in some cases, one may also want to explicitly instantiate
edge feature vectors. Such instantiation would be used in, for example, edge
classification or edge regression tasks. An example GNN formulation that
enables this is Graph Networks by Battaglia et
al.~\cite{battaglia2018relational}, also an LC formulation. The insights about
parallelism in such a formulation are not different than the ones provided in
this section; the central difference lies in the fact that, in MP-GNNs, edge
feature vectors~$\psi$ are ``transient'' and used primarily as input for
obtaining vertex feature vectors.}
\fi

\textbf{Analysis of $\bigoplus$}
The aggregate operator $\bigoplus_{j \in N(i)}$ is almost always commutative and
associative (e.g., min, max, or plain sum~\cite{wang2019deep,
fey2019fast}). While it operates on vectors~$\mathbf{x}_j$ of
dimensionality~$k$, each dimension can be computed independently of others.
Thus, to compute $\bigoplus_{j \in N(i)}$, one needs $O(\log d_i)$ depth and
$O(k d_i)$ work, using established parallel tree reduction
algorithms~\cite{blelloch1996programming}. 
Hence, $\bigoplus$ is the bottleneck in depth 
in all the considered models.
This is because $d$ (maximum vertex degree) is usually much larger than $k$.
%
%
% In case of computing the average, the additional averaging operation can be
% done in parallel for each dimension and for each vertex, and it does not
% increase the above bounds.

\textbf{Analysis of $\phi$}
The analysis of $\phi$ is shown in Table~\ref{tab:models-fg-2}
(for the same models as in Table~\ref{tab:models-fg-1}). We show the total model work
and depth. 
All the models entail matrix-vector dense products and a sum of up to $d$ dense
vectors.
Depth is logarithmic. Work varies, being the highest
for GAT.
%
%
% Both properties also heavily depends on a specific
% model. Overall, $\phi$ fulfills two roles: (1) it determines how the previous
% version~$\mathbf{h}^{(l)}_i$ of a vector being processed~$\mathbf{h}^{(l+1)}_i$
% is updated and combined with the neighbors of vertex~$i$, and (2) it determines
% how the activation is conducted. 
%
% While task~(1) is model specific, task~(2) is almost always an element-wise
% simple operation such as ReLU or sigmoid~\cite{??}. Thus, the activation part
% takes $O(1)$ work and depth.

We also illustrate the operator parallelism in the LC formulation, focusing on
the GNN programming kernels, in the top part of Figure~\ref{fig:op-par}. We
provide the corresponding generic work-depth analysis in
Table~\ref{tab:ops-wd}.
The four programming kernels follow the work and depth of the corresponding
LC functions ($\psi, \oplus, \phi$).

\begin{table}[h]
%\vspace{-1em}
\centering
%\setlength{\tabcolsep}{1pt}
%\renewcommand{\arraystretch}{1.8}
% \ifcnf
% \renewcommand{\arraystretch}{0.5}
% \else
% \renewcommand{\arraystretch}{1.2}
% \fi
\scriptsize
%\footnotesize
%\small
%\sf
%
\begin{tabular}{@{}lllll@{}}
\toprule
\makecell[l]{\textbf{Kernel}} &
\makecell[l]{\textbf{Work}} &
\makecell[l]{\textbf{Depth}} & 
\makecell[l]{\textbf{Comm.}} & 
\makecell[l]{\textbf{Sync.}} \\
\midrule
%
\if 0
%
\multicolumn{5}{c}{Example specific scheme (GCN~\cite{kipf2016semi}):} \\
%
\midrule
%
Scatter ($\psi$) & $O(1)$ & $O(1)$ & $O(mk)$ & $O(1)$ \\
UpdateEdge ($\psi$) & $O(mk)$ & $O(1)$ & $O(1)$ & $O(1)$ \\
Aggregate ($\oplus$) & $O(mk)$ & $O(\log d)$ & $O(mk)$ & $O(1)$ \\
UpdateVertex ($\phi$) & $O(mk)$ & $O(1)$ & $O(1)$ & $O(1)$ \\
%
\midrule
%
\multicolumn{5}{c}{Generic equations:} \\
%
\midrule
%
\fi
%
Scatter ($\psi$) & $O(1)$ & $O(1)$ & $O(mk)$ & $O(1)$ \\
UpdateEdge ($\psi$) & $O(m W_{\psi})$ & $O(D_\psi)$ & $O(1)$ & $O(1)$ \\
Aggregate ($\oplus$) & $O(n W_\oplus)$ & $O(D_\oplus \log d)$ & $O(mk)$ & $O(1)$ \\
UpdateVertex ($\phi$) & $O(n W_\phi)$ & $O(D_\phi)$ & $O(1)$ & $O(1)$ \\
\bottomrule
\end{tabular}
\vspaceSQ{-1em}
\caption{\textbf{Work-depth analysis of GNN operators (kernels).}
%
% ``\textbf{[F]}'': Full-batch.
% ``\textbf{[M]}'': Mini-batch.
}
\label{tab:ops-wd}
\vspaceSQ{-1em}
\end{table}

\begin{table*}[hbtp]
%\vspace{-1em}
\centering
\setlength{\tabcolsep}{3pt}
%\renewcommand{\arraystretch}{1.8}
% \ifcnf
% \renewcommand{\arraystretch}{0.5}
% \else
% \renewcommand{\arraystretch}{1.2}
% \fi
\scriptsize
\footnotesize
%\small
%\sf
%
\begin{tabular}{@{}llllllll@{}}
\toprule
\makecell[l]{\textbf{Reference}} &
\makecell[l]{\textbf{Class}} &
%\makecell[l]{\textbf{Pw}} &
\makecell[l]{\textbf{Formulation for} $\psi\fRB{\mathbf{h}_i, \mathbf{h}_j}$} & 
\makecell[l]{\textbf{Dimensions \& density of one}\\ \textbf{execution of $\psi\fRB{\mathbf{h}_i, \mathbf{h}_j}$}} &
\makecell[l]{{\textbf{Pr?}}} &
\multicolumn{2}{l}{\makecell[l]{\textbf{Work \& depth of one}\\ \textbf{execution of $\psi\fRB{\mathbf{h}_i, \mathbf{h}_j}$}}} \\
%\makecell[c]{\textbf{Remarks}} \\
%
\midrule
GCN~\cite{kipf2016semi} & C-GNN  & $\frac{1}{\sqrt{d_i d_j}} \mathbf{h}_j$ & $c$ $\cdot$ \MdnKO & \faY & $O(k)$  & $O(1)$ \\
\makecell[l]{GraphSAGE~\cite{hamilton2017inductive}\\ (mean)} & C-GNN  & $\mathbf{h}_j$  & \MdnKO & \faY & $O(1)$ & $O(1)$ \\
\vspace{0.5em}
\hspace{-0.25em}\makecell[l]{GIN~\cite{xu2018powerful}} & C-GNN  & $\mathbf{h}_j$  & \MdnKO & \faY & $O(1)$ & $O(1)$ \\
\vspace{0.5em}
\hspace{-0.25em}\makecell[l]{CommNet~\cite{sukhbaatar2016learning}} & C-GNN  & $\mathbf{h}_j$  & \MdnKO & \faY & $O(1)$ & $O(1)$ \\
%
\if 0
\makecell[l]{GraphSAGE~\cite{hamilton2017inductive}\\ (mean)} & C-GNN  & $\mathbf{h}_j$  & $\mathbb{R}$ & \faY & $O(1)$ & $O(1)$ \\
%
% \makecell[l]{GraphSAGE~\cite{hamilton2017inductive}\\ (LSTM)} & C-GNN ??   & ?? & $\mathbb{R}$ & \faY &  \\
%
\hspace{-0.25em}GIN~\cite{xu2018powerful} & C-GNN  & $\mathbf{h}_j$ & $\mathbb{R}$ & \faY & $O(1)$ & $O(1)$ \\
%
\hspace{-0.25em}CommNet~\cite{sukhbaatar2016learning} & C-GNN  & $\mathbf{h}_j$  & $\mathbb{R}$ & \faY & $O(1)$ & $O(1)$ \\
\fi
%
% \midrule
%
\makecell[l]{Vanilla\\ attention~\cite{vaswani2017attention}} & A-GNN  & $\fRB{\mathbf{h}_i^T \cdot \mathbf{h}_j} \mathbf{h}_j$ & \bigg( \MdnOK $\cdot$ \MdnKO \bigg) $\cdot$ \MdnKO & \faN & $O \fRB{k}$ & $O(\log k)$ \\
%
%\vspace{0.5em} 
MoNet~\cite{monti2017geometric} & A-GNN  & $\text{exp}\fRB{-\frac{1}{2} \fRB{\mathbf{h}_j - \mathbf{w}_j}^T \mathbf{W}_j^{-1} \fRB{\mathbf{h}_j - \mathbf{w}_j}}$ & exp\bigg( \MdnOK $\times$ \MdnKK $\times$ \MdnKO \bigg) & \faTimes & $O\fRB{ k^2 }$ & $O(\log k )$ \\
%
%\vspace{1.0em}
GAT~\cite{velivckovic2017graph} & A-GNN  & $\frac{\text{exp}\left( \sigma \left( \mathbf{a}^T \cdot \left[ \mathbf{W} \mathbf{h}_i \big\Vert \mathbf{W} \mathbf{h}_j \right] \right) \right)}{\sum_{y \in \widehat{N}(i)} \text{exp}\left( \sigma \left( \mathbf{a}^T \cdot \left[ \mathbf{W} \mathbf{h}_i \big\Vert \mathbf{W} \mathbf{h}_y \right] \right) \right)} \mathbf{h}_j$ & $\frac{\text{exp}\bigg( \MdnOK \cdot \bigg[ \MdnKK \times \MdnKO \bigg\Vert\ \MdnKK \times \MdnKO \bigg] \bigg)}{\mathlarger{\sum} \text{exp}\bigg( \MdnOK \cdot \bigg[ \MdnKK \times \MdnKO \bigg\Vert\ \MdnKK \times \MdnKO \bigg] \bigg)}  \cdot \MdnKO$ & \faN & $O\fRB{ d k^2 }$ & $O\fRB{\log k + \log d}$ \\
\vspace{0.5em}
\hspace{-0.25em}\makecell[l]{Attention-based\\ GNNs~\cite{thekumparampil2018attention}} & A-GNN  & $w \frac{\mathbf{h}_i^T \cdot \mathbf{h}_j} {\left\lVert\mathbf{h}_i \right\rVert \left\lVert\mathbf{h}_j \right\rVert} \mathbf{h}_j$ & \bigg( \MdnOK $\cdot$ \MdnKO \bigg) $\cdot$ \MdnKO & \faN & $O(k)$ & $O(\log k)$ \\
%
% \midrule
%
%GG-NN~\cite{li2015gated} $ MP-GNN  & $$ \\
%
G-GCN~\cite{bresson2017residual} & MP-GNN  & $\sigma \left( \mathbf{W}_1 \mathbf{h}_i + \mathbf{W}_2 \mathbf{h}_j \right) \odot \mathbf{h}_j$ & \bigg( \MdnKK $\times$ \MdnKO \bigg) $\odot$ \MdnKO & \faN & $O( k^2 )$ & $O(\log k)$ \\
\makecell[l]{GraphSAGE~\cite{hamilton2017inductive}\\ (pooling)} & MP-GNN  & $\sigma\fRB{\mathbf{W} \mathbf{h}_j + \mathbf{w}}$ & \MdnKK $\times$ \MdnKO & \faTimes & $O( k^2 )$ & $O(\log k)$  \\
\makecell[l]{EdgeConv~\cite{wang2019dynamic}\\ ``choice 1''} & MP-GNN  & $\mathbf{W} \mathbf{h}_j$ & \MdnKK $\times$ \MdnKO & \faN & $O( k^2 )$ & $O(\log k)$  \\ 
\makecell[l]{EdgeConv~\cite{wang2019dynamic}\\ ``choice 5''} & MP-GNN  & $\sigma\fRB{\mathbf{W}_1 \fRB{\mathbf{h}_j - \mathbf{h}_i} + \mathbf{W}_2 \mathbf{h}_i}$ & \MdnKK $\times$ \MdnKO & \faN & $O( k^2 )$ & $O(\log k)$  \\ 
\bottomrule
\end{tabular}
\vspaceSQ{-1em}
\caption{\textbf{Comparison of local (LC) formulations of GNN models with respect to the inner function $\psi\fRB{\mathbf{h}_i, \mathbf{h}_j}$.}
For clarity and brevity of equations, when it is obvious, we omit the 
matrix multiplication symbol $\times$ and the indices of a given
iteration (GNN layer) number~$(l)$.
``\textbf{Class}'': class of a GNN model with respect to
the complexity of $\psi$, details
are in Section~\ref{sec:gnns}.
All models considered in this table feature aggregations over 1-hop neighbors (``Type~L'', details are in Section~\ref{sec:gnns}).
``\textbf{Dimensions \& density}'': dimensions and density of the most important tensors and tensor operations when computing~$\psi\fRB{\mathbf{h}_i, \mathbf{h}_j}$ in a given model.
``\textbf{Pr}'': can coefficients in $\psi$ be preprocessed (\faY),
or do they have to be learnt (\faN)?
%
% ``\textbf{D}'': dimensionality of a coefficient.
%
When listing the most important tensor operations, we focus on multiplications.
%
\if 0
``\faBatteryFull'': Support.
%
``\faBatteryHalf'': Partial / limited support.
%
``\faTimes'': No support.
%
``\noAnswer'': Unknown.
\fi
}
\label{tab:models-fg-1}
\vspaceSQ{-1.75em}
\end{table*}

\begin{table*}[hbtp]
%\vspace{-1em}
\centering
\setlength{\tabcolsep}{1.5pt}
\renewcommand{\arraystretch}{1.8}
% \ifcnf
% \renewcommand{\arraystretch}{0.5}
% \else
% \renewcommand{\arraystretch}{1.2}
% \fi
\scriptsize
\footnotesize
%\ssmall
%\sf
%
\begin{tabular}{@{}lllllll@{}}
\toprule
\makecell[l]{\textbf{Reference}} &
\makecell[l]{\textbf{Class}} &
%\makecell[l]{\textbf{Pw}} &
\makecell[l]{\textbf{Formulation of $\phi$ for $\mathbf{h}^{(l)}_i$};\\ $\psi\fRB{\mathbf{h}_i, \mathbf{h}_j}$ are stated in Table~\ref{tab:models-fg-1}} & 
\makecell[l]{\textbf{Dimensions \& density of}\\ \textbf{computing $\phi(\cdot)$, excluding $\psi(\cdot)$}} &
\multicolumn{2}{l}{\makecell[l]{\textbf{Work \& depth} (a whole training iteration\\ or inference, \textbf{including $\psi$ from Table~\ref{tab:models-fg-1}})}} \\ 
\midrule
GCN~\cite{kipf2016semi} & C-GNN  & ${ \mathbf{W} \times \fRB{\sum_{j \in \widehat{N}(i)} \psi\fRB{\mathbf{h}_j}}}$ & \MdnKK $\times {\sum}$ \MdnKO & $O( L m k +  L n k^2 )$ & $O(L\log d + L\log k)$ \\
\makecell[l]{GraphSAGE~\cite{hamilton2017inductive}\\ (mean)} & C-GNN  & ${ \mathbf{W} \times \fRB{\frac{1}{d_i} \cdot \fRB{\sum_{j \in \widehat{N}(i)} \psi\fRB{\mathbf{h}_j}}}}$ & \MdnKK $\times \sum$ \MdnKO & $O( L m k + L n k^2 )$ & $O(L\log d + L\log k)$  \\ 
%
% \makecell[l]{GraphSAGE~\cite{hamilton2017inductive}\\ (LSTM)} & C-GNN ??   &  \\
%
GIN~\cite{xu2018powerful} & C-GNN  &  $\text{MLP}\fRB{(1+\epsilon) \mathbf{h}_i + \sum_{j \in N(i)} \psi\fRB{\mathbf{h}_j}}$ & $\overbrace{\;\MdnKK \times ... \times \MdnKK}^{K \text{times}} \times \sum$ \MdnKO & $O(L m k + L K n k^2 )$ & $O(L \log d + L K \log k)$ \\
CommNet~\cite{sukhbaatar2016learning} & C-GNN  & $\mathbf{W}_1 \mathbf{h}_i + \mathbf{W}_2 \times \left( \sum_{j \in N^+(i)} \psi\fRB{\mathbf{h}_j} \right)$ & \MdnKK $\times \sum$ \MdnKO & $O(L m k + L n k^2 )$ & $O(L \log d + L \log k)$ \\ 
%
% \midrule
%
\makecell[l]{Vanilla\\ attention~\cite{vaswani2017attention}} & A-GNN  & $\mathbf{W} \times \left( \sum_{j \in \widehat{N}(i)} \psi\fRB{\mathbf{h}_i, \mathbf{h}_j} \right)$ & \MdnKK $\times \sum$ \MdnKO & $O( L m k + L n k^2 )$ & $O(L \log d + L \log k)$ \\
\makecell[l]{GAT~\cite{velivckovic2017graph}} & A-GNN  & $\mathbf{W} \times \left( \sum_{j \in \widehat{N}(i)} \psi\fRB{\mathbf{h}_i, \mathbf{h}_j} \right)$ & \MdnKK $\times \sum$ \MdnKO & $O( L m d k^2 + L n k^2)$ & $O(L \log d + L \log k)$ \\
\makecell[l]{Attention-based\\ GNNs~\cite{thekumparampil2018attention}} & A-GNN  & $\mathbf{W} \times \left( \sum_{j \in \widehat{N}(i)} \psi\fRB{\mathbf{h}_i, \mathbf{h}_j} \right)$ & \MdnKK $\times \sum$ \MdnKO & $O( L m k + L n k^2 )$ & $O(L \log d + L \log k)$ \\
%
% \vspace{0.5em} 
MoNet~\cite{monti2017geometric} & A-GNN  & $\mathbf{W} \times \left( \sum_{j \in \widehat{N}(i)} \psi\fRB{\mathbf{h}_j} \right)$ & \MdnKK $\times \sum$ \MdnKO & $O( L m k^2 + L n k^2 )$ & $O(L \log d + L \log k)$ \\
%
% \midrule
%
%GG-NN~\cite{li2015gated} $ MP-GNN  & $$ \\
%
G-GCN~\cite{bresson2017residual} & MP-GNN  & $\mathbf{W} \times \left( \sum_{j \in N^+(i)} \psi\fRB{\mathbf{h}_i, \mathbf{h}_j} \right)$ & \MdnKK $\times \sum$ \MdnKO & $O( L m k^2 + L n k^2 )$ & $O(L \log d + L \log k)$ \\
\makecell[l]{GraphSAGE~\cite{hamilton2017inductive}\\ (pooling)} & MP-GNN  & $\fRB{\mathbf{W} \times \fRB{ \mathbf{h}_i \Big\Vert \fRB{\max_{j \in N(i)} \psi\fRB{\mathbf{h}_i, \mathbf{h}_j}} }}$ & \MdnKK $\times$ \bigg( \MdnKO \bigg\Vert\ \bigg( \MdnKK $\times \sum$ \MdnKO \bigg) \bigg) & $O( L m k^2 + L n k^2 )$ & $O(L \log d + L \log k)$ \\
\makecell[l]{EdgeConv~\cite{wang2019dynamic}\\ ``choice 1''} & MP-GNN  & $\sum_{j \in N^+(i)} \psi\fRB{\mathbf{h}_j}$ & $\sum$ \MdnKO & $O( L m k^2 + L n k^2 )$ & $O(L \log d + L \log k)$ \\
\makecell[l]{EdgeConv~\cite{wang2019dynamic}\\ ``choice 5''} & MP-GNN  & $\max_{j \in N^+(i)} \psi\fRB{\mathbf{h}_i, \mathbf{h}_j}$ & $ \sum$ \MdnKO & $O( L m k^2 + L n k^2 )$ & $O(L \log d + L \log k)$ \\ 
\ifFORM
SGC~\cite{wu2019simplifying} & C-GNN  & $\mathbf{W} \times \left( \sum_{j \in N^s(i)} \psi\fRB{\mathbf{h}_j^{(0)}} \right)$ & \MdnKK $\times$ \MdnKO & $O( n^s + m k + k^s )$ & $O(\log k + \log d^s)$  \\ 
%
%\vspace{0.5em}
DeepWalk~\cite{perozzi2014deepwalk}, ChebNet~\cite{defferrard2016convolutional} & C-GNN  & $\mathbf{W} \times \left( \sum_{j \in \overline{\widehat{N}(i)}^s} \psi\fRB{\mathbf{h}_j^{(0)}} \right)$ & \MdnKK $\times$ \MdnKO & \\ 
DCNN~\cite{atwood2016diffusion}, GDC~\cite{klicpera2019diffusion} & C-GNN  & $\mathbf{W} \times \left( \sum_{j \in \overline{N(i)}^s} \psi\fRB{\mathbf{h}_j^{(0)}} \right)$ & \MdnKK $\times$ \MdnKO & \\ 
Node2Vec~\cite{grover2016node2vec} & C-GNN  & $\mathbf{W} \times \left( \sum_{j \in \overline{\widehat{N}(i)}^2} \psi\fRB{\mathbf{h}_j^{(0)}} \right)$ & \MdnKK $\times$ \MdnKO &  \\ 
LINE~\cite{lin2015learning}, SDNE~\cite{wang2016structural} & C-GNN  & $\mathbf{W} \times \left( \sum_{j \in \overline{N(i)}^2} \psi\fRB{\mathbf{h}_j^{(0)}} \right)$ & \MdnKK $\times$ \MdnKO & \\ 
\makecell[l]{Auto-Regress\\ \cite{zhou2004learning, zhu2003semi}} & & R \\
PPNP~\cite{ying2018graph, klicpera2018predict, bojchevski2020scaling} & & R \\
ARMA~\cite{bianchi2021graph} & & R \\
ParWalks~\cite{wu2012learning} & & R \\
RationalNet~\cite{chen2021bridging} & & R \\
\fi
\bottomrule
\end{tabular}
\vspaceSQ{-1em}
\caption{\textbf{Comparison of local (LC) formulation of GNN models with respect to the outer function $\phi$.}
For clarity and brevity of equations, when it is obvious, we omit the matrix
multiplication symbol $\times$ and the indices of a given iteration
number~$(l)$; we also omit activations from the formulations (these are
elementwise operations, not contributing to work or depth).
``\textbf{Class}'': class of a GNN model with respect to
the complexity of $\psi$, details
are in Section~\ref{sec:gnns}.
All models considered in this table feature aggregations over 1-hop neighbors (``Type~L'', details are in Section~\ref{sec:gnns}).
``\textbf{Dimensions \& density}'': dimensions and density of the most important tensors and tensor operations in a given model when computing $\mathbf{h}^{(l)}_i$.
When listing the most important tensor operations, we focus on multiplications.
%
%
\if 0
``\faBatteryFull'': Support.
%
``\faBatteryHalf'': Partial / limited support.
%
``\faTimes'': No support.
%
``\noAnswer'': Unknown.
\fi
}
\label{tab:models-fg-2}
\vspaceSQ{-1.75em}
\end{table*}

\begin{figure*}[hbtp]
\vspace{-1em}
\includegraphics[width=1.0\textwidth]{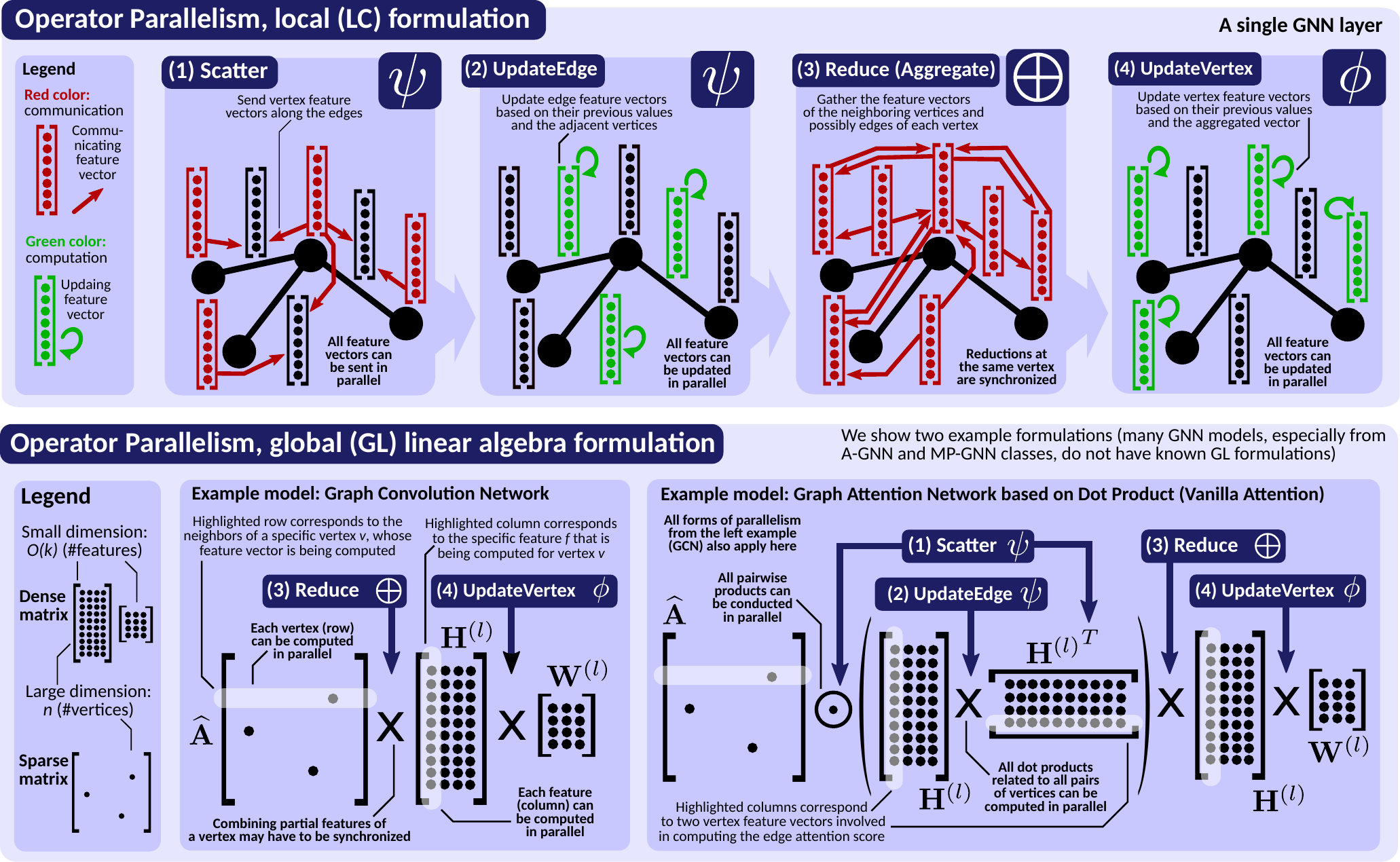}
\vspace{-2em}
\caption{\textbf{Operator parallelism in GNNs} for LC formulations (top) and GL formulations (bottom).}
\vspaceSQ{-1em}
\label{fig:op-par}
\end{figure*}

\begin{table*}[hbtp]
\vspace{-1em}
\centering
\setlength{\tabcolsep}{1.8pt}
\renewcommand{\arraystretch}{2.0}
% \ifcnf
% \renewcommand{\arraystretch}{0.5}
% \else
% \renewcommand{\arraystretch}{1.2}
% \fi
\scriptsize
\footnotesize
%\small
%\sf
%
\begin{tabular}{@{}llllllll@{}}
\toprule
\makecell[l]{\textbf{Reference}} &
\makecell[l]{\textbf{Type}} &
%\makecell[l]{\textbf{Pw}} &
\makecell[l]{\textbf{Algebraic formulation}\\ \textbf{ for} $\mathbf{H}^{(l+1)}$} & 
\makecell[l]{\textbf{Dimensions \& density}\\ \textbf{of deriving $\mathbf{H}^{(l+1)}$}} &
\makecell[c]{\textbf{\#I}} & 
\multicolumn{2}{l}{\makecell[l]{\textbf{Work \& depth} (one whole\\ training iteration or inference)}} \\ 
\midrule
GCN~\cite{kipf2016semi} & L & $\mathbf{\widehat{A}}  \mathbf{H}  \mathbf{W}$ & \MspNN $\times$ \MdnNK $\times$ \MdnKK & $L$ & $O(mkL + Lnk^2)$ & $O(L\log k + L\log d)$ \\
\makecell[l]{GraphSAGE~\cite{hamilton2017inductive}\\ (mean)} & L  & $\mathbf{\widehat{A}}  \mathbf{H}  \mathbf{W}$ & \MspNN $\times$ \MdnNK $\times$ \MdnKK & $L$ & $O(mkL + Lnk^2)$ & $O(L\log k + L\log d)$ \\ 
GIN~\cite{xu2018powerful} & L  & $\text{MLP}\fRB{((1 + \epsilon) \mathbf{I} + \mathbf{\widehat{A}})  \mathbf{H}}$ & \MspNN $\times$ \MdnNK $\times$ \MdnKK $\times$ ... $\times$ \MdnKK & $L$ & $O(mkL + KLnk^2)$ & $O(LK\log k + LK\log d)$ \\ 
CommNet~\cite{sukhbaatar2016learning} & L  & $ \mathbf{A}  \mathbf{H}  \mathbf{W}_2 + \mathbf{H}  \mathbf{W}_1$ & \MspNN $\times$ \MdnNK $\times$ \MdnKK + \MdnNK $\times$ \MdnKK & $L$ & $O(mkL + Lnk^2)$ & $O(L\log k + L\log d)$ \\ 
%
% \midrule
%
Dot Product~\cite{vaswani2017attention} & L  & $\fRB{ \mathbf{A} \odot \fRB{\mathbf{H}  {\mathbf{H}}^T}}  \mathbf{H}  \mathbf{W}$ & \MspNN $\odot$ $\bigg($ \MdnNK $\times$ $\MdnKN$ $\bigg)$ $\times$ \MdnNK $\times$ \MdnKK & $L$ & $O(L m k + Lnk^2)$ & $O(L\log k + L\log d)$ \\ 
%
% \makecell[l]{GraphSAGE~\cite{hamilton2017inductive}\\ (pooling)} & MP-GNN  & $\sigma\fRB{\mathbf{W} \mathbf{h}_j + \mathbf{w}}$ & $L$ \\ 
%
\makecell[l]{EdgeConv~\cite{wang2019dynamic}\\ ``choice 1''} & L  & $\mathbf{A}  \mathbf{H}  \mathbf{W}$ & \MspNN $\times$ \MdnNK $\times$ \MdnKK & $L$ & $O(mkL + Lnk^2)$ & $O(L\log k + L\log d)$\\ 
%
% \makecell[l]{EdgeConv~\cite{wang2019dynamic}\\ ``choice 5''} & MP-GNN  & $\sigma\fRB{\mathbf{W}_1  \fRB{\mathbf{h}_j - \mathbf{h}_i} + \mathbf{W}_2 \mathbf{h}_i}$ \\ 
%
%
SGC~\cite{wu2019simplifying} & P  & $\mathbf{\widehat{A}}^s  \mathbf{H}  \mathbf{W}$ & $\MspNN^s$ $\times$ \MdnNK $\times$ \MdnKK & 1 & $O(mn \log s + nk^2)$ & $O(\log k + \log s \log d)$ \\ 
%
% \vspace{0.5em}
DeepWalk~\cite{perozzi2014deepwalk} & P  & $\fRB{\sum_{s = 0}^T \mathbf{\overline{A}}^s}  \mathbf{H}  \mathbf{W}$ & $\bigg($ $\MspNN^0$ + ...+  $\MspNN^T$ $\bigg)$ $\times$ \MdnNK $\times$ \MdnKK & 1 & $O(mn \log T + nk^2)$ & $O(\log k + \log T \log d)$ \\ 
%
% \vspace{0.5em}
ChebNet~\cite{defferrard2016convolutional} & P  & $\fRB{ \sum_{s = 0}^T \theta_s \mathbf{\overline{A}}^s}  \mathbf{H}  \mathbf{W}$ & $\bigg($ $\MspNN^0$ + ...+  $\MspNN^T$ $\bigg)$ $\times$ \MdnNK $\times$ \MdnKK & 1 & $O(mn \log T + nk^2)$ & $O(\log k + \log T \log d)$  \\ 
%
% \vspace{0.5em}
\makecell[l]{DCNN~\cite{atwood2016diffusion},\\ GDC~\cite{klicpera2019diffusion}} & P  & $\fRB{\sum_{s = 1}^T w_s \mathbf{\overline{A}}^s}  \mathbf{H}  \mathbf{W}$ & $\bigg($ $\MspNN^1$ + ...+  $\MspNN^T$ $\bigg)$ $\times$ \MdnNK $\times$ \MdnKK & 1& $O(mn \log T + nk^2)$ & $O(\log k + \log T \log d)$  \\ 
%
% \vspace{0.5em}
% GDC~\cite{klicpera2019diffusion} & C-GNN  &  $\fRB{\sum_{s = 1}^T w_s \mathbf{\overline{A}}^s} \times \mathbf{H} \times \mathbf{W}$ &  & 1 \\ 
%
%SIGN~\cite{frasca2020sign} &  & \\
%
% \vspace{0.5em}
Node2Vec~\cite{grover2016node2vec} & P  & $\fRB{\frac{1}{p} \mathbf{I} + \fRB{1 - \frac{1}{q}} \mathbf{\overline{A}} + \frac{1}{q} \mathbf{\overline{A}}^2 }  \mathbf{H}  \mathbf{W}$ & $\bigg($ $\MspNN^0$ + ... + $\MspNN^2$ $\bigg)$ $\times$ \MdnNK $\times$ \MdnKK & 1 & $O(mn + nk^2)$ & $O(\log k + \log d)$  \\ 
%
% \vspace{0.5em}
\makecell[l]{LINE~\cite{lin2015learning},\\ SDNE~\cite{wang2016structural}} & P  & $\fRB{\mathbf{\overline{A}} + \theta \mathbf{\overline{A}}^2 }  \mathbf{H}  \mathbf{W}$ & $\bigg($ $\MspNN$ + $\MspNN^2$ $\bigg)$ $\times$ \MdnNK $\times$ \MdnKK & 1& $O(mn + nk^2)$ & $O(\log k + \log d)$  \\ 
%
% \vspace{0.5em}
% SDNE~\cite{wang2016structural} & C-GNN  &$\fRB{\mathbf{\overline{A}} + \theta \mathbf{\overline{A}}^2}  \mathbf{H}  \mathbf{W}$ &  & 1 \\ 
%
% \vspace{0.5em}
\makecell[l]{Auto-Regress\\ \cite{zhou2004learning, zhu2003semi}} & R & $\fRB{(1+\alpha)\mathbf{I} -\alpha \mathbf{\widehat{A}}}^{-1} \mathbf{H} \mathbf{W}$ & $\MspNN^{-1}$ $\times$ \MdnNK $\times$ \MdnKK & 1 & $O(n^3 + n k^2)$ & $O(\log^2 n + \log k)$ \\
\makecell[l]{PPNP\\ \cite{ying2018graph, klicpera2018predict, bojchevski2020scaling}} & R & $\alpha \fRB{\mathbf{I} -(1 - \alpha) \mathbf{\widehat{A}}}^{-1} \mathbf{H} \mathbf{W}$ & $\MspNN^{-1}$ $\times$ \MdnNK $\times$ \MdnKK & 1 & $O(n^3 + n k^2)$ & $O(\log^2 n + \log k)$ \\
\makecell[l]{ARMA~\cite{bianchi2021graph},\\ ParWalks~\cite{wu2012learning}} & R & $b \fRB{\mathbf{I} - a \mathbf{\widehat{A}}}^{-1} \mathbf{H} \mathbf{W}$ & $\MspNN^{-1}$ $\times$ \MdnNK $\times$ \MdnKK & 1 & $O(n^3 + n k^2)$ & $O(\log^2 n + \log k)$ \\
%
%
% RationalNet~\cite{chen2021bridging} & R \\
%
%
\bottomrule
\end{tabular}
\vspace{-1em}
\caption{\footnotesize\textbf{Comparison of global (GL) linear algebra formulations of GNN models.}
For clarity and brevity of equations, when it is obvious, we omit the matrix
multiplication symbol $\times$ and the indices of a given iteration
number~$(l)$; we also omit activations from the formulations (these are
elementwise operations, not contributing to work or depth).
``\textbf{Type}'': type of a GNN model with respect to
the scope of accessed vertex neighbors, details
are in Section~\ref{sec:gnns} (``\textbf{L}'': adjacency matrix is used in its 1st power, ``\textbf{P}'': adjacency matrix is used in its polynomial power, ``\textbf{R}'': adjacency matrix is used in its rational power).
``\textbf{\#I}'': the number of GNN layers (GNN iterations).
``\textbf{Dimensions \& density}'': dimensions and density of the most important tensors and tensor operations in a given model.
When listing the most important tensor operations, we focus on multiplications.
%
%
%
% ``\faBatteryFull'': Support.
%
% ``\faBatteryHalf'': Partial / limited support.
%
% ``\faTimes'': No support.
%
% ``\noAnswer'': Unknown.
}
\label{tab:models-la}
\vspace{-2.5em}
\end{table*}

%\marginpar{\large\vspace{3em}\colorbox{ly}{\textbf{R2}}\\\colorbox{ly}{\textbf{3.1}}}

\sethlcolor{ly}
\textbf{{Communication \& Synchronization}}
% \label{sec:comm-sync}
% \vspace{-0.25em}
%
{Communication in the LC formulations takes place in the Scatter kernel (a
part of~$\psi$) if vertex feature vectors are communicated to form edge feature
vectors; transferred data amounts to $O(mk)$. Similarly, during the Aggregate kernel
($\bigoplus$), there can also be $O(mk)$ data moved. Both UpdateEdge ($\psi$)
and UpdateVertex ($\phi$) do not explicitly move data. However, they may be
associated with communication intense operations; especially A-GNNs and MP-GNNs
often have complex processing associated with $\psi$ and $\phi$,
cf.~Tables~\mbox{\ref{tab:models-fg-1}} and~\mbox{\ref{tab:models-fg-2}}. While this
processing entails matrices of dimensions of up to $O(k) \times O(k)$, which
easily fit in the memory of a single machine, this may change in the future, \emph{if}
the feature dimensionality~$k$ is increased in future GNN models.}

\if 0
%
Synchronization in the LC formulations is $O(1)$ in all the kernels, except for
$\bigoplus$, where parallel binary tree reductions may require up to $O(\log
d)$ steps. 
%
\fi

{In the default synchronized variants of GNN, computing all kernels of the same
type must be followed by global synchronization, to ensure that all data
has been received by respective workers (after Scatter and Aggregate) or
that all feature vectors have been updated (after UpdateEdge and UpdateVertex).
In Section~\mbox{\ref{sec:async}}, we discuss how this requirement can be
relaxed by allowing asynchronous execution.}

\sethlcolor{yellow}

\subsubsection{Parallelism in GL Formulations of GNN Models}
\vspaceSQ{-0.25em}

Parallelism in GL formulations is analyzed in Table~\ref{tab:models-la}.
The models with both LC and FG formulations (e.g., GCN) have the same work and
depth. Thus, fundamentally, they offer the same amount of parallelism. 
However, the GL formulations based on matrix operations come with potential for
different parallelization approaches than the ones used for the LC
formulations.
\iftr
For example, there are more opportunities to use vectorization, because one is
not forced to vectorize the processing of feature vectors for each vertex or
edge separately (as in the LC formulation), but instead one could vectorize the
derivation of the whole matrix~$\mathbf{H}$~\cite{besta2017slimsell}.
\else
For example, there are more opportunities to use vectorization, because one is
not forced to vectorize the processing of feature vectors for each vertex or
edge separately (as in the LC formulation), but instead one could vectorize the
derivation of the whole matrix~$\mathbf{H}$.
\fi

There are also models designed in the GL formulations with no known LC
formulations, cf.~Tables~\ref{tab:models-fg-1}--\ref{tab:models-fg-2}.
These are models that use polynomial and rational powers of the adjacency
matrix, cf.~\cref{sec:global-forms} and Figure~\ref{fig:gnn-models-cats}.
These models have only one iteration.
They also offer parallelism, as indicated by the logarithmic depth (or square
logarithmic for rational models requiring inverting the adjacency
matrix~\cite{mulmuley1986fast}). While they have one iteration, making the $L$
term vanish, they require deriving a given power~$x$ of the adjacency
matrix~$\mathbf{A}$ (or its normalized version).  Importantly, as computing
these powers is not interleaved with non-linearities (as is the case with many
models that only use linear powers of~$\mathbf{A}$), the increase in work and
depth is \emph{only logarithmic}, indicating more parallelism. Still,
their representative power may be lower, due to the lack of non-linearities.

We overview two example GL models (GCN and vanilla graph
attention) in Figure~\ref{fig:op-par} (bottom). In this figure, we also
indicate how the LC GNN kernels are reflected in the flow of matrix operations
in the GL formulation.

\iftr
\begin{figure*}[b]
\vspace{-1em}
\centering
\includegraphics[width=1.0\textwidth]{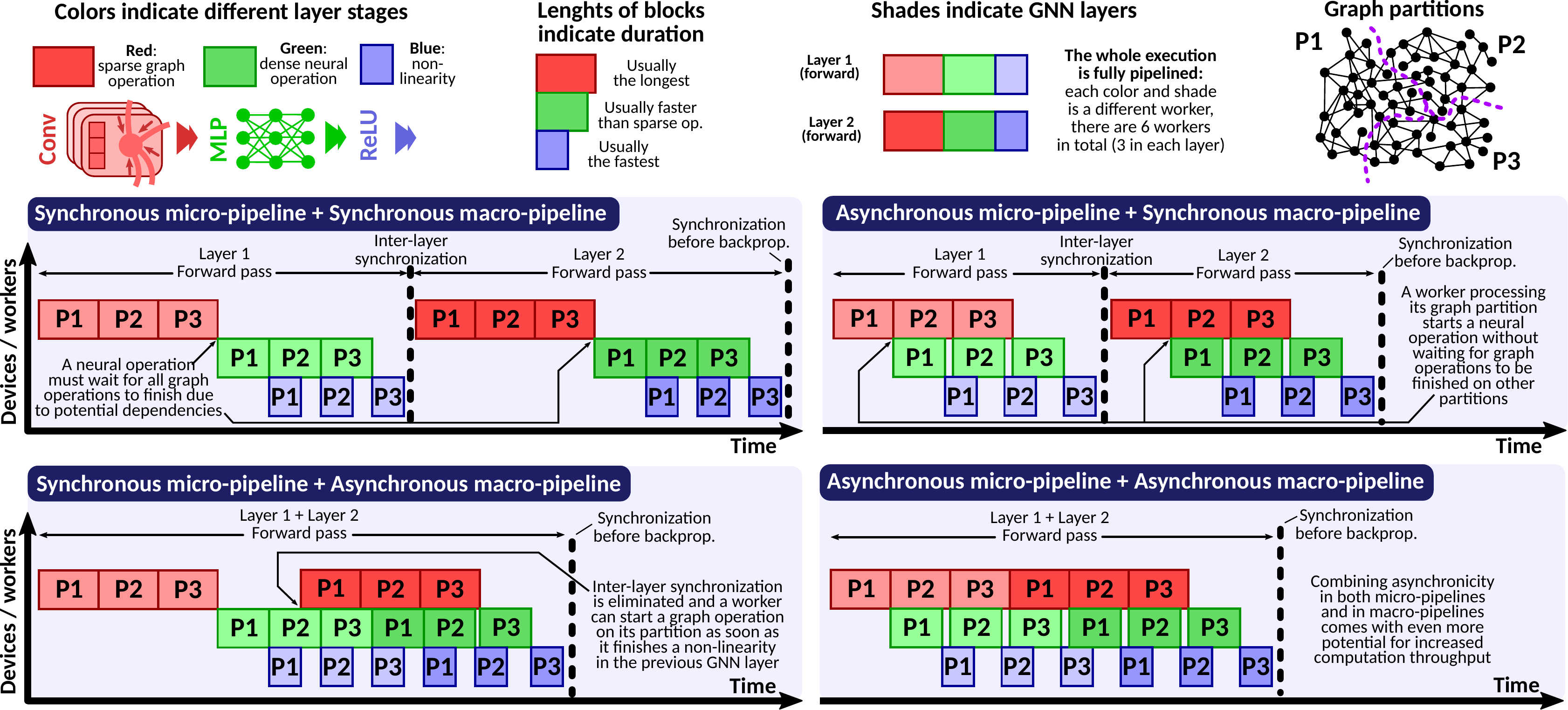}
\vspace{-1.5em}
\caption{\textbf{(\cref{sec:model-pipeline}) Overview of pipelining combined
with graph partition parallelism} (top-left panel), and of \textbf{(\cref{sec:async})
asynchronous pipelined execution (other panels)}.
Each of four example GNN executions processes three graph partitions (P1, P2,
P3) on three stages: \textcolor{red}{\textbf{red}} (a sparse graph operation
such as convolution), \textcolor{green}{\textbf{green}} (a dense neural
operation such as MLP), and \textcolor{blue}{\textbf{blue}}) (a non-linearity
such as ReLU). We use two GNN layers  
(shades indicate layers). The whole execution is fully pipelined, i.e.,
there are 6 workers in total (three workers for each stage in each layer).}
\label{fig:par-model-pipeline}
\vspaceSQ{-1.0em}
\end{figure*}
\else
\begin{figure*}[t]
\vspaceSQ{-1.25em}
\centering
\includegraphics[width=0.9\textwidth]{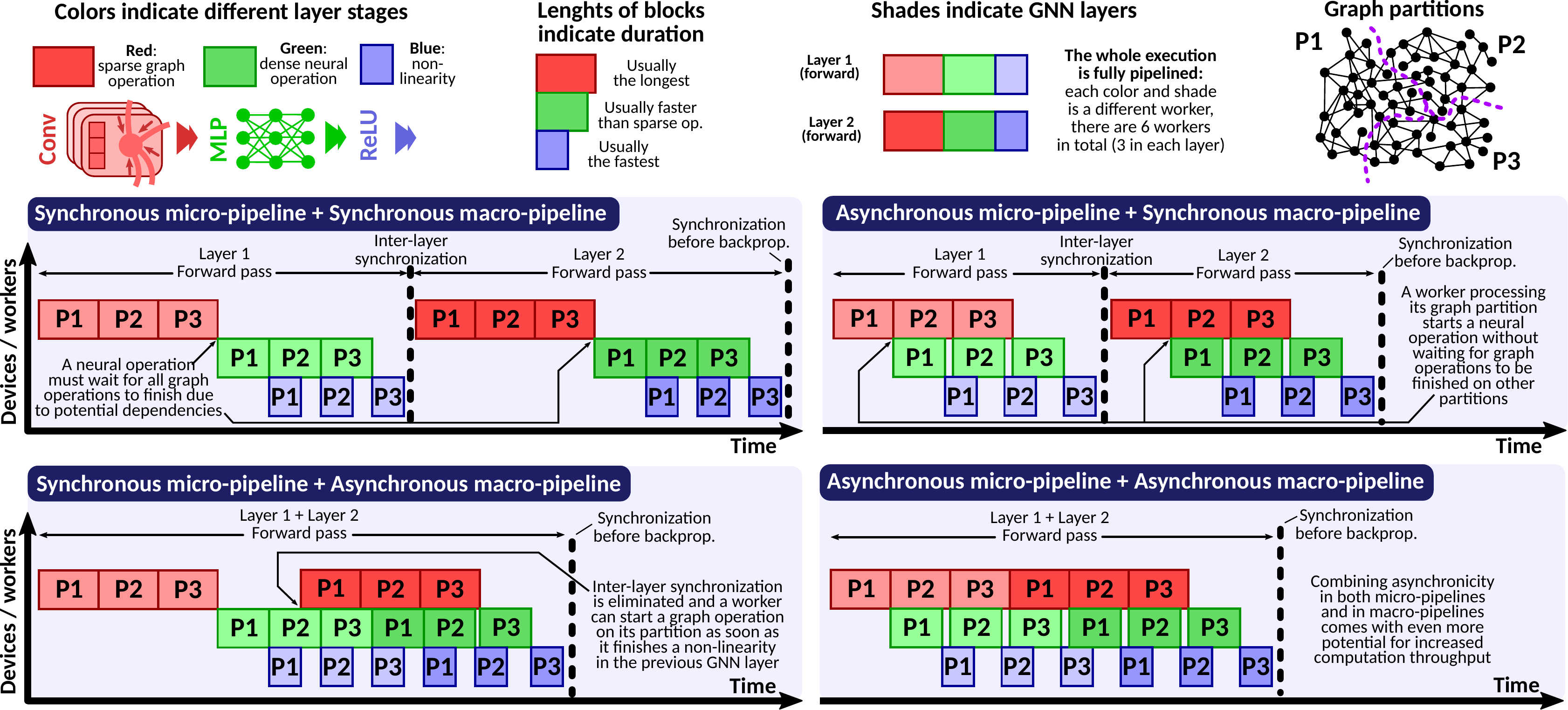}
\vspace{-1em}
\caption{\textbf{(\cref{sec:model-pipeline}) Overview of pipelining combined
with graph partition parallelism} (top-left panel), and of \textbf{(\cref{sec:async})
asynchronous pipelined execution (other panels)}.
Each of four example GNN executions processes three graph partitions (P1, P2,
P3) on three stages: \textcolor{red}{\textbf{red}} (a sparse graph operation
such as convolution), \textcolor{green}{\textbf{green}} (a dense neural
operation such as MLP), and \textcolor{blue}{\textbf{blue}}) (a non-linearity
such as ReLU). We use two GNN layers  
(shades indicate layers). The whole execution is fully pipelined, i.e.,
there are 6 workers in total (three workers for each stage in each layer).}
\label{fig:par-model-pipeline}
\vspaceSQ{-1.0em}
\end{figure*}
\fi

%\marginpar{\large\vspace{3em}\colorbox{ly}{\textbf{R2}}\\\colorbox{ly}{\textbf{3.1}}}

\sethlcolor{ly}
\textbf{{Communication \& Synchronization}}
{in the GL formulations heavily depend on the
used matrix representations and operations. Specifically, there have been a
plethora of works into communication reduction in matrix operations, for
example targeting dense matrix
multiplications~\mbox{\cite{Georganas:2012:CAO:2388996.2389132,
kwasniewski2021parallel, kwasniewski2021pebbles, kwasniewski2019red}} or sparse
matrix multiplications~\mbox{\cite{DBLP:journals/corr/SolomonikH15,
solomonik2017scaling, solomonik2014tradeoffs, gleinig2022io}}.
They could be used with different GNN operations (cf.~Table~\mbox{\ref{tab:la-ops}})
and different models (cf.~Table~\mbox{\ref{tab:models-la}}).
The exact bounds would depend on the selected schemes. Importantly,
many works propose to trade more storage for less communication
by different forms of input matrix replication~\mbox{\cite{solomonik2014tradeoffs}}.
This could be used in GNNs for more performance.}

\sethlcolor{yellow}

% \subsubsection{Instantiation of Edge Feature Vectors}
% \vspaceSQ{-0.25em}

%\marginpar{\large\vspace{0em}\colorbox{ly}{\textbf{R2}}\\\colorbox{ly}{\textbf{3.1}}}
\subsubsection{Discussion}

% \subsubsection{Feature vs.~Structure vs.~Model Weight Parallelism}
% \label{sec:model-par-f-s-m}
% \vspaceSQ{-0.25em}

\sethlcolor{ly}
\textbf{{Feature vs.~Structure vs.~Model Weight Parallelism}}
\sethlcolor{yellow}
Feature parallelism is straightforward in both LC and GL formulations
(cf.~Figure~\ref{fig:op-par}). In the former, one can execute binary tree
reductions over different features in parallel (feature parallelism
in~$\bigoplus$), or update edge or vertex features in parallel (feature
parallelism in~$\psi$ and $\phi$). In the latter, one can multiply a row
of an adjacency matrix with any column of the latent matrix~$\mathbf{H}$
(corresponding to different features) in parallel. As feature vectors are
dense, they can be stored contiguously in memory and easily used with
vectorization.

%\marginpar{\large\vspace{0em}\colorbox{yellow}{\textbf{R2}}\\\colorbox{yellow}{\textbf{3.2}}}

Graph {neighborhood} parallelism is available in both LC and GL formulations. In
the former, it is present via parallel execution of $\bigoplus$ (for a single
specific feature).  In the latter, one parallelizes the multiplication
of a given adjacency matrix row with a given feature matrix column.

Traditional model weight parallelism, in which one partitions the weight
matrix~$\mathbf{W}$ across workers, is also possible in GNNs. Yet, due to
the small sizes of weight matrices used so far in the
literature~\cite{hu2020open, dwivedi2020benchmarking}, it was not yet the focus
of research. If this parallelism becomes useful in the feature, one could use
traditional deep learning techniques to parallelize the model weight
processing~\cite{ben2019demystifying}.

% \subsubsection{Communication \& Synchronization}
% \label{sec:comm-sync}
% \vspace{-0.25em}

%\marginpar{\large\vspace{3em}\colorbox{yellow}{\textbf{R2}}\\\colorbox{yellow}{\textbf{3.2}}}

\textbf{{Graph [Neighborhood] vs.~Graph [Partition] Parallelism}}
{Graph [partition] parallelism is used to partition the graph as a whole
among workers in order to contain the graph fully in memory. Graph [neighborhood]
parallelism is used solely among the neighbors of a single vertex during
aggregation, and it can be applied orthogonally to graph [partition]
parallelism. 
For example, a single graph partition may contain a high-degree vertex,
and the aggregation applied to this vertex can be parallelized within
this partition.}

{The relation between graph partition parallelism and graph neighborhood 
parallelism is similar to that of macro-pipeline and micro-pipeline
parallelism. Specifically, while in macro-pipeline parallelism, one
partitions whole GNN layers among the workers, parts of this single layer
can be further parallelized using micro-pipeline parallelism, orthogonally
to the macro-pipeline structure.}

\begin{figure*}[b]
\vspaceSQ{-1.5em}
%\small
\begin{flalign}
\multicolumn{2}{l}{\text{\textbf{Standard computation with graph partition parallelism:}}}\nonumber\\
\mathbf{h}^{(t, l)}_i & = \phi \left( \mathbf{h}^{(t,l-1)}_i, \bigoplus_{j \in N^\mathcal{L}(i)} \psi\left(\mathbf{h}^{(t,l-1)}_i, \mathbf{h}^{(t,l-1)}_j \right) \bigoplus_{j \in N^\mathcal{R}(i)} \psi\left(\mathbf{h}^{(t,l-1)}_i, \mathbf{h}^{(t,l-1)}_j \right) \right) \label{eq:mpgnn-s} \\
\multicolumn{2}{l}{\text{\textbf{Using bounded stale feature vectors with graph partition parallelism (worst case):}}}\nonumber\\
\mathbf{h}^{(t, l)}_i & = \phi \left( \mathbf{h}^{\fRB{t-T_{\phi},l-L_{\phi}}}_i, \bigoplus_{j \in N^\mathcal{L}(i)} \psi\left(\mathbf{h}^{\fRB{t-T_{\phi},l-L_{\phi}}}_i, \mathbf{h}^{\fRB{t-T^\mathcal{L}_{\psi},l-L^\mathcal{L}_{\psi}}}_j \right) \bigoplus_{j \in N^\mathcal{R}(i)} \psi\left(\mathbf{h}^{\fRB{t-T_{\phi},l-L_{\phi}}}_i, \mathbf{h}^{\fRB{t-T^\mathcal{R}_{\psi},l-L^\mathcal{R}_{\psi}}}_j \right) \right) \label{eq:mpgnn-as}
\end{flalign}
\vspaceSQ{-1.5em}
\caption{\textbf{(\cref{sec:async}) Message Passing GNN
formulation that includes graph partition parallelism combined with fully
synchronous (top) and potentially stale asynchronous computation (bottom).}
The equation generalizes the Message Passing formulation~\cite{gilmer2017neural} and
past synchronous GCN models~\cite{wan2022pipegcn}.}
\label{fig:part-async}
\vspaceSQ{-1em}
\end{figure*}

\subsection{Global Operator Parallelism}

% \textcolor{red}{We do not know of... global operator parallelism in GNNs. Thus, we only sketch its
% characteristics, indicating that it could be a source of future efficient
% parallelization designs.}

{In global operator parallelism, one executes Scatter \& Gather
in parallel over \emph{collections} of vertices or edges.
This approach could utilize additional structure on top of the processed graph
for more performance.
For example, one could harness primitives similar to graph contraction and supervertices (used in Karger's algorithm for
min-cuts~\cite{karger1993global}) or hooking trees (used in Shiloach and Vishkin's algorithm for
connected components~\cite{shiloach1982logn}). Such supervertices or trees could potentially
be used to speed up Gather.}

\textbf{Global Operator vs.~Graph Partition Parallelism}
Graph partition parallelism is a straightforward approach that only distributes
vertices and edges across different workers so that they can all fit into
memory.
Global operator parallelism, on the contrary, is more complex and it can harness additional
structure, such as supervertices, on top of the processed collections of
vertices/edges.

% Global operator parallelism is similar to graph partition parallelism in that
% both parallelize the processing of the graph structure. 

\subsection{Pipeline Parallelism}
\label{sec:model-pipeline}
\vspaceSQ{-0.25em}

Pipelining has two general benefits. First, it increases the throughput of a
computation, lowering the overall processing runtime. This is because more
operations can finish in a time unit. Second, it reduces memory pressure in the
computation. Specifically, one can divide the input dataset into chunks, and
process these chunks separately via pipeline stages, having to keep a fraction
of the input in memory at a time.
In GNNs, pipelining is often combined with graph partition parallelism, with
partitions being such chunks. 
We distinguish two main forms of GNN pipelines: micro-pipelines and
macro-pipelines, see Figure~\ref{fig:par-overview} and~\cref{sec:gnns-taxonomy}.

\subsubsection{Micro-Pipeline Parallelism}

In micro-pipeline parallelism, the pipeline stages correspond to the operations
within a GNN layer. Here, for simplicity, we consider a \textcolor{red}{graph
operation} followed by a \textcolor{green}{neural operation}, followed by
a \textcolor{blue}{non-linearity}, cf.~Figure~\ref{fig:gnn-layer}.  One can
equivalently consider kernels (Scatter, UpdateEdge, Aggregate, UpdateVertex) or
the associated functions ($\psi, \oplus, \phi$).
Such pipelining enables reducing the length of the sequence of executed
operators by up to 3$\times$, effectively forming a 3-stage operator
micro-pipeline.
There have been several practical works into micro-pipelining GNN operators,
especially using HW accelerators; we discuss them in Section~\ref{sec:systems}.

We show an example micro-pipeline (synchronous) in the top panel of
Figure~\ref{fig:par-model-pipeline}. Observe that each \textcolor{green}{neural
operation} must wait for \emph{all} \textcolor{red}{graph operations} to
finish, because -- in the worst case -- in each partition, there may be
vertices with edges to all other partitions.
This is an important difference to traditional deep learning (and
to a GNN setting with independent graphs, cf.~Figure~\ref{fig:gnn-samples}), where
chunks have no inter-chunk dependencies, and thus \textcolor{green}{neural
processing} of P1 could start right after finishing the \textcolor{red}{graph
operation} on P1.

The exact benefits from micro-pipelining in depth depend on a concrete
GNN model. Assuming a simple GCN, the four operations listed above take,
respectively, $O(\log d)$, $O(1)$, and $O(1)$ depth. Thus, as Aggregate takes
asymptotically more time, one could replicate the remaining stages,
in order to make the pipeline balanced.

\subsubsection{Macro-Pipeline Parallelism}

In macro-pipeline parallelism, pipeline stages are GNN layers.
Such pipelines are subject to intense research in traditional deep learning,
with designs such as GPipe~\cite{huang2019gpipe},
PipeDream~\cite{narayanan2019pipedream}, or Chimera~\cite{li2021chimera}.
However, pipelining GNN layers is more difficult because of dependencies
between data samples, and it is only in its early development stage~\cite{thorpe2021dorylus}. In
Figure~\ref{fig:par-model-pipeline}, the execution is fully pipelined, i.e.,
all layers are processed by different workers.

\subsubsection{Asynchronous Pipelining}
\label{sec:async}

In asynchronous pipelining, pipeline stages proceed without waiting for the
previous stages to finish~\cite{narayanan2019pipedream}.
This notion can be applied to both micro- and macro-pipelines in GNNs.
First, in \textbf{asynchronous micro-pipelines}, a worker processing its graph
partition starts a \textcolor{green}{neural operation} without waiting for
\textcolor{red}{graph operations} to be finished on other partitions
(Figure~\ref{fig:par-model-pipeline}, top-right panel).
Second, in \textbf{asynchronous macro-pipelines}, the inter-layer
synchronization is eliminated and a worker can start a \textcolor{red}{graph
operation} on its partition as soon as it finishes a
\textcolor{blue}{non-linearity} in the previous GNN layer
(Figure~\ref{fig:par-model-pipeline}, bottom-left panel).
Finally, \textbf{both forms can be combined}, see
Figure~\ref{fig:par-model-pipeline} (bottom-right).

Note that asynchronous pipelining can be used with \emph{both graph partitions}
(i.e., asynchronous processing of different graph partitions)
\emph{and with mini-batches} (i.e., asynchronous processing of different mini-batches).

\subsubsection{Theoretical Formulation of Arbitrarily Deep Pipelines}

To understand GNN pipelining better, we first provide a variant of
Eq.~(\ref{eq:mpgnn}), namely Eq.~(\ref{eq:mpgnn-s}), which defines a
\emph{synchronous Message-Passing GNN execution with graph partition
parallelism}. In this equation, we explicitly illustrate that, when computing
Aggregation ($\oplus$) of a given vertex~$i$, some of the aggregated neighbors
may come from ``remote'' graph partitions, where $i$ does not belong; such
$i$'s neighbors form a set $N^\mathcal{R}(i)$. Other neighbors come from the
same ``local'' graph partition, forming a set $N^\mathcal{L}(i)$. Note that
$N^\mathcal{R}(i) \cup N^\mathcal{L}(i) = N(i)$.
Moreover, in Eq.~(\ref{eq:mpgnn-s}), we also explicitly indicate the current
training iteration~$t$ in addition to the current layer~$l$ by using a double
index $(t,l)$.
Overall, Eq.~(\ref{eq:mpgnn-s}) describes a synchronous standard execution
because, to obtain a feature vector in the layer~$l$ and in the training
iteration~$t$, all used vectors come from the previous layer~$l-1$, in the same
training iteration~$t$.

Different forms of staleness and asynchronicity can be introduced by modifying
the layer indexes so that they ``point more to the past'', i.e., use stale
feature vectors from past layers. For this, we generalize
Eq.~(\ref{eq:mpgnn-s}) into Eq.~(\ref{eq:mpgnn-as}) by incorporating parameters
to fully control the scope of such staleness. These parameters are $L_{\phi}$
(controlling the staleness of $i$'s own previous feature vector),
$L^\mathcal{L}_{\psi}$ (controlling the staleness of feature vectors coming
from $i$'s local neighbors from $i$'s partition), and $L^\mathcal{R}_{\psi}$
(controlling the staleness of feature vectors coming from $i$'s remote
neighbors in other partitions).
Moreover, to also \emph{allow for staleness and asynchronicity with respect to
training iterations}, we introduce the analogous parameters $T_\phi,
T^\mathcal{L}_\psi, T^\mathcal{R}_\psi$.
We then define the behavior of Eq.~(\ref{eq:mpgnn-as}) such that these six
parameters \emph{upper bound the maximum allowed staleness}, i.e.,
Eq.~(\ref{eq:mpgnn-as}) can use feature vectors from past layers/iterations
\emph{at most} as old as controlled by the given respective index parameters.

Now, first observe that when setting $L_\phi = L^\mathcal{L}_\psi =
L^\mathcal{R}_\psi = 1$ and $T_\phi = T^\mathcal{L}_\psi = T^\mathcal{R}_\psi =
0$, we obtain the standard synchronous equation 
(cf.~Figure~\ref{fig:par-model-pipeline}, top-left panel).
Setting any of these parameters to be larger than this introduces staleness.
For example, PipeGCN~\cite{wan2022pipegcn} proposes to pipeline communication
and computation \emph{between training iterations} in the GCN
model~\cite{kipf2016semi} by using $T^\mathcal{R}_\psi = 1$ (all
other parameters are zero).  This way, the model is allowed to use stale
feature vectors coming from \emph{remote partitions in previous training
iterations}, enabling communication-computation overlap (at the cost of
somewhat longer convergence).
Another option would be to \emph{only} set $L^\mathcal{R}_\psi = 2$ (or to a
higher value). \emph{This would enable asynchronous macro-pipelining}, because
one does not have to wait for the most recent GNN layer to finish processing
other graph partitions to start processing its own feature vector.
We leave the exploration of other asynchronous designs based on
Eq.~(\ref{eq:mpgnn-as}) for future work.

\begin{figure}[h]
\vspaceSQ{-1em}
\footnotesize
\begin{flalign}
\multicolumn{2}{l}{\text{\textbf{Standard computation:}}}\nonumber\\
\grad \mathbf{h}^{(t, l)}_j & = \sum_{\substack{i \in V:\\ j \in N^\mathcal{L}(i)}} \grad \mathbf{h}^{\fRB{t,l+1}}_i + \sum_{\substack{i \in V:\\ j \in N^\mathcal{R}(i)}} \grad \mathbf{h}^{\fRB{t,l+1}}_i \label{eq:grad-s} \\
\multicolumn{2}{l}{\text{\textbf{Using bounded stale gradients (worst case):}}}\nonumber\\
\grad \mathbf{h}^{(t, l)}_j & = \sum_{\substack{i \in V:\\ j \in N^\mathcal{L}(i)}} \grad \mathbf{h}^{\fRB{t-T^\mathcal{L},l+L^\mathcal{L}}}_i + \sum_{\substack{i \in V:\\ j \in N^\mathcal{R}(i)}} \grad \mathbf{h}^{\fRB{t-T^\mathcal{R},l+L^\mathcal{R}}}_i\;\;\;\; \label{eq:grad-a}
\end{flalign}
\vspaceSQ{-1em}
\caption{\textbf{(\cref{sec:async}) Generalization of computing gradients in
GNNs} to include graph partition parallelism combined with fully synchronous
(top) and potentially stale asynchronous computation (bottom).}
\label{fig:part-async-grad}
\vspaceSQ{-1em}
\end{figure}

Finally, we also obtain the equivalent formulations for the asynchronous
computation of \emph{stale gradients}, see Figure~\ref{fig:part-async-grad}.
This establishes a similar approach for optimizing backward propagation passes.

\subsubsection{Beyond Micro- and Macro-Pipelining}

We note that the above two forms of pipelining do not necessarily exhaust all
opportunities for pipelined execution in GNNs. For example, there is extensive
work on parallel pipelined reduction trees~\cite{hoefler2014energy} that could
be used to further accelerate the Aggregate operator ($\bigoplus$).

%\marginpar{\large\vspace{0em}\colorbox{yellow}{\textbf{R2}}\\\colorbox{yellow}{\textbf{(4)}}}

\subsection{Artificial Neural Network (ANN) Parallelism}
\label{sec:ann-dets}

{In some GNN models such as GIN~\mbox{\cite{xu2018powerful}}, the dense
UpdateVertex or UpdateEdge kernels are MLPs.  They can be parallelized with
traditional DL approaches, which are not the focus of this work; they have been
extensively described elsewhere~\mbox{\cite{ben2019demystifying}}.
Overall, one can use \textbf{ANN-operator parallelism} (parallel processing of
single NN operators within one layer, e.g., computing the value of a single
neuron) and \textbf{ANN-pipeline parallelism} (parallel pipelined processing of
consecutive MLP layers), cf.~Figure~\mbox{\ref{fig:par-overview}}.}

{We identify} \sethlcolor{ly}{an interesting \textbf{difference between
GNN macro-pipelines and the traditional ANN pipelines}. Specifically, in the
latter, the data is only needed at the pipeline beginning. In the former, the
data (i.e., the graph structure) is needed \emph{at every GNN
layer}.}\sethlcolor{yellow}

\subsection{Other Forms of Parallelism in GNNs}
\vspaceSQ{-0.25em}

One could identify other forms of parallelism in GNNs. First, by combining
model and data parallelism, one obtains -- as in traditional deep learning --
\emph{hybrid parallelism}~\cite{krizhevsky2014one}.
More elaborate forms of model parallelism are also possible. An example is
Mixture of Experts (MoE)~\cite{masoudnia2014mixture}, in which different models
could be evaluated in parallel.  Currently, MoE usage in GNNs is in its
infancy~\cite{zhou2019explore, hu2021graph}.

\section{Frameworks, Accelerators, Techniques}
\label{sec:systems}

We finally analyze existing GNN SW frameworks and HW
accelerators\footnote{\scriptsize We encourage participation in this analysis.
In case the reader possesses additional relevant information, such as important
details of systems not mentioned in the current paper version, the authors
would welcome the input.}. For this, we first describe parallel and distributed
architectures used by these systems.

\subsection{Parallel and Distributed Computing Architectures}
\label{sec:par-arch}

\iftr
There are both single-machine (single-node, often shared-memory) or
multi-machine (multi-node, often distributed-memory) GNN systems.
\fi

\subsubsection{Single-Machine Architectures}
\vspaceSQ{-0.25em}

% Parallelism is omnipresent in a single machine. 
%
Multi- or manycore parallelism is usually included in {general-purpose CPUs}.
{Graphical Processing Units (GPUs)} offer massive amounts of
parallelism in a form of a large number of simple cores. However, they
often require the compute problems to be structured so that they fit the
``regular'' GPU hardware and parallelism. 
Moreover, {Field Programmable Gate Arrays (FPGAs)} are well suited for
problems that easily form pipelines. 
\iftr
Finally, novel proposals include {processing-in-memory
(PIM)}~\cite{besta2021sisa, mutlu2020modern} that brings computation closer to
data.
\else
Finally, novel proposals include {processing-in-memory
(PIM)}~\cite{mutlu2020modern} that brings computation closer to
data.
\fi

\if 0
%
Parallelism is omnipresent in a single machine. It is deployed internally in a
single processing core, with mechanisms such as \emph{pipelining} or
\emph{out-of-order execution}. Then, it is also exposed to the developer by
providing \emph{multiple cores}, potentially grouped in \emph{multiple sockets}
and even more complex hierarchies.  Such multi- or manycore parallelism is
usually included in \textbf{general-purpose CPUs}.

There are other classes of single-machine parallel architectures.
%
\textbf{Graphical Processing Units (GPUs)} offer massive amounts of simple
parallelism in a form of a large number of very simple cores. However, they
often require the compute problems to be structured so that they fit the
``regular'' GPU hardware and parallelism. 
%
Moreover, \textbf{Field Programmable Gate Arrays (FPGAs)} enable creating
arbitrary hardware designs suited for solving specific problems, but they offer
limited amounts of on-chip memory and are well suited for problems that easily
form pipelines. 
%
Finally, novel proposals include \textbf{processing-in-memory
(PIM)}~\cite{besta2021sisa, mutlu2020modern} that brings computation closer to
data.
%
\fi

GNNs feature both irregular operations that are ``sparse'' (i.e., entailing
many random memory accesses), such as reductions over neighborhoods, and
regular ``dense'' operations, such as transformations of feature vectors, that
are usually dominated by sequential memory access
patterns~\cite{abadal2021computing}.
The latter are often suitable for effective GPU processing while the former are
easier to be processed effectively on the CPU. Thus, both architectures are
highly relevant in the context of GNNs. Our analysis (Table~\ref{tab:systems},
the top part) indicates that they are both supported by more than 50\% of
the available GNN processing frameworks. 
We observe that most of these designs focus on executing regular dense GNN
operations on GPUs, leaving the irregular sparse computations for the CPU.
While being an effective approach, we note that GPUs were successfully used to
achieve very high performance in irregular graph
processing~\cite{shi2018graph}, and they thus have high potential for also
accelerating sparse GNN operations. 

There is also interest in HW accelerators for GNNs (Table~\ref{tab:systems},
the bottom part). Most are ASIC proposals (some are evaluated using FPGAs);
several of them incorporate PIM.
With today's significance of heterogeneous computing, developing GNN-specific
accelerators and using them in tandem with mainstream architectures is an
important thread of work that, as we predict, will only gain more significance
in the foreseeable future.

\subsubsection{Multi-Machine Parallelism}
\vspaceSQ{-0.25em}

While shared-memory systems are sufficient for processing many datasets, a recent
trend in GNNs is to increase the size of input graphs~\cite{hu2020open}, which
often requires multi-node settings to avoid expensive I/Os.
We observe (Table~\ref{tab:systems}, the top part) that different GNN software
frameworks support distributed-memory environments.
\iftr
However, the majority of them focus on training, leaving much room for
developing efficient distributed-memory frameworks and techniques for GNN
inference. We also note high potential in incorporating  high-performance
interconnect related mechanisms such as Remote Direct Memory Access
(RDMA)~\cite{gerstenberger2013enabling}, SmartNICs~\cite{besta2015active,
hoefler2017spin, di2019network}, or novel network topologies and
routing~\cite{besta2014slim, besta2020fatpaths} into the GNN domain.
\else
However, the majority of them focus on training, leaving much room for
developing efficient distributed-memory frameworks and techniques for GNN
inference.
\fi

\begin{table*}[hbtp]
\vspaceSQ{-0.5em}
\centering
\setlength{\tabcolsep}{1pt}
\ifcnf
\renewcommand{\arraystretch}{0.5}
\fi
% \ifcnf
% \renewcommand{\arraystretch}{0.5}
% \else
% \renewcommand{\arraystretch}{1.2}
% \fi
\scriptsize
%\footnotesize
%\ssmall
%\sf
%
\begin{tabular}{@{}llllllllllll@{}}
\toprule
\makecell[c]{\textbf{Reference}} &
%\makecell[c]{\textbf{Data} \textbf{location}} &
\makecell[c]{\textbf{Arch.}} &
\makecell[c]{{\textbf{Ds?}}} &
\makecell[c]{\textbf{T?}} &
\makecell[c]{\textbf{I?}} &
\makecell[c]{\textbf{Op?}} &
\makecell[c]{\textbf{mp?}} &
\makecell[c]{\textbf{Mp?}} &
\makecell[c]{\textbf{Dp?}} &
\makecell[c]{\textbf{Dpp}} &
\makecell[c]{\textbf{PM}} &
\makecell[c]{\textbf{Remarks}} \\
\midrule
\textbf{[SW]} PipeGCN~\cite{wan2022pipegcn} & CPU+GPU & \faY & \faY\ (fb) & \faN & \faY & \faY & \faN & \faY & sh & LC \\
\textbf{[SW]} BNS-GCN~\cite{wan2022bns} & GPU & \faY & \faY\ (fb) & \faN & \faU & \faU & \faN & \faY & sh \\
\textbf{[SW]} PaSca~\cite{zhang2022pasca} & GPU & \faY & \faY & \faY & \faU & \faU & \faU & \faY & \\
\textbf{[SW]} Marius++~\cite{waleffe2022marius++} & CPU & \faN & \faY\ (mb) & \faN & \faY\ (v) & \faU & \faN & \faY & & LC (SU) & Focus on using disk \\
\textbf{[SW]} BGL~\cite{liu2021bgl} & GPU & \faY & \faY\ (mb) & \faN & \faY & \faY & \faN & \faY & sh & --- \\
\textbf{[SW]} DistDGLv2~\cite{zheng2021distributed} & CPU+GPU & \faY & \faY\ (mb) & \faN & \faY & \faY & \faN & \faY & sh & --- \\
\textbf{[SW]} SAR~\cite{mostafa2021sequential} & CPU & \faY & \faY\ (fb) & \faN & \faN & \faN & \faN & \faY & \\
\textbf{[SW]} DeepGalois~\cite{hoang1050efficient} & CPU & \faY & \faY\ (fb) & \faN & \faU & \faN & \faN & \faY\ (v) & sh & LC (AU) & \\
\textbf{[SW]} DistGNN~\cite{md2021distgnn} & CPU & \faY & \faY\ (fb) & \faN & \faU & \faN & \faN & \faY\ (v) & sh & LC( AU) \faU & \\
\textbf{[SW]} DGCL~\cite{cai2021dgcl} & GPU & \faH$^*$ & \faY & \faN & \faU & \faN & \faN & \faY\ (v) & & LC (AU) & $^*$Only two servers used. \\
\textbf{[SW]} Seastar~\cite{wu2021seastar} & GPU & \faN & \faY & \faN & \faY\ (f) & \faN & \faN & \faY\ (v, t) & & LC (VC) & \\
\textbf{[SW]} \makecell[l]{Chakaravarthy~\cite{chakaravarthy2021efficient}} & GPU & \faY & \faY\ (fb) & \faN & \faN & \faN & \faN & \faY\ (v, sn)  & & \faN & \\
\textbf{[SW]} Zhou et al.~\cite{zhou2021optimizing} & CPU & \faN & \faN & \faY & \faY\ (f) & \faN & \faN & \faU & & \faN & \\
\textbf{[SW]} MC-GCN~\cite{balin2021mg} & GPU & \faH$^*$ & \faY\ (fb) & \faTimes & \faY\ (f) & \faN & \faN & \faY\ (v) & & GL & $^*$Multi-GPU within one node. \\
% \textbf{[SW]} BDS-GCN~\cite{wan2020bds} \\
% \textbf{[SW]} Jangda et al.~\cite{jangda2021accelerating} \\
\textbf{[SW]} Dorylus~\cite{thorpe2021dorylus} & CPU & \faY & \faY\ (fb) & \faN & \faN & \faN & \faY & \faY\ (v) & & LC (SAGA) & \\
%\textbf{[SW]} Min et al.~\cite{min2021large} & GPU & \faH$^*$ & \faY\ (mb) & \faN & \faU & \faN & \faY & \faY\ (v) & & GL & $^*$Multi-GPU within one node. \\
\textbf{[SW]} GNNAdvisor~\cite{wang2020gnnadvisor} & GPU & \faN & \faY & \faY & \faY\ (f, n) & \faN & \faN & \faU & & GL, LC &  \\
\textbf{[SW]} AliGraph~\cite{zhu2019aligraph} & CPU & \faY & \faY & \faY & \faU & \faN & \faU & \faY & & LC (NAU) & \\
\textbf{[SW]} FlexGraph~\cite{wang2021flexgraph} & CPU & \faY & \faY\ (fb) & \faN & \faY\ (n) & \faU & \faN & \faY & & LC (NAU) & \\
% \textbf{[SW]} Lin et ak.~\cite{lin2021accelerating} &  \\
\textbf{[SW]} Kim et al.~\cite{kim2021accelerating} & CPU+GPU & \faN & \faY\ (mb) & \faN & \faY\ (n) & \faU & \faN & \faY & & LC (AU) & \\
\textbf{[SW]} AGL~\cite{zhang2020agl} & CPU & \faY & \faY\ (mb) & \faY & \faN & \faY & \faY & \faY & & MapReduce & \\
\textbf{[SW]} ROC~\cite{jia2020improving} & CPU+GPU & \faY & \faY\ (fb) & \faY & \faN & \faN & \faN & \faY & & \faN & \\
\textbf{[SW]} DistDGL~\cite{zheng2020distdgl} & CPU & \faY & \faY\ (mb) & \faN & \faU & \faN & \faN & \faY & & \faN &   \\
\textbf{[SW]} PaGraph~\cite{lin2020pagraph, bai2021efficient} & GPU & \faH$^*$ & \faY\ (mb) & \faN & \faU & \faN & \faN & \faY & & \faN & $^*$Multi-GPU within one node. \\
\textbf{[SW]} 2PGraph~\cite{zhang20212pgraph} & GPU & \faY & \faY\ (mb) & \faN & \faU & \faY & \faU & \faY & & --- & \\
\textbf{[SW]} GMLP~\cite{zhang2021gmlp} & GPU & \faY & \faY\ (mb) & \faN & \faU & \faN & \faU & \faY & & LC & \\
\textbf{[SW]} fuseGNN~\cite{chen2020fusegnn} & GPU & \faN & \faY & \faN & \faY & \faN & \faN & \faY & & LC (AU)$^*$ & $^*$Two aggregation schemes are used. \\
\textbf{[SW]} P$^3$~\cite{gandhi2021p3} & CPU+GPU & \faY & \faY\ (mb) & \faN & \faY & \faY & \faN & \faY & & LC (SAGA)$^*$ & $^*$A variant called P-TAGS \\
\textbf{[SW]} QGTC~\cite{wang2021qgtc} & GPU & \faN & \faN & \faY & \faY & \faH & \faH & \faY & sh & GL &  \\
% \textbf{[SW]} GEBT~\cite{you2021gebt} \\
% \textbf{[SW]} VQ-GNN~\cite{ding2021vq} \\
% \textbf{[SW]} Degree-Quant~\cite{tailor2020degree} \\
% \textbf{[SW]} Tailor et al.~\cite{tailor2021adaptive} \\
\textbf{[SW]} CAGNET~\cite{tripathy2020reducing} & CPU+GPU & \faY & \faY\ (fb) & \faN & \faY\ (f, n) & \faN & \faN & \faY\ (v, e) & sh+rep & GL & \\
\textbf{[SW]} PCGCN~\cite{tian2020pcgcn} & CPU+GPU & \faN & \faY & \faN & \faN & \faN & \faN & \faY\ (e) & sh & --- & \\
% \textbf{[SW]} HAG~\cite{jia2020redundancy} \\
%\textbf{[SW]} Zhang et al.~\cite{zhang2021understanding} & CPU, GPU & \faN \\
\textbf{[SW]} FeatGraph~\cite{hu2020featgraph} & CPU, GPU & \faN & \faY\ (fb) & \faY & \faY\ (f, n) & \faN & \faN & \faY\ (v) & sh & GL & \\
\textbf{[SW]} G$^3$~\cite{liu2020g3} & GPU & \faN & \faY & \faY & \faU & \faU & \faU & \faU & & GL & \\
\textbf{[SW]} NeuGraph~\cite{ma2019neugraph} & GPU & \faY & \faY & \faN & \faH & \faY & \faN & \faY\ (v, e) & sh & LC (SAGA) \\
\textbf{[SW]} PyTorch-Direct~\cite{min2021large} & GPU & \faY & \faY & \faY & \faY & \faU & \faU & \faY\ (v, e) & & GL, LC & \\ 
\textbf{[SW]} PyG~\cite{fey2019fast} & CPU, GPU & \faY & \faY$^*$ & \faY & \faY & \faU & \faU & \faY\ (v, e) & & GL, LC & $^*$Mini-batching for graph components  \\
\textbf{[SW]} DGL~\cite{wang2019deep} & CPU, GPU & \faY & \faY$^*$ & \faY & \faY & \faU & \faH & \faY &  & GL, LC & $^*$Mini-batching for graph components \\
\midrule
\textbf{[HW]} ZIPPER~\cite{zhang2021zipper} & new & \faN & \faN & \faY & \faH & \faY & \faN & \faY\ (v, e) & sh & GL, LC & \\
\textbf{[HW]} GCNear~\cite{zhou2021gcnear} & new (PIM) & \faN & \faY\ (fb) & \faN & \faY & \faY & \faN & \faY\ (v, e) & sh & LC (AU) & \\
% \textbf{[HW]} Cambricon-G~\cite{song2021cambricon} &  \\
\textbf{[HW]} BlockGNN~\cite{zhou2021blockgnn} & new & \faN & \faN & \faY & \faH & \faY & \faN & \faY & & LC \\
\textbf{[HW]} TARe~\cite{he2021tare} & new (ReRAM) & \faN & \faN & \faY & \faH & \faY & \faN & \faU & & GL &  \\
\textbf{[HW]} Rubik~\cite{chen2021rubik} & new & \faN & \faY\ (mb) & \faN & \faH & \faU & \faN & \faY\ (v, e) & sh & LC (AU) \\
\textbf{[HW]} GCNAX~\cite{li2021gcnax} & new & \faN & \faN & \faY & \faH & \faU & \faN & \faU & & GL \\
\textbf{[HW]} Li et al.~\cite{li2021hardware} & new & \faN & \faN & \faY & \faY & \faY & \faY & \faY & & LC (AU) \\
% \textbf{[HW]} G-Nmp~\cite{tian4007736g} \\
\textbf{[HW]} GReTA~\cite{kiningham2020greta} & new & \faN & \faN & \faY & \faU & \faU & \faU & \faY & sh & LC (GReTA) & \\
\textbf{[HW]} GNN-PIM~\cite{wang2020gnn} & new (PIM) & \faN & \faN & \faY & \faU & \faY & \faN & \faY\ (v) & sh & LC (SAGA) & \\
\textbf{[HW]} EnGN~\cite{liang2020engn} & new & \faN & \faN & \faY & \faH & \faU & \faN & \faY\ (v, e) & sh & \multicolumn{2}{l}{LC (AU + ``feature extraction'' stage)} \\
\textbf{[HW]} HyGCN~\cite{yan2020hygcn} & new & \faN & \faN & \faY & \faY & \faY & \faN & \faY\ (v, e) & sh & LC (AU) \\
\textbf{[HW]} AWB-GCN~\cite{geng2020awb} & new & \faN & \faN & \faY & \faH & \faY & \faY & \faY\ (v, e) & sh & GL \\
\textbf{[HW]} GRIP~\cite{kiningham2020grip} & new & \faN & \faN & \faY & \faU & \faY & \faH & \faY\ (v, e) & sh & \multicolumn{2}{l}{GL, LC (GReTA)} \\
\textbf{[HW]} Zhang et al.~\cite{zhang2020hardware} & new & \faN & \faN & \faY & \faY & \faY & \faH & \faY\ (v, e) & sh & GL  \\ 
\textbf{[HW]} GraphACT~\cite{zeng2020graphact} & new & \faN & \faY\ (mb) & \faN & \faH & \faY & \faN & \faU & & GL \\
\textbf{[HW]} Auten et al.~\cite{auten2020hardware} & new & \faN & \faN & \faY & \faU & \faU & \faU & \faU & & LC (AU) \\
\bottomrule
\end{tabular}
\vspaceSQ{-1em}
\caption{Comparison of GNN processing frameworks and analyses of GNN
implementations. They are grouped by the targeted architecture. Within each
group, the systems are sorted by publication date. ``\textbf{[SW]}'': a
software framework or package, ``\textbf{[HW]}'': a hardware accelerator.
``---'': not relevant for a given system.
{``\textbf{Ds?}''} (distributed memory): does a design target distributed environments such as clusters, supercomputers, or data centers?
``\textbf{Arch.}'': targeted architecture.
``\textbf{new}'': a new proposed architecture. 
``\textbf{T}'': Focus on training? (if more details are provided, we distinguish between ``\textbf{fb}'': full batch, and ``\textbf{mb}'': mini batch).
``\textbf{I}'': Focus on inference?
``\textbf{Op}'': Support for operator parallelism (``\textbf{f}'': feature parallelism, ``\textbf{n}'': {neighborhood} parallelism)?
``\textbf{mp}'': Support for micro-pipeline parallelism?
``\textbf{Mp}'': Support for macro-pipeline parallelism?
``\textbf{Dp}'': Support for graph partition parallelism (if more details are provided, we distinguish between ``\textbf{v}'': vertex
partitioning (also called 1D partitioning), ``\textbf{e}'': edge partitioning
(also called 2D partitioning), ``\textbf{t}'': type
partitioning (in heterogeneous GNNs, where a vertex can have multiple types), or ``\textbf{sn}'': snapshot partitioning
(in dynamic GNNs, where a graph can be stored in multiple snapshots)).
``\textbf{Dpp}'': data partitioning policy (if more details are provided, we distinguish between ``\textbf{sh}'': sharding, and
``\textbf{rep}'': replication).
``\textbf{PM}'': Used programming model or paradigm:
``\textbf{GL}'': Global, linear algebra based, focusing on operations on matrices such
as SpMM or GEMM,
``\textbf{LC}'': Fine Grained, focusing on operations of graph elements,
such as neighborhood aggregation. If more details are provided, we further distinguish
``\textbf{AU}'': Aggregate + Update.
``\textbf{NAU}'': Neighborhood selection + Aggregate + Update.
``\textbf{SAGA}'': Scatter + ApplyEdge (i.e., Update Edge) + Gather + ApplyVertex (i.e., Update Vertex).
``\textbf{GReTA}'': Gather + Reduce (i.e., Aggregate) + Transform (i.e., Update) + Activate.
``\faBatteryFull'': Support.
``\faBatteryHalf'': Partial / limited support.
``\faTimes'': No support.
``\noAnswer'': Unknown.
}
\label{tab:systems}
\vspaceSQ{-1.75em}
\end{table*}

\subsection{General Categories of Systems \& Design Decisions}
\vspaceSQ{-0.25em}

The first systems supporting GNNs usually belonged to one of two categories.
The first category are systems constructed on top of graph processing
frameworks and paradigms that add neural network processing capabilities (e.g.,
Neugraph~\cite{ma2019neugraph} or Gunrock~\cite{wang2016gunrock}).
On the contrary, systems in the second category (e.g., the initial PyG
design~\cite{fey2019fast}) start with deep learning frameworks, and extend it
towards graph processing capabilities.
These two categories usually focus on -- respectively -- the LC and GL
formulations and associated execution paradigms.

The third, most recent, category of GNN systems, does not start
from either traditional graph processing or deep learning. Instead, they target
GNN computations from scratch, focusing on GNN-specific workload
characteristics and design decisions~\cite{rahman2021fusedmm,
qiu2021optimizing, huang2020ge, wang2020gnnadvisor}. For example, Zhang et
al.~\cite{zhang2021understanding} analyze the computational graph of GNNs, and
propose optimizations tailored specifically for GNNs.

\if 0
%
The authors of DGL state two main benefits of the GL approach.  First, the
unified programming and execution paradigm, consisting of only two operations,
facilitates parallelization and other enhancements.  Second, it reduces data
copying, by eliminating intermediate message storage.
%
\fi

\subsection{Parallelism in GNN Systems}
\label{sec:systems-par}
\vspaceSQ{-0.25em}

The most commonly supported form of parallelism is \textbf{graph partition parallelism}.
Here, one often reuses a plethora of established techniques from the graph processing
domain~\cite{bulucc2016recent}. Unsurprisingly, most schemes used for graph
partitioning are based on assigning vertices to workers (``1D partitioning'').
This is easy to develop, but comes with challenges related to load balancing.
Better load balancing properties can often be obtained when also partitioning
each neighborhood among workers (``2D partitioning''). While some frameworks
support this variant, there is potential for more development into this
direction. We also observe that most systems support sharding, attacking a
node or edge classification/regression scenario with a single large
input graph. 
\iftr
Here, CAGNET~\cite{tripathy2020reducing} combines
sharding with replication, in order to accelerate GNN training by communication
avoidance~\cite{solomonik2011communication}, a technique that have been
used to speed up many linear algebra based computations~\cite{besta2020communication}.
\else
Here, CAGNET~\cite{tripathy2020reducing} combines
sharding with replication, in order to accelerate GNN training by communication
avoidance~\cite{solomonik2011communication}.
\fi

The majority of works use graph partition parallelism on its own, to alleviate
large sizes of graph inputs (by distributing it over different memory
resources) and to accelerate GNN computations (by parallelizing the execution
of one GNN layer). Some systems~ZIPPER~\cite{zhang2021zipper} combine graph partition parallelism
with pipelining, offering further speedups and reductions in used storage
resources.

\textbf{Operator parallelism} is supported by most systems. Both
\textbf{feature parallelism} and {\textbf{neighborhood parallelism}} have been
investigated, and there are systems supporting both
(FeatGraph~\cite{hu2020featgraph}, GNNAdvisor~\cite{wang2020gnnadvisor},
CAGNET~\cite{tripathy2020reducing}). Most of these systems target these forms
of parallelism explicitly (e.g., by programming binary reduction
trees processing Reduce). For example, Seastar~\cite{wu2021seastar} focuses on
combining feature-level, vertex-level,
and edge-wise parallelism.
On the other hand, CAGNET is an example design that supports operator
parallelism implicitly, by incorporating 2D and 3D distributed-memory matrix
products and the appropriate partitioning of $\mathbf{A}$ and $\textbf{H}$
matrices.
We note that existing works often refer to operator parallelism differently
(e.g., intra-phase dataflow''~\cite{garg2022understanding}).

%\marginpar{\large\vspace{-12em}\colorbox{yellow}{\textbf{R2}}\\\colorbox{yellow}{\textbf{3.2}}}

\textbf{Micro-pipeline parallelism} is widely supported by HW accelerators.  We
conjecture this is because it is easier to use a micro-pipeline in the HW
setting, where there already often exist such pipeline dedicated HW resources.

\textbf{Macro-pipeline parallelism} is least often supported. This is caused by its
complexity: one must consider how to pipeline the processing of interrelated
nodes, edges, or graphs, across GNN layers. While it is
relatively straightforward to use pipeline parallelism when processing graph
samples in the context of graph classification or regression, most 
systems focus on the more complex node/edge classification or regression.
Here, two examples are GRIP~\cite{kiningham2020grip} and work by Zhang et
al.~\cite{zhang2020hardware}, where pipelining is enabled by simply loading the
weights of the next GNN layer, while the current layer is being processed.
%
% Other systems, such as Dorylus~\cite{thorpe2021dorylus}, offer
% pipelining of graph partitions.

\if 0

Traditional model parallelism forms that, for example, parallelize obtaining
gradients, are not a subject of focus in the current GNN landscale. This is
because the parameter counts in GNN models are usually very small compared to
traditional deep learning, and one simply replicates the associated data
structures~\cite{hoang1050efficient}.
%
However, we expect that, with the growing complexity of graph problems being
targeted, and the ever-growing compute capabilities, this aspect of parallelism
will also undergo growth. 

\fi

\subsection{Optimizations in GNN Systems}
\label{sec:systems-opt}
\vspaceSQ{-0.25em}

We also summarize parallelism related optimizations.

Frameworks that enable data parallelism use different forms of \textbf{graph
partitioning}.  The primary goal is to minimize the edges crossing graph
partitions in order to reduce communication.  Here, some designs (e.g.,
DeepGalois~\cite{hoang1050efficient}, DistGNN~\cite{md2021distgnn}) use {vertex
cuts}.  Others (e.g., DGCL~\cite{cai2021dgcl}, QGTC~\cite{wang2021qgtc})
incorporate {METIS partitioning}~\cite{karypis1995metis}.
ROC~\cite{jia2020improving} proposes an interesting scheme, in which it uses
{online linear regression} to effectively repartition the graph across
different training iterations, minimizing the execution time.
\iftr
Similar approaches are employed by I-GCN~\cite{geng2021gcn} and Point-X~\cite{zhang2021point}.
\fi

\if 0
%
PCGCN 2D

2PGraph supports locality-aware mini-batch sampling
%
\fi

There are numerous other schemes used for \textbf{reducing communication
volume}.
For example, DeepGalois~\cite{hoang1050efficient} and DGCL~\cite{cai2021dgcl}
use {message combination and aggregation}, DGCL also
balances loads on different interconnect links,
DistGNN~\cite{md2021distgnn} delays updates to optimize communication, Min et
al.~\cite{min2021large} use zero copy in the CPU--GPU setting (GPU threads
access host memory without requiring much CPU interaction) and computation \&
communication overlap, and GNNAdvisor~\cite{wang2020gnnadvisor} and
2PGraph~\cite{zhang20212pgraph} renumber nodes to achieve more locality~\cite{tate2014programming}.
\iftr
GNNAdvisor also provides locality-aware task
scheduling~\cite{huang2021understanding}. 
\fi
2PGraph also increases the amount of
locality with cluster based mini-batching (it increases the number of sampled
vertices that belong to the same neighborhood in a mini-batch), a scheme
similar to the approach taken by the Cluster-GCN model~\cite{chiang2019cluster}.

Moreover, there are many interesting \textbf{operator related optimizations}.
A common optimization is {operator fusion} (sometimes referred to as ``kernel
fusion''~\cite{chen2020fusegnn}), in which one merges the execution of
different operators, for example Aggregate and UpdateVertex~\cite{chen2020fusegnn,
cai2021dgcl, wu2021seastar, wang2019deep}. This enables better on-chip data
reuse and reduces the number of invoked operators. While most systems fuse
operators within a single GNN layer, QGTC offers operator fusion also across
different GNN layers~\cite{wang2021qgtc}.
Other interesting schemes include operator
reorganization~\cite{zhang2021understanding}, in which one ensures that
operators first perform neural operations (e.g., MLP) and only then the updated
feature vectors are propagated. This limits redundant computation.

\iftr

Many systems come with optimizations related to the used \textbf{graph
representation}.  For example, PCGCN~\cite{tian2020pcgcn} provides a hybrid
representation, in which it uses CSR to maintain sparse parts of the adjacency
matrix, and dense bitmaps for dense edge blocks.
Further, many designs focus on ensuring cache friendlines:
DistGNN~\cite{md2021distgnn} uses tiling (referred to as ``blocking''),
PaGraph~\cite{lin2020pagraph} and 2PGraph~\cite{zhang20212pgraph} provide
effective feature caching, and Lin et al.~\cite{lin2021accelerating} propose
lossy compression to fit into GPU memory.

\fi

\if 0
%
\textbf{Thread mapping} Seastar (feature-adaptive thread mapping)
%
\fi

%\marginpar{\large\vspace{-15em}\colorbox{yellow}{\textbf{R2}}\\\colorbox{yellow}{\textbf{3.2}}}

Some systems \textbf{hybridize} various aspects of GNN computing.
For example, Dorylus~\cite{thorpe2021dorylus} executes graph-related sparse operators
(e.g., Aggregate) on CPUs, while dense tensor related operators
(e.g., UpdateVertex) run on Lambdas.
fuseGNN~\cite{chen2020fusegnn} dynamically switches between incorporated
execution paradigms: dense operators such as UpdateVertex are executed with the
GL paradigm and the corresponding GEMM kernels, while sparse computations such
as Aggregate use the graph-related paradigms such as SAGA.
\iftr
Gcod~\cite{you2022gcod} proposes an algorithm and accelerator codesign approach for GCNs.
\fi

\iftr

Many other schemes exist.
For example, Zhang et al.~\cite{zhang2021understanding} reduce memory
consumption (intermediate data are
recomputed in the backward pass, instead of being cached), while He et
al.~\cite{he2021spreadgnn} incorporate serverless
computing~\cite{copik2020sebs} for more efficient GNNs. 
%
% QGTC~\cite{wang2021qgtc} is the first approach for quantized GNNs.

\fi

\iftr

\subsection{Analyses and Evaluations of GNN Systems}

Finally, there are several works dedicated to analyses and evaluations of
various techniques.
Garg et al.~\cite{garg2022understanding} investigate different strategies for
mapping sparse and dense GNN operators on various HW accelerators.
Serafini et al.~\cite{serafini2021scalable} compare sample-based and full-graph
trainig.
Yan et al.~\cite{yan2020characterizing} analyze the performance of GCNs
on GPUs. 
Wang et al.~\cite{wang2021empirical} and Zhang et
al.~\cite{zhang2020architectural} extend this analysis to a wider choice of GNN
models. 
Baruah et al.~\cite{baruah2021gnnmark} introduce GNNMark, a benchmarking suite
for GNNs on GPUs.
Tailor et al~\cite{tailor2021towards} analyze the performance of point cloud
GNNs. 
Qiu et al.~\cite{qiu2021optimizing} analyze the impact of sparse
matrix data format on the performance of GNN computations.

\fi

\iftr
\section{Other Categories \& Aspects of GNNs}
\label{sec:others}
\vspaceSQ{-0.25em}

We also briefly discuss other categories, variants, and aspects of GNN models.

\subsection{Classes of Local GNN Models}

Depending on the details of $\bigoplus$, $\psi$, and $\phi$, one distinguishes
three \textbf{GNN classes}~\cite{petar-gnns}: \emph{Convolutional GNNs}
(C-GNNs), \emph{Attentional GNNs} (A-GNNs), and \emph{Message-Passing GNNs}
(MP-GNNs).
Each class is defined by an equation specifying the feature
vector~$\mathbf{h}^{(l+1)}_i$ of a vertex~$i$ in the next GNN layer $l+1$.
The three equations to update a vertex are as follows:

\vspace{-1em}
\footnotesize
\begin{align}
\mathbf{h}^{(l+1)}_i &= \phi \left( \mathbf{h}^{(l)}_i, \bigoplus_{j \in N(i)} c_{ij} \cdot \psi\left(\mathbf{h}^{(l)}_j\right) \right) & \text{(C-GNNs)} \label{eq:cgnn} \\
\mathbf{h}^{(l+1)}_i &= \phi \left( \mathbf{h}^{(l)}_i, \bigoplus_{j \in N(i)} a\left(\mathbf{h}^{(l)}_i, \mathbf{h}^{(l)}_j \right) \cdot \psi\left(\mathbf{h}^{(l)}_j \right) \right) & \text{(A-GNNs)} \label{eq:agnn} \\
\mathbf{h}^{(l+1)}_i &= \phi \left( \mathbf{h}^{(l)}_i, \bigoplus_{j \in N(i)} \psi\left(\mathbf{h}^{(l)}_i, \mathbf{h}^{(l)}_j \right) \right) & \text{(MP-GNNs)} \label{eq:mpgnn}
\end{align}
\normalsize
\vspace{-1em}

The overall difference between these classes lies in the expressiveness of
operations acting on edges. C-GNNs enable fixed precomputed \emph{scalar} edge
weights~$c_{ij}$, A-GNNs enable \emph{scalar} edge weights that may be the
result of arbitrarily complex operations on learnable
parameters~$a(\mathbf{h}_i, \mathbf{h}_j)$, and MP-GNNs enable arbitrarily
complex edge weights~$\psi(\mathbf{h}_i, \mathbf{h}_j)$.

Specifically, in \textbf{C-GNNs}, Eq.~(\ref{eq:cgnn}), the feature vector
$\mathbf{h}^{(l)}_j$ of each neighbor~$j$ is first transformed using a
function~$\psi$. The outcome is then multiplied by a scalar~$c_{ij}$.
Importantly, $c_{ij}$ is a value that is fixed, i.e., it is known upfront
(before the GNN computation starts), and it does not change across iterations.
Example GNN models that fall into the class of C-GNNs are
GCN~\cite{kipf2016semi} or SGC~\cite{wu2019simplifying}.

In \textbf{A-GNNs}, Eq.~(\ref{eq:agnn}), $\mathbf{h}^{(l)}_j$ is also first
transformed with a function~$\psi$. However, the outcome is then multiplied by
a value~$a \fRB{\mathbf{h}^{(l)}_i, \mathbf{h}^{(l)}_j}$ which is no longer
precomputed. Instead, $a$ is a parameterized function of the model and is
obtained during training. Importantly, while the derivation of this weight can
be arbitrarily complex (e.g., the weight can be computed via attention), the
edge weight itself is a scalar. Overall, A-GNNs come with richer expressive
capabilities and can be used with models such as
GAT~\cite{velivckovic2017graph} or Gated Attention Network
(GaAN)~\cite{zhang2018gaan}.

Finally, \textbf{MP-GNNs} are the most complex LC class, with edge
weights~$\psi$ that can be arbitrary vectors or even more complex objects, as
specified by $\psi$. Example MP-GNN models are G-GCN~\cite{bresson2017residual}
or different models under the umbrella of the Message Passing paradigm
(MPNN)~\cite{gilmer2017neural}.

Explicit division of GNNs models into three classes (C-GNNs, A-GNNs, MP-GNNs)
has its benefits, as it makes it easier to apply various optimizations. For
example, whenever handling a C-GNN model, one can simply precompute edge
weights instead of deriving them in each GNN layer.

We illustrate generic work and depth equations in 
Figures~\ref{fig:work}-\ref{fig:depth}. Overall, work is the sum of any
preprocessing costs~$W_{pre}$, post-processing costs~$W_{post}$, and work of a
single GNN layer~$W_l$ times the number of layers~$L$. In the MP-GNN generic
formulation, $W_l$ equals to work needed to evaluate
$\psi$ for each edge ($m W_\psi$), $\bigoplus$ for each vertex ($n
W_{\oplus}$), and $\phi$ for each vertex ($n W_\phi$).
In C-GNNs, the preprocessing cost belongs to $W_{pre}$ and $D_{pre}$. In
A-GNNs, one needs to also consider the work and depth needed to obtain the
coefficients~$a$.
Depth is analogous, with the main difference that the depth of a single GNN
layer is a plain sum of depths of computing $\psi$, $\bigoplus$, and $\phi$
(each function is evaluated in parallel for each vertex and edge, hence no
multiplication with $n$ or $m$).

\begin{figure}[h]
\footnotesize
\begin{align}
\text{Generic} & = W_{pre} + L  W_{l} + W_{post} \nonumber\\
\text{MP-GNNs} & = W_{pre} + L \cdot \fRB{m  W_\psi + n  W_\oplus + n  W_\phi} + W_{post}\nonumber\\
\text{C-GNN} & = W_{pre} + L \cdot \fRB{m  W_\psi + n  W_\oplus + n  W_\phi} + W_{post} \nonumber\\
\text{A-GNNs} & = W_{pre} + L \cdot \fRB{m  W_\psi + m  W_a + n  W_\oplus + n  W_\phi} + W_{post} \nonumber
\end{align}
\vspaceSQ{-2em}
\caption{Generic equations for work in GNN model classes.}
\label{fig:work}
\end{figure}

\begin{figure}[h]
\vspaceSQ{-2em}
\footnotesize
\begin{align}
\text{Generic} & = D_{pre} + L  D_{l} + D_{post} \nonumber\\
\text{MP-GNNs} & = D_{pre} + L \cdot \fRB{D_\psi + D_\oplus + D_\phi} + D_{post}\nonumber\\
\text{C-GNN} & = D_{pre} + L \cdot \fRB{D_\psi + D_\oplus + D_\phi} + D_{post} \nonumber\\
\text{A-GNNs} & = D_{pre} + L \cdot \fRB{D_\psi + D_a + D_\oplus + D_\phi} + D_{post} \nonumber
\end{align}
\vspaceSQ{-2em}
\caption{Generic equations for depth in GNN model classes.}
\label{fig:depth}
\end{figure}

\subsection{Reductions over $H$-Hop Neighborhoods}
\label{sec:h-hop}
\vspaceSQ{-0.25em}

The LC formulations described by Eq.~(\ref{eq:cgnn})--(\ref{eq:mpgnn}) enable
reductions over $H$-hop neighborhoods, where $H > 1$. This would enable
expressing models such as MixHop~\cite{abu2019mixhop} or
SGC~\cite{wu2019simplifying}.
Such $H$-hop reductions are similar to using polynomial powers of the adjacency
matrix in the GL formulation, i.e., they also enable using the knowledge from
regions of the graph located further than the direct 1-hop neighbors, within
one GNN layer.
In terms of parallelization, the main difference from reductions over 1-hop
neighborhoods is that the number of vertices being reduced may be much larger
(up to $n$), which means that the work and depth of such a reduction become
$O(n)$ and $O(\log n)$, respectively.
Simultaneously, such reductions would require large preprocessing costs, i.e.,
one needs to derive (and maintain) the information of $H$-hop neighbors for
each vertex.

\subsection{Imposing Vertex Order}
\vspaceSQ{-0.25em}

Nearly all considered GNN models assume that $\bigoplus$ is permutation
invariant. However, some GNN models, such as
PATCHY-SAN~\cite{niepert2016learning}, impose explicit vertex ordering.
In such models, the outcome of $\bigoplus$ would be different based on the
order of the input vertices.
This still enables parallelism: analogously to the established parallel prefix
sum problem~\cite{blelloch1990pre}, one could compute such a $\bigoplus$ in
$O(\log d)$ depth and $O(d)$ work (for $d$ input vertices) -- assuming that
$\bigoplus$ is associative. However, there may
be constant overhead of 2$\times$ for both depth and work, compared to the case
where $\bigoplus$ is permutation-invariant.

\subsection{Heterogeneous GNNs}
\label{sec:het-gnns}
\vspaceSQ{-0.25em}

Heterogeneous graphs (HetG)~\cite{shi2016survey, cai2018comprehensive,
yang2020heterogeneous, besta2019demystifying, wang2020survey} generalize simple graphs defined as a
tuple~$(V,E)$ (cf.~Section~\ref{sec:gnns}) in that vertices and edges may have
arbitrary \emph{types} (e.g., person, paper, journal) and \emph{attributes}
(e.g., age, page count, ranking).
There has been research into HetG learning, with recent models such as
Heterogeneous Graph Neural Network~\cite{zhang2019heterogeneous}, Heterogeneous
Graph Attention Network~\cite{wang2019heterogeneous}, or Heterogeneous Graph
Transformer~\cite{hu2020heterogeneous}.
Computations on such models can benefit from forms of parallelism discussed in
this work. However, in addition to this, they also come with potential for
novel forms of parallelism.  For example, in HetG, one often uses different
adjacency matrices to model edges of different types. Such matrices could be
processed in parallel, introducing \textbf{type parallelism}.
Similarly, different attributes can also be processed in parallel,
offering \textbf{attribute parallelism}.

\subsection{Dynamic and Temporal GNNs}
\label{sec:dyn-gnns}
\vspaceSQ{-0.25em}

Another line of works addresses dynamic graph neural
networks~\cite{wu2020comprehensive} with potential for
\textbf{snapshot parallelism}. Specifically, one may
parallelize the processing of different \emph{snapshots} of a graph, taken at
different time points~\cite{chakaravarthy2021efficient}. Such snapshots can be
processed with both the FG and the GL approach. This introduces potential for
new optimizations, such as using different approaches with different snapshots
based on their sparsity properties.

\subsection{Hierarchical GNNs}
\label{sec:h-gnns}
\vspaceSQ{-0.25em}

Some GNN models are \emph{hierarchical}, i.e., one deals with different input
granularities in the same training and inference process. For example, in the
SEAL-C and SEAL-AI models~\cite{li2019semi}, one is primarily interested in
graph classification. However, before running training and inference using
graphs as data samples, the authors first execute GNNs separately on each
graph, using nodes as data samples. In this process, each graph obtains an
embedding vector, that is then used as input for the training process in graph
classification. Such hierarchical approach comes with opportunities for more
elaborate forms of parallelism, for example parallel pipelining of stages of
computations belonging to the different hierarchy levels.

\if 0
%
\subsection{Preprocessing in GNN Models}

\subsection{Higher-Order GNNs}

\maciej{A big dangerous}

- parallelize like in operators, but across respective higher-order structures
%
\fi

\subsection{Spectral GNN Models}

GL formulations of GNNs can be \emph{spatial} or \emph{spectral}; the
difference is in using either the adjacency or the Laplacian matrix.
As Chen et al.~\cite{chen2021bridging} shows, one can transform a spatial
formulation into a spectral one, and vice versa. For simplicity, we focus on
spacial formulations.

\subsection{Preprocessing in GNNs}
\vspaceSQ{-0.25em}

There are different forms of preprocessing in GNNs.
First, different GNN models often require \textbf{preprocessing the adjacency
matrix} and the degree matrix: by \emph{incorporating self-loops}
($\mathbf{\widetilde{A}} = \mathbf{A} + \mathbf{I}$, $\mathbf{\widetilde{D}} =
\mathbf{D} + \mathbf{I}$), with \emph{symmetric normalization} ($\mathbf{A'} =
\mathbf{\widetilde{D}}^{-\frac{1}{2}} \mathbf{\widetilde{A}}
\mathbf{\widetilde{D}}^{-\frac{1}{2}}$), or with \emph{random-walk
normalization} ($\mathbf{\overline{A}} = \mathbf{D}^{-1} \mathbf{A}$).
Second, some spectral GNN models often require a \textbf{spectral
decomposition} of the Laplacian matrix~\cite{chen2021bridging}.
Third, many GNN works propose to reduce or even eliminate using several GNN
layers, and instead \textbf{push the bulk of computation to preprocessing}.
This could involve explicitly considering \emph{multihop
neighborhoods}~\cite{frasca2020sign, abu2019mixhop} or \emph{polynomial} as
well as \emph{rational powers of the adjacency matrix}~\cite{frasca2020sign,
wu2019simplifying, chen2021bridging}.
All these preprocessing routines come with potential for parallel and
distributed execution.

\subsection{Global vs.~Local Approach for GNNs}
\vspaceSQ{-0.25em}

On one hand, there exist many GNN models defined using the GL approach, that
cannot be easily expressed with the LC formulation. Examples are numerous GNN
models that use rational powers of the adjacency or Laplacian matrix, such as
Auto-Regress~\cite{zhou2004learning, zhu2003semi, bengio200611},
PPNP~\cite{ying2018graph, klicpera2018predict, bojchevski2020scaling},
ARMA~\cite{bianchi2021graph}, ParWalks~\cite{wu2012learning}, or
RationalNet~\cite{chen2021bridging}.
On the other hand, many GNN models that were defined with the LC approach, have
no known GL formulations, for example GAT~\cite{velivckovic2017graph},
MoNet~\cite{monti2017geometric}, G-GCN~\cite{bresson2017residual}, or
EdgeConv~\cite{wang2019dynamic}.

\fi

\section{Selected Insights}
\label{sec:eval}
\vspaceSQ{-0.25em}

We now summarize our {insights} into parallel and distributed GNNs, 
pointing to parts with more details.

%\marginpar{\large\vspace{2em}\colorbox{yellow}{\textbf{R1}}}

\begin{itemize}[noitemsep, leftmargin=0.75em]
\item {\textbf{Two biggest bottlenecks in LC GNNS are reduction $\bigoplus$
and the number of layers $K$ in MLPs (in UpdateVertex/UpdateEdge kernels)}
First, our formal analysis illustrates that the depth of a single GNN layer in
all the considered LC models (except for GIN) is $O(\log d + \log k)$. In
today's models, \#features $k$ is many orders of magnitude smaller than the
maximum degree $d$. Thus, the reduction $\bigoplus$ -- which is responsible for
the $\log d$ term -- is the bottleneck in today's LC GNNs, and it is important
to optimize it (e.g., using pipelined reduction
trees~\mbox{\cite{hoefler2014energy}}.
Second, while most GNN models use linear projections to implement
UpdateVertex/UpdateEdge, some models (e.g., GIN) use MLPs with at least one
hidden layer $K$. The depth of such GNN models depends linearly on $K$, and it
then becomes another bottleneck for a GNN layer. In such models, it is
important to also optimize the depth of MLP (e.g., using the ANN-pipeline
parallelism).}
\item {\textbf{Polynomial GL models are the most parallelizable} 
Models such as LINE, DCNN, or GDC, come with lowest overall depth of
one training iteration, being $O(\log k + \log d \log T)$ (where $T$ is the
power of the adjacency matrix used in a given model).  Simultaneously, all the
LC GNN models have the depth of a single iteration being linear with respect to
$L$ (\#GNN layers). However, polynomial models do not use non-linearities,
making their predictive power potentially lower than that of LC GNNs. This
indicates that when scalability is of top importance, it may be desirable to
consider a polynomial GL model.
Moreover, our blueprint for asynchronous LC GNNs may be the key to LC GNNs that
have both high predictive power \emph{and} high parallelism, as asynchronicity
directly reduces the overheads due to $L$; this is one of the
research opportunities that we detail in the next section.}
\item {\textbf{GAT has the highest amount of work, followed by other A-GNNs
and by MP-GNNs} The established Graph Attention Network model comes with the
highest amount of work. Simultaneously, it still is \emph{as parallelizable as
other models} (i.e., its depth is equal to the depth of other models). Hence,
when one has enough parallel workers, the high amount of work needed for GAT is
alleviated. However, with limited parallel resources, it
may be better to consider other GNN models that need less work.}
%
%\marginpar{\large\vspace{4em}\colorbox{yellow}{\textbf{R2}}\\\colorbox{yellow}{\textbf{3.2}}}
\item \textbf{GNNs come with new forms of parallelism} Graph partition parallelism
(Section~\ref{sec:data-par}), closely related to the graph partitioning problem
in graph processing, is more challenging than equivalent forms of parallelism
in traditional deep learning, due to the dependencies between data samples.
Another form, characteristic to GNNs, is graph {neighborhood} parallelism
(\cref{sec:systems-par}).
\item \textbf{GNNs come with rich diversity of tensor operations} Different GNN
models use a large variety of tensor operations. While today's works focus on
the C-GNN style of computations, that only uses two variants of matrix
products, there are numerous others, listed in Table~\ref{tab:la-ops} and
assigned to models in Tables~\ref{tab:models-fg-1}, \ref{tab:models-fg-2}, and
\ref{tab:models-la}.
\item \textbf{Even local GNN formulations heavily use tensor operations} One
could conclude that efficient tensor operations are crucial only for the global
GNN formulations, as these formulations focus on expressing the whole GNN model
with matrices and operations on matrices (cf.~Table~\ref{tab:models-la}).
However, many local GNN formulations have complex tensor operations associated
with UpdateEdge or UpdateVertex operators, see Tables~\ref{tab:models-fg-1}
and~\ref{tab:models-fg-2}.
%
%\marginpar{\large\vspace{2em}\colorbox{ly}{\textbf{R2}}\\\colorbox{ly}{\textbf{1.1}}}
\sethlcolor{ly}
\item {\textbf{Both local and global GNN formulations are important} There are
many GNN models formulated using the local approach, that have no known global
formulation, and vice versa. Thus, effective parallel and distributed execution
is important in both formulations.}
%
%\marginpar{\large\vspace{2em}\colorbox{ly}{\textbf{R2}}\\\colorbox{ly}{\textbf{3.3}}}
\item {\textbf{Local and global GNN formulations welcome different
optimizations} While having similar or the same work and depth, LC 
and GL formulations have potential for different types of
optimizations. For example, global GNN models focus on operations on large
matrices, which immediately suggests optimizations related to -- for example --
communication-avoiding linear algebra~\mbox{\cite{solomonik2014tradeoffs,
solomonik2011communication, kwasniewski2019red}}.  On the other hand, local GNN
models use operators as the ``first class citizens'', suggesting that an
example important line of work would be operator
parallelism such as the one proposed by the Galois
framework~\mbox{\cite{kulkarni2007optimistic}}.}
\sethlcolor{yellow}
\end{itemize}

\section{Challenges \& Opportunities}
\label{sec:challenges}
\vspaceSQ{-0.25em}

Many of the considered parts of the parallel and distributed GNN landscape were
not thoroughly researched. Some were not researched at all. We now list such
challenges and opportunities for future research.

\begin{itemize}[noitemsep, leftmargin=0.75em]
\item \textbf{Efficient GNN inference} Most GNN frameworks focus on training,
but fewer of them also target inference. There is a large potential for developing
high-performance schemes targeting fast inference. 
\iftr
\item \textbf{Advanced mini-batching in GNNs} There is very little work on
advanced mini-batch training and GNN layer pipelining. Mini-batch training of
GNNs is by nature complex, due to the dependencies between node, edge, or graph
samples. While the traditional deep learning saw numerous interesting works
such as GPipe~\cite{huang2019gpipe}, PipeDream~\cite{narayanan2019pipedream},
or Chimera~\cite{li2021chimera}, that investigate mechanisms such as
asynchronous or bi-directional mini-batching, such works are yet to be
developed for GNNs.
\item \textbf{Asynchronous GNNs} The landscape of asynchronous GNNs is largely
unexplored. While we outline the overall framework defined in
Eq.~(\ref{eq:mpgnn-s}), (\ref{eq:mpgnn-as}), (\ref{eq:grad-s}), and
(\ref{eq:grad-a}), implementations, optimizations, and analyses are missing.
There could be a plethora of works improving the GNN performance by
incorporating different forms of asynchrony.
\else
\item \textbf{Advanced mini-batching in GNNs} There is very little work on
advanced mini-batch training and GNN layer pipelining. Mini-batch training of
GNNs is by nature complex, due to the dependencies between node, edge, or graph
samples. While the traditional deep learning saw numerous interesting works
such as GPipe~\cite{huang2019gpipe} or PipeDream~\cite{narayanan2019pipedream},
that investigate 
asynchronous or bi-directional mini-batching, such works are yet to be
developed for GNNs.
Here, our blueprint for
\textbf{asynchronous GNNs} can spearhead novel works.
\fi
\item \textbf{More performance in GNNs via replication} Many GNN works have
explored graph partitioning. However, very few (e.g.,
CAGNET~\cite{tripathy2020reducing}) uses replication for more
performance (e.g., through 2.5D \& -3D matrix multiplications).
\item \textbf{Parallelization of GNN models beyond simple C-GNNs} There is not 
much work on parallel and distributed GNN models beyond the simple seminal
ones from the C-GNN or A-GNN classes, such as GCN~\cite{kipf2016semi}
or GraphSAGE~\cite{hamilton2017inductive}. One 
would welcome works on more complex models from the MP-GNN class,
cf.~Tables~\ref{tab:models-fg-1}-\ref{tab:models-fg-2}.
\item \textbf{Parallelization of GNN models beyond linear ones}
Virtually no research exists on parallel and distributed GNN models of
polynomial and rational types, cf.~Table~\ref{tab:models-la}.
\item \textbf{Parallelization of other GNN settings} Besides very few
attempts~\cite{zhu2019aligraph}, there is no work on parallelizing
heterogeneous GNNs~\cite{zhang2019heterogeneous}, dynamic and temporal
GNNs~\cite{wu2020comprehensive}, or hierarchical GNNs~\cite{li2019semi}. We
predict that parallel and distributed schemes targeting these works will come
with a large number of research opportunities, due to the rich diversity of
these GNN models and the associated graph related problems. 
\iftr
One example idea
would be to use the available techniques from dynamic and streaming graph
processing~\cite{besta2019practice} for GNNs.
\fi
\item \textbf{Achieving large scales} A large challenge is to further
push the scale of GNN computations. When comparing the scale and parameter
counts of models such as CNNs or Transformers with GNNs, it can be
seen that there is a large gap and a lot of research opportunities.
\iftr
\item \textbf{New forms of parallelism} It would be interesting to investigate
effective utilization of other forms of parallelism in GNNs, for example using
Mixture of Experts~\cite{masoudnia2014mixture}.
\fi
\iftr
\item \textbf{Incorporating new HW architectures} While some initial works
exist, there are not many designs on using GNNs with architectures such as
FPGAs~\cite{besta2019graph, de2018transformations}, transactional
processing~\cite{besta2015accelerating}, or processing in
memory~\cite{besta2021sisa, mutlu2019, mutlu2020modern, ghose2019processing,
seshadri2013rowclone, gomez2021benchmarking, oliveira2021damov,
hajinazar2021simdram, seshadri2017ambit, ahn2015scalable_tes}.
\fi
\item \textbf{Incorporating high-performance distributed-memory capabilities}
\iftr
CAGNET~\cite{tripathy2020reducing} illustrated how to scalably execute GNN
training across many compute nodes. It would be interesting to push this
direction and use high-performance distributed-memory developments and
interconnects, and the associated mechanisms for more performance of
distributed-memory GNN computations, using -- for example -- RDMA and RMA
programming~\cite{gerstenberger2013enabling, schmid2016high},
SmartNICs~\cite{di2019network, besta2015active}, serverless
computing~\cite{copik2020sebs}, of high-performance networking
architectures~\cite{besta2018slim, besta2020fatpaths, besta2020highr,
besta2014slim}. Such techniques have been successfully used to accelerate the
related graph processing field~\cite{strausz2022asynchronous}.
\else
CAGNET~\cite{tripathy2020reducing} illustrated how to scalably execute GNN
training across many compute nodes. It would be interesting to push this
direction and use high-performance distributed-memory developments and
interconnects, and the associated mechanisms for more performance of
distributed-memory GNN computations, using -- for example -- RDMA and RMA
programming.
\fi
\iftr
\item \textbf{Incorporating techniques from graph processing} There is more
potential in using graph processing techniques for GNNs. While many such
schemes have already been incorporated, there are
many more to be tried, 
for example sketching and sampling~\cite{besta2019slim, besta2020substream,
  gianinazzi2018communication} or various forms of approximate graph
  processing~\cite{besta2020high, riondato2016fast, borassi2016kadabra,
  riondato2018abra, geisberger2008better, bader2007approximating,
  chazelle2005approximating, dumbrava2018approximate, slota2014complex,
  roditty2013fast, boldi2011hyperanf, echbarthi2017lasas}.
\fi
\iftr
\item \textbf{Incorporating techniques from linear algebra computations}
A lot of work has been done into optimization algebraic operations such as
matrix products~\cite{Georganas:2012:CAO:2388996.2389132,
kwasniewski2021parallel, kwasniewski2021pebbles, kwasniewski2019red,
DBLP:journals/corr/SolomonikH15,
solomonik2014tradeoffs, gleinig2022io}. Many of them could be applied
in the GNN setting, especially in the context of GL GNN formulations. 
\fi
\item {\textbf{System designs for graph predictions}
There are few studies and systems focusing on graph-related predictions.
For example, developing load balancing schemes for mini-batch parallelism where
samples are graphs would be a promising area of study.}
\end{itemize}

%\marginpar{\large\vspace{-4em}\colorbox{yellow}{\textbf{R1}}\\\colorbox{yellow}{\textbf{(A)}}}

\section{Conclusion}

Graph neural networks (GNNs) are one of the most important and fastest growing
parts of machine learning. Parallel and distributed execution of GNNs is a key
to GNNs achieving large scales, high performance, and possibly accuracy. In this work,
we conduct an in-depth analysis of parallelism and distribution in GNNs.
We provide a taxonomy of parallelism, use it to analyze a large number of GNN
models, and synthesise the outcomes in a set of insights as well as
opportunities for future research.
Our work will propel the development of next-generation GNN computations.

\vspace{1em}
{\noindent\footnotesize\textbf{Acknowledgements}
We thank Petar Veličković for useful comments. We thank Hussein Harake, Colin
McMurtrie, Mark Klein, Angelo Mangili, and the whole CSCS team granting access
to the Ault and Daint machines, and for their excellent technical support. We
thank Timo Schneider for help with computing infrastructure at SPCL.
This project received funding from the European Research Council
\raisebox{-0.25em}{\includegraphics[height=1em]{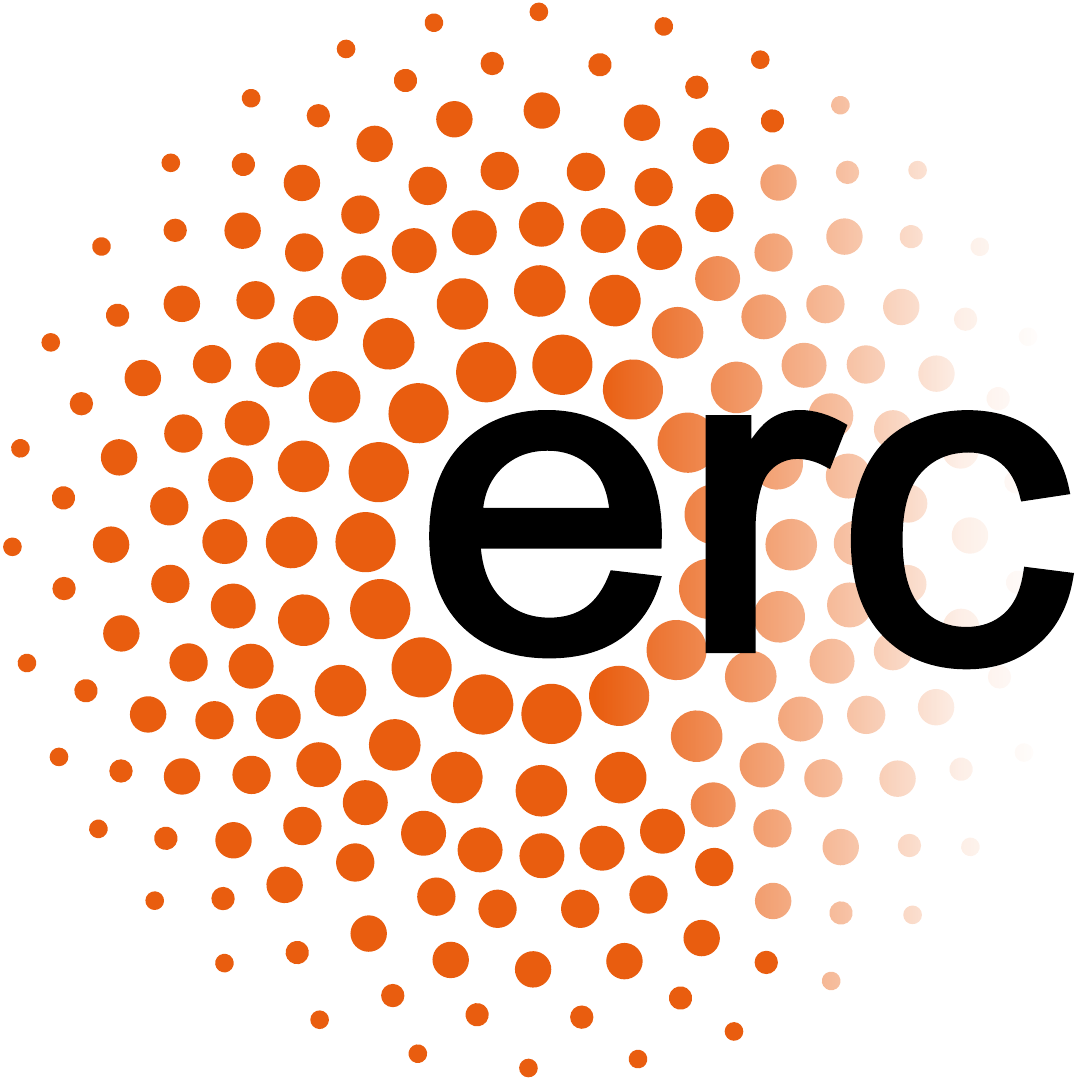}} (Project PSAP,
No.~101002047), and the European High-Performance Computing Joint Undertaking
(JU) under grant agreement No.~955513 (MAELSTROM).
This project was supported by the ETH Future Computing Laboratory (EFCL),
financed by a donation from Huawei Technologies.
This project received funding from the European Union's HE research and
innovation programme under the grant agreement No.~101070141 (Project
GLACIATION).}

\ifcnf
\printbibliography
\else

%\if 0
{%\scriptsize
\bibliographystyle{abbrv}
\bibliography{references}
}
\fi

\ifcnf

\vspace{-3.5em}
\begin{IEEEbiographynophoto}{\ssmall Maciej Besta}
\ssmall
is a researcher at ETH Zurich. He works on understanding and accelerating
large-scale irregular computations, such as graph streaming, graph neural
networks, or graph databases, at all levels of the computing stack.
\end{IEEEbiographynophoto}
\vspace{-4em}
\begin{IEEEbiographynophoto}{\ssmall Torsten Hoefler}
\ssmall
is a Professor at ETH Zurich, where he leads the Scalable Parallel Computing
Lab. His research aims at understanding performance of parallel computing
systems ranging from parallel computer architecture through parallel programming
to parallel algorithms.
\end{IEEEbiographynophoto}

\fi

\end{document}